\documentclass{article}

\usepackage[final,main]{neurips_2026}

\usepackage[utf8]{inputenc}
\usepackage[T1]{fontenc}
\usepackage{textcomp}
\usepackage{marvosym}
\usepackage{microtype}
\usepackage{graphicx}
\usepackage{booktabs}
\usepackage{tabularx}
\usepackage{adjustbox}
\usepackage[table]{xcolor}
\newif\ifshowchanges
\showchangestrue
\newcommand{\revisioncolor}{\ifshowchanges\color{blue}\fi}

\usepackage{subcaption}

\usepackage{tikz}
\usetikzlibrary{positioning}
\usepackage{pgfplots}
\pgfplotsset{compat=1.18}

\usepackage[ruled,vlined]{algorithm2e}
\usepackage{amsmath}
\usepackage{amssymb}
\usepackage{mathtools}
\usepackage{amsthm}

\usepackage{enumitem}
\usepackage{needspace}
\usepackage{etoolbox}
\newcommand{\markrevisionbibitem}[2][]{%
  \color{black}\originalbibitem[#1]{#2}%
  \ifinlist{#2}{\revisedbibkeys}{\revisioncolor}{}%
}
\newcommand{\revisedbibkeys}{}
\AtBeginEnvironment{thebibliography}{%
  \let\originalbibitem\bibitem
  \let\bibitem\markrevisionbibitem
}

\usepackage{listings}
\usepackage{wrapfig}
\usepackage{multirow}
\usepackage{caption}
\usepackage{comment}

\usepackage{tcolorbox}
\tcbuselibrary{breakable}
\usepackage{fvextra}

\usepackage{hyperref}
\hypersetup{hidelinks,
  pdftitle={PEARL: Solver-in-the-Loop Interactive Optimization Modeling from Natural Language},
  pdfauthor={Hongliang Lu, Zhong Li, Yuxuan Chen, Yuan Lan, Fan Zhang, Zaiwen Wen}}
\usepackage[capitalize,noabbrev]{cleveref}

\tcbuselibrary{skins, listings}
\theoremstyle{plain}

\theoremstyle{definition}

\theoremstyle{remark}

\newcommand{\projectname}
{\text{PEARL}}

\title{PEARL: Solver-in-the-Loop Interactive Optimization Modeling from Natural Language}
\author{%
  Hongliang Lu$^{1*}$ \quad Zhong Li$^{2*}$ \\
  \textbf{Yuxuan Chen}$^{1}$ \quad \textbf{Yuan Lan}$^{3}$ \quad
  \textbf{Fan Zhang}$^{3}$ \quad \textbf{Zaiwen Wen}$^{1}$\textsuperscript{(\Letter)} \\
  $^{1}$Peking University \quad $^{2}$Great Bay University \\
  $^{3}$Huawei Technologies Co., Ltd \\
  \textsuperscript{(\Letter)}Corresponding author:
  \texttt{wenzw@pku.edu.cn} (Zaiwen Wen) \\
  $^{*}$Equal contributions.
}

\begin{document}

\maketitle

{
\begin{abstract}
\noindent
Optimization modeling is the process of translating real-world decision problems, often described in natural language, into formal mathematical formulations and executable solver code. While recent advances in large language models have shown promise in automating this process, many approaches focus on one-shot generation or rely on predefined execution-and-repair workflows, without explicitly learning a policy for tool use and revision. This stands in sharp contrast to real-world optimization modeling, which is inherently interactive and proceeds through repeated solve--debug--revise cycles. We introduce PEARL, a system for multi-turn solver-interactive optimization modeling that uses Python execution and mathematical programming solvers inside this loop. Rather than relying on a fixed repair workflow, PEARL learns when to test partial models, how to revise from solver diagnostics, and when to stop. It operates in a multi-turn tool-integrated setting where intermediate execution results, feasibility signals, and solution checks are used to improve both formulations and solver code before finalization. Across diverse optimization benchmarks, PEARL substantially improves executable objective agreement over the same-scale base model and is competitive with strong one-shot and tool-augmented baselines. Notably, PEARL-Qwen3-\textbf{4B} outperforms the much larger DeepSeek-V3.2-\textbf{685B} baseline under direct prompting on both macro- and micro-averaged Pass@1.
Code and datasets are provided at \url{https://github.com/AuroraLHL/PEARL}.
\end{abstract}
}

\section{Introduction}
\label{sec:intro}

Optimization modeling is the process of translating real-world decision problems, often described in natural language (NL), into formal mathematical models and generating corresponding solver code. It provides the critical link between high-level problem descriptions and the low-level solvers~\citep{anand2017comparative} that compute optimal or feasible decisions. Across domains~\citep{singh2012overview,antoniou2007practical} ranging from energy systems to logistics, finance, and manufacturing, the quality of this modeling step largely determines whether optimization can be applied. However, modeling typically requires substantial domain knowledge, careful specification of variables and constraints, and repeated debugging of data pipelines and solver implementations~\citep{huang2025orlm,jiang2025llmopt}. In practice, even seemingly minor issues---such as a missing constraint, an incorrect index, or inconsistent data bindings---can make a model infeasible or lead to silent but severe errors.

Recent advances in large language models (LLMs) have spurred growing interest in automating optimization modeling from natural language~\citep{ijcai2025p1192}. Existing approaches can be broadly grouped into three categories. \emph{Prompting-based} methods~\citep{prasath2023synthesis} treat optimization modeling as a single-pass or weakly iterative generation task. \emph{Workflow-based} systems~\citep{xiao_chain--experts_2024,pmlr-v235-ahmaditeshnizi24a} decompose modeling into multiple stages such as drafting, execution, and repair, typically orchestrated by fixed prompting logic or search procedures. \emph{Training-based} approaches~\citep{huang2025orlm,lu2025optmath,chen2025solverinformed} improve base models through supervised fine-tuning or reinforcement learning (RL) on curated optimization-modeling data.

\begin{figure}[h]
    \centering
    \includegraphics[width=0.95\linewidth]{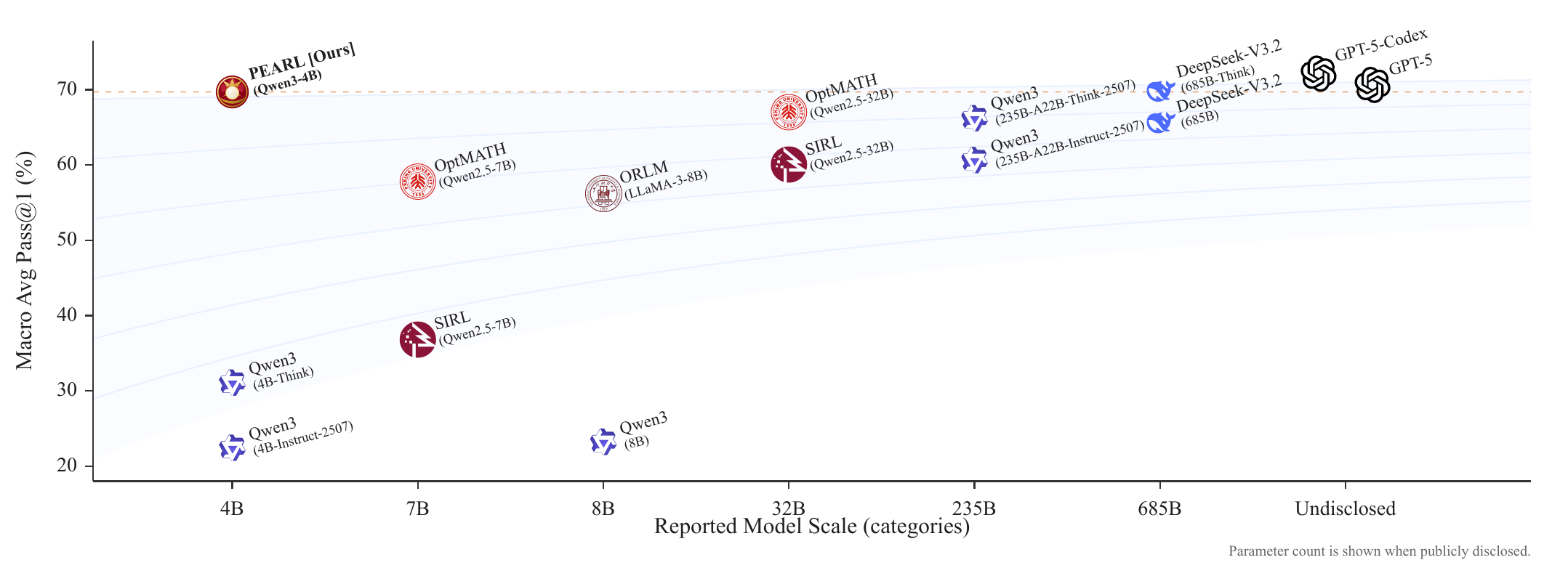}
    \caption{{Macro-averaged Pass@1 versus reported model scale across representative optimization-modeling methods. PEARL (Ours) uses a 4B backbone yet matches or exceeds much larger general-purpose models such as DeepSeek-V3.2-685B, highlighting that solver-mediated interaction and optimization-specific training can deliver strong accuracy in a compact model regime that is more practical for repeated agentic invocation~\citep{kwon2023efficient,abdin2024phi3,belcak2025small}.}}
    \label{fig:main_results_comp}
\end{figure}

\looseness=-1
Across these paradigms, LLMs demonstrate a nontrivial ability to capture modeling structure.
However, {{many training-based approaches} still {optimize single-pass} generation{, while workflow-based systems often prescribe the interaction policy}}. Even in workflow-based systems~\citep{xiao_chain--experts_2024,pmlr-v235-ahmaditeshnizi24a}, the control flow that governs tool use and revision is typically hand-designed, and models are not explicitly trained to make multi-turn decisions based on solver feedback. {{In the single-pass setting}}{, given} a natural-language (NL) specification, the model is expected to produce a complete formulation and solver implementation in a single pass, without executing partial models, conditioning on solver diagnostics, or revising errors. As a result, generated outputs frequently hallucinate constraints~{\citep{li2026optargus,wu2026fix}}, produce infeasible formulations, or fail at runtime. Some recent systems~\citep{chen2025solverinformed} iteratively execute and repair solver code using tool feedback, but these interactions are orchestrated by heuristics rather than learned policies. Crucially, the underlying models are not trained to decide when to invoke tools, how to interpret solver feedback, or whether to revise or terminate. This mismatch between training objectives and real-world modeling workflows limits performance, particularly as problem descriptions become longer, noisier, and {more {structurally complex}}.

In contrast, real-world optimization modeling is inherently interactive~\citep{zhang-etal-2024-solving}. Human modelers rarely produce a correct model in a single pass; instead, they follow a solve--debug--revise loop. Partial formulations are tested, solver errors or infeasibility reports are inspected, constraints or data processing are adjusted, and the process repeats until the model passes verification. This workflow naturally involves temporally extended decision-making, partial observability, and tool-mediated feedback.
These characteristics align closely with the emerging paradigm of \emph{agentic reinforcement learning} \citep{zhang_agentrl_2025}, which treats LLMs as policies operating in multi-turn, tool-integrated environments rather than as one-shot generators.

Motivated by this perspective, we argue that optimization modeling should be formulated not as static text generation, but as an \emph{interactive decision process}. To this end,
we introduce \projectname\ (interactive o\underline{\textbf{p}}timization mod\underline{\textbf{e}}ling with tool-integrated \underline{\textbf{a}}gentic \underline{\textbf{r}}einforcement \underline{\textbf{l}}earning), a method for modeling and solving optimization problems via tool-integrated agentic reinforcement learning (Agentic RL).
In \projectname, an agent incrementally constructs mathematical formulations and solver implementations, invokes tools such as Python execution and optimization solvers, observes execution traces and solver feedback, and revises its outputs across multiple turns {before finalizing its answer}. Rather than relying on fixed heuristics or prompting logic, the agent is trained to learn \emph{how} to interact with tools, \emph{when} to test partial models, \emph{how} to repair failures, and \emph{when} to terminate. {Here, interaction refers to iterative execution and revision using solver feedback; it does not include clarification of missing requirements with a human user.}

To support this behavior, we design an optimization-specific Agentic RL environment with structured tool interfaces and solver-based observations. We further introduce a verifiable reward decomposition that combines interaction-format validity with {{objective agreement after successful execution}}, while execution traces and feasibility signals are fed back as observations for revision. This design enables long-horizon credit assignment over modeling decisions, rather than only over final outputs. Building on recent advances in tool-integrated Agentic RL, we adapt training recipes and curricula to the unique challenges of optimization tasks, including sparse terminal rewards, solver nondeterminism, and heterogeneous failure modes arising from  data processing, modeling, and implementation errors.

We evaluate \projectname\ on a diverse set of optimization modeling benchmarks, ranging from standard natural-language-to-optimization (NL-to-Opt) tasks to long-context and industrial-style problem descriptions. Across these settings, training optimization modeling as an interactive agent leads to substantial improvements in verified solve rates over strong one-shot and tool-augmented baselines. In particular, \projectname\ {reduces {code execution errors} and {output-format violations}}, highlighting the benefits of explicitly learning solve--debug--revise behavior. As previewed in Figure~\ref{fig:main_results_comp}, this is not merely a scale effect: a 4B PEARL policy trained for solver-mediated interaction outperforms the much larger DeepSeek-V3.2-685B in macro-averaged Pass@1, while its compact backbone makes repeated agentic invocation more practical by reducing the serving burden of multi-turn tool use~\citep{kwon2023efficient,abdin2024phi3,chen2024octopusv2,belcak2025small}.

\textbf{Contributions.}
Our main contributions are threefold:
(1) We reconceptualize optimization modeling as an inherently \emph{interactive} process rather than a one-shot generation task, and formalize it as a multi-turn, tool-integrated decision problem in which solver execution and diagnostic feedback are treated as part of the agent’s observable state;
(2) Building on this formulation, we introduce {{an optimization-specific}} agentic reinforcement learning approach, enabling the agent to learn when to execute, revise, and terminate through solver-grounded rewards and an optimization-specific training curriculum;
(3) Through extensive experiments across diverse NL-to-Opt benchmarks, we demonstrate that PEARL substantially improves verified solve rates over existing baselines; notably, our PEARL-Qwen3-\textbf{4B} outperforms the much larger DeepSeek-V3.2-\textbf{685B}~\citep{liu2025deepseek} in both macro- and micro-averaged accuracy on optimization modeling tasks (see Figure~\ref{fig:main_results_comp}){.}
{{Our methodological contribution lies in the optimization-specific interaction environment and policy training, built on established RL optimization methods}.}

\section{Related Work}
\label{sec:related}

\noindent\textbf{LLMs for optimization modeling.}
A growing body of work studies natural-language-to-optimization modeling, covering decision variables, objectives, constraints, and executable solver code~\citep{ijcai2025p1192}. Prompting-based methods rely on templates or in-context examples to elicit formulations from pretrained LLMs~\citep{prasath2023synthesis}. Workflow-based systems decompose modeling into drafting, execution, and repair, using hand-designed control logic or search, as in Autoformulation~\citep{astorga2025autoformulation}, Chain-of-Experts~\citep{xiao_chain--experts_2024}, and OptiMUS~\citep{pmlr-v235-ahmaditeshnizi24a}. Supervised fine-tuning methods such as ORLM~\citep{huang2025orlm}, LLMOPT~\citep{jiang2025llmopt}, and OptMATH~\citep{lu2025optmath} improve formulation and code generation.
OptiMind~\citep{chen_optimind_2025} uses class-specific error analysis to clean fine-tuning data and guide multi-turn inference with hints and solver feedback.

Recent RL-based methods use solver outcomes as verifiable supervision. SIRL~\citep{chen2025solverinformed} optimizes complete generations using executability, feasibility, and correctness rewards, while OR-R1~\citep{ding2026or}, StepORLM~\citep{zhou_steporlm_2025}, and STORM~\citep{tang_calm_2025} explore richer objectives, curricula, or step-level signals. MIND~\citep{liu2026localizable} pairs error-targeted data synthesis with reinforcement learning and supervised updates on teacher-corrected rollouts. Several recent systems also use iterative or agent-style optimization workflows, including OptLLM~\citep{zhang-etal-2024-solving}, OptiTree~\citep{liu2026optitree}, LLMOPT~\citep{jiang2025llmopt}, and MM-Agent~\citep{liu2026mmagent}.
MURKA~\citep{xie2025murka} combines a GRPO-trained Extractor, a distilled Solver, and Checker-guided refinement using solver feedback.
PEARL instead uses multi-turn RL to train the interaction policy itself, including tool invocation, revision, and termination.
A systematic comparison with prior RL-based methods is provided in Table~\ref{tab:agentic_rl_comparison} (Appendix~\ref{app:subsec:rl_comparison}).

For post-hoc verification and repair, OptArgus~\citep{li2026optargus} uses specialist agents to audit problem--model--code consistency, while OPT-Kraken~\citep{wu2026fix} uses source-grounded evidence to constrain repair scope and preserve unaffected model and code components.

\noindent\textbf{Agentic reinforcement learning.}
Agentic reinforcement learning formulates LLMs as policies in temporally extended, partially observable environments whose actions may include reasoning steps and tool calls~\citep{zhang2026the}. Recent systems such as rStar2-Agent~\citep{shang_rstar2-agent_2025}, ReTool~\citep{feng2026retool}, SimpleTIR~\citep{xue2026simpletir}, and Agent-R1~\citep{cheng_agent-r1_2025} show that end-to-end RL can induce tool use, self-correction, and verification behavior with code interpreters, calculators, or search engines. Optimization modeling adds domain-specific structure: solvers expose partially verifiable signals such as feasibility, status, and objective values, while failures may arise from data processing, formulation, implementation, or solver limits. PEARL adapts the Agentic RL paradigm to this setting by making solver interaction, revision, and termination trainable decisions in a multi-turn optimization-modeling environment. This emphasis on defining states, actions, rewards, and termination conditions follows the MDP-design perspective discussed by \citet{yang2026modularized}.

\section{Methodology: Optimization Modeling via Agentic Reinforcement Learning}
\label{sec:method}

\noindent\textbf{Problem Statement.}
We study automated optimization modeling from natural language as a sequential decision-making problem. Given a natural-language problem specification and associated data artifacts, the goal is to produce a mathematically correct optimization formulation together with executable solver code that passes verification. Crucially, correctness cannot be reliably assessed from the generated text alone: {execution, solver feedback, and {reference-based checks provide operational evidence for this goal}}. This setting naturally induces multi-step interaction, partial observability, and delayed rewards.

\noindent\textbf{Optimization Modeling as a POMDP.} We formalize agentic optimization modeling as an episodic partially observable Markov decision process (POMDP) defined by the tuple~\citep{zhang2026the}
$$
\langle \mathcal{S}, \mathcal{O}, \mathcal{A}, \mathcal{P}, \mathcal{R}, T \rangle .
$$
$\mathcal{S}$ denotes the latent environment state, which includes the true optimization problem semantics, hidden modeling errors, solver-internal states, and execution side effects that are not directly observable to the agent. The agent does not observe $s_t \in \mathcal{S}$ directly; instead, it receives observations $o_t \in \mathcal{O}$ consisting of the natural-language
specification, the conversation history, intermediate code artifacts, and tool
feedback such as runtime outputs, solver logs, feasibility status, and validation reports.

The action space $\mathcal{A}$ consists of two classes of actions: (i) natural-language actions that revise or extend the current mathematical
formulation and implementation plan, and (ii) tool-invocation actions that execute code or trigger validation procedures.
Transitions $\mathcal{P}(s_{t+1} \mid s_t, a_t)$ are induced by the effects of code execution, solver behavior, and internal environment updates. {The {scalar} reward {combines interaction-format rewards with terminal objective agreement. Runtime errors}, solver {status}, and {feasibility signals are observations for revision}, not {additional scalar reward terms}.} An episode terminates when the agent emits a final answer (formulation plus solver code) or when a turn or wall-clock budget $T$ is exhausted.

\begin{figure}[!ht]
\centering
\resizebox{0.97\columnwidth}{!}{%
\begin{tikzpicture}[
  node distance=0.65cm,
  observation/.style={draw,rounded corners,align=center,font=\scriptsize,
    minimum width=1.75cm,minimum height=0.62cm,fill=blue!35},
  policy/.style={draw,rounded corners,align=center,font=\scriptsize,
    minimum width=1.75cm,minimum height=0.62cm,fill=red!35},
  action/.style={draw,rounded corners,align=center,font=\scriptsize,
    minimum width=1.85cm,minimum height=0.58cm,fill=violet!25},
  transition/.style={draw,rounded corners,align=center,font=\scriptsize,
    minimum width=1.75cm,minimum height=0.62cm,fill=orange!30},
  rewardbox/.style={draw,dashed,rounded corners,align=center,font=\scriptsize,
    minimum width=2.00cm,minimum height=0.62cm,fill=green!30},
  terminal/.style={draw,rounded corners,align=center,font=\scriptsize,
    minimum width=1.85cm,minimum height=0.58cm,fill=gray!20},
  arrow/.style={->,thick,shorten >=2pt, shorten <=2pt},
  dashedline/.style={dashed,thick,shorten >=2pt, shorten <=2pt}
]
\node[observation] (x) {Problem\\ Instance $x$};
\node[observation, right=of x] (obs) {Observation\\ NL + History\\ + Feedback};
\node[policy, right=of obs] (pi) {Agent Policy\\ $\pi_\theta$};
\node[action, above right=0.08cm and 0.75cm of pi] (nl) {NL Action\\ Draft / Revise};
\node[action, below right=0.08cm and 0.75cm of pi] (tool) {Tool Action\\ {Execute {Code}}};
\node[transition, right=0.80cm of tool] (solver) {{{Python} /}\\ {Solver}};
\node[terminal, right=0.80cm of nl] (out) {Final\\ Answer};
\node[transition, right=of solver] (fb) {Execution\\ Feedback};
\node[rewardbox, anchor=west] (rew) at (fb.west |- out.west) {Reward $R(\tau)$\\
{\scriptsize $r_{\text{fmt}} + r_{\text{corr}}$}};
\draw[arrow] (x) -- (obs);
\draw[arrow] (obs) -- (pi);
\draw[arrow] (pi) -- (nl);
\draw[arrow] (pi) -- (tool);
\draw[arrow] (tool) -- (solver);
\draw[arrow] (solver) -- (fb);
\draw[arrow] (fb.south) -- ++(0,-0.70cm) -| (obs.south);
\draw[dashedline] (out.east) -- (rew.west);

\draw[arrow] (nl) -- (out);
\draw[arrow] (tool.north east) -- (out.south west);

\end{tikzpicture}%
}
\caption{Workflow of \textsc{PEARL}. Colors indicate elements of the POMDP
$\langle \mathcal{S}, \mathcal{O}, \mathcal{A}, \mathcal{P}, \mathcal{R}, T \rangle$:
observations $\mathcal{O}$ (blue), agent policy $\pi_\theta$ (red), actions
$\mathcal{A}$ (purple), environment transitions $\mathcal{P}$ (orange),
reward $\mathcal{R}$ (green, dashed), and termination $T$ (gray).
The latent state $\mathcal{S}$ is not shown. {{Terminal scoring executes the final code and compares its objective with the reference; this comparison is not an in-loop tool observation.}}}
\label{fig:pearl_workflow}
\end{figure}
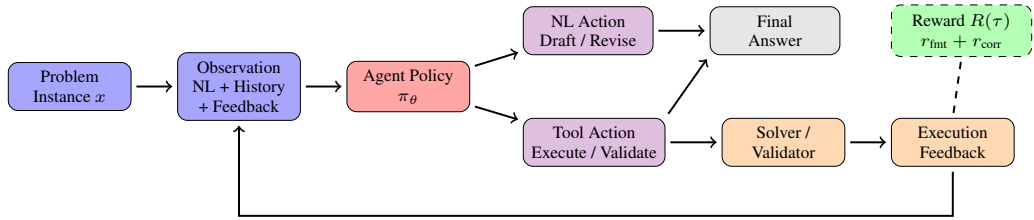

This problem formalization explicitly distinguishes agentic optimization modeling from traditional one-shot generation, which corresponds to a degenerate MDP with horizon $T=1$ and no intermediate observations. We next describe the environment and tool interface (Section~\ref{subsec:env}), action space and output format (Section~\ref{subsec:actions}), reward design (Section~\ref{subsec:reward}), and training objective and policy optimization (Section~\ref{subsec:training}). The overall workflow is demonstrated in Figure~\ref{fig:pearl_workflow}.

\subsection{Environment and Tool Interface}
\label{subsec:env}

At each step, the agent conditions on the full observation history $o_{\le t}$, including the NL specification, intermediate modeling artifacts, and feedback from prior tool interactions. {The environment exposes {a Python execution interface through} \texttt{{<code>...</code>}} {actions}.}

{The {execution interface} runs model-generated Python code {for} data parsing, model construction (e.g., via Pyomo), solver invocation, and solution extraction. {It} returns {execution output and errors inside} \texttt{{<interpreter>...</interpreter>}} {blocks;} solver status, objective values, solutions, and {constraint checks are visible when the executed code reports them. Such checks are performed by model-generated code}, {not by} a {separate trusted} \textit{{Validate}} {action}. {The} agent can {use this feedback to} revise {its implementation or terminate. Reference-objective comparison is performed separately by the terminal scorer}, {which does not append the label or comparison result} to {the interaction history}.}

\subsection{Action Space and Output Format}
\label{subsec:actions}

At each turn $t$, the agent selects an action $a_t \in \mathcal{A}$ that explicitly controls the optimization modeling workflow. The action space consists of two categories: (1) \textit{natural-language actions} modify or extend the current modeling state, including revising decision variables, objectives, constraints, indexing schemes, or implementation plans. These actions correspond to reasoning and editing steps typically performed by human modelers prior to execution; and (2) \textit{tool-invocation actions} trigger interaction with the environment via {the {Python execution interface}}, enabling the agent to test partial formulations, inspect solver behavior, and {{run checks on its candidate solution}}.

The choice between natural-language revision and tool invocation is part of the learned policy, allowing the agent to decide \emph{when} to execute, \emph{what} to test, and \emph{whether} further revision is warranted. An episode terminates when the agent emits a designated terminal action, which outputs a final answer consisting of: (1) a complete mathematical formulation—including decision variables, objective, and constraints—expressed in a standardized template, and (2) executable solver code that produces a solution and an accompanying verification report.
This explicit termination decision enables learning over long-horizon solve--debug--revise trajectories rather than fixed-length generation.
{{Training and inference allow at most eight generation turns, including the final-answer turn. The agent may terminate earlier with an} \texttt{{<answer>...</answer>}} {response. Thus, a trajectory ending with this action can contain at most seven preceding tool calls; exhausting all eight turns on tool calls does not leave an extra final-answer turn.}}

\subsection{Reward Design}
\label{subsec:reward}

We design rewards to be \emph{verifiable}, \emph{solver-grounded}, and aligned with the semantics of optimization modeling.
Let $\tau = (a_{1:T})$ denote a multi-turn interaction trajectory and $R(\tau)$ its return.
Because the correctness of an optimization model cannot be reliably assessed from text alone and typically only becomes evident after execution, reward signals are sparse and delayed.

In practice, we define the return as the sum of two components:
$
R(\tau) = r_{\text{fmt}}(\tau) + r_{\text{corr}}(\tau),
$
which jointly encourage correct interaction protocols and solver-verifiable modeling outcomes.

\noindent\textbf{Format reward.}
{The format reward $r_{\text{fmt}}$ {is computed once per trajectory:} ${+0.1}$ {if the response contains both opening and closing} \texttt{{interpreter}} {tags}, and ${+0.2}$ {if} the {final-answer parser extracts} a {nonempty code block in} the required format{. The} ${0.1}$ {bonus is neither accumulated per call nor conditional on tool success}.}
This component stabilizes training by discouraging invalid interactions that would otherwise prevent execution-based evaluation.

\noindent\textbf{{{Objective-agreement} reward.}}
The {{objective-agreement}} reward $r_{\text{corr}}$ is grounded in solver execution.
After extracting and executing the generated optimization code, we obtain the solver-reported objective value $v_{\text{pred}}$.
{{All retained training instances have} a {verified} reference optimal {objective} value $v_{\text{label}}${. If execution succeeds and an objective value} is {extracted}, the {trajectory earns} ${+1.0}$ {when}}
$
|v_{\text{pred}} - v_{\text{label}}| \le \epsilon,
$
where {$\epsilon{=10^{-3}}$ {is the absolute tolerance for training rewards and online evaluation during training; the final benchmark evaluation protocol is specified separately in Section}~{\ref{sec:exp}}}{.}
{{Otherwise, this reward is zero. Code extraction and execution are prerequisites, not separate rewards, and the scorer does not independently enforce optimal solver termination. Objective agreement} is a {practical proxy}, {not a certificate of semantic correctness; see Appendix}~{\ref{app:limitations_impact}}.}

{By grounding rewards in executable code and solver-verifiable outcomes, this design avoids text-based correctness proxies.
While sparse at the modeling level, the reward provides a clear supervision signal aligned with the end goal of optimization modeling: producing executable models that yield correct solver behavior.
{Further parsing} and {scoring} details are provided in Appendix~\ref{app:subsec:reward_func}.
\subsection{Training Objective and Policy Optimization}
\label{subsec:training}
We train the optimization-modeling agent using agentic reinforcement learning with multi-turn rollouts in a tool-augmented POMDP. The policy is initialized from a pretrained language model and subsequently optimized via GRPO-style reinforcement learning with DAPO-inspired stabilization details such as decoupled clipping~\citep{shao2024deepseekmath,yu_dapo_2025}. This recipe is well suited for sparse, delayed rewards and high-variance tool-augmented trajectories. The learning procedure is summarized in Algorithm~\ref{alg:pearl} (see Appendix~\ref{app:subsec:training_algorithm}).}

\noindent\textbf{Objective.}
We train the agent to maximize expected performance over multi-turn optimization-modeling episodes. Let $x \sim \mathcal{D}$ denote a single optimization modeling instance, consisting of a natural-language problem description and associated data artifacts. Given $x$, the agent induces a multi-turn trajectory $\tau = (o_{1:T}, a_{1:T}) \sim \pi_\theta(\cdot \mid x)$, where each observation $o_t$ encapsulates the conversation history and tool feedback accumulated up to turn $t$, and each action $a_t$ corresponds to either a natural-language modeling step or a tool invocation. The episode terminates after $T$ turns or upon producing a final formulation and solver implementation. Each trajectory $\tau$ receives a scalar return $R(\tau)$ derived from solver execution and validation signals, as described in Section~\ref{subsec:reward}. We optimize the expected trajectory-level return:
\begin{equation}
\max_{\theta}\; J(\theta)
=
\mathbb{E}_{x \sim \mathcal{D}}
\;\mathbb{E}_{\tau \sim \pi_\theta(\cdot \mid x)}
\!\left[R(\tau)\right].
\label{eq:objective}
\end{equation}
where $\pi_\theta$ is the agent policy being optimized. 
This objective encourages improvement in long-horizon, solver-verified performance in a high-variance, tool-augmented environment.

\noindent\textbf{Policy update with decoupled clipping.} We perform on-policy policy optimization using trajectories sampled from the current policy. Let $\pi_{\theta_{\mathrm{old}}}$ denote the policy used to generate rollouts in the current iteration. For each token-level action $a_t$ conditioned on observation $o_t$, we define the standard importance sampling ratio
$
\rho_t(\theta)
=
\frac{\pi_\theta(a_t \mid o_t)}{\pi_{\theta_{\mathrm{old}}}(a_t \mid o_t)} ,
$
where $\pi_\theta$ is the updated policy being optimized.

Given an advantage estimate $A_t$ for action $a_t$, we adopt the decoupled clipping strategy introduced in DAPO \citep{yu_dapo_2025}, which applies asymmetric clipping depending on the sign of the advantage:
\begin{align}
\bar{\rho}_t(\theta)
&=
\begin{cases}
\min(\rho_t(\theta),\,1+\epsilon_{+}), & A_t \ge 0,\\
\max(\rho_t(\theta),\,1-\epsilon_{-}), & A_t < 0,
\end{cases}\\
\mathcal{L}_{\mathrm{pg}}(\theta)
&=
\mathbb{E}
\!\left[
\min\!\left(
\rho_t(\theta)\,A_t,\;
\bar{\rho}_t(\theta)\,A_t
\right)
\right],
\end{align}
where $\epsilon_{+}$ and $\epsilon_{-}$ control the clipping range for positive and negative advantages. This asymmetric design stabilizes learning under high-variance returns in tool-augmented, multi-turn  environments.

{Because rewards in optimization modeling are {computed} at the trajectory level, for each {problem} $x_i$ we sample a {group of} ${G=8}$ {trajectories} ${\tau_{i,j} \sim \pi_{\theta_{\mathrm{old}}}}({\cdot\mid x_i})$. {In the main training configuration,} returns {are normalized} within each {problem's group:}}
{
\begin{align}
{A_{i,j}} &{= \frac{R(\tau_{i,j})-\mu_i}{s_i+\delta},} \qquad
{\mu_i = \frac{1}{G}\sum_{j=1}^{G} R(\tau_{i,j}), \nonumber}\\
{s_i^2} &{= \frac{1}{G-1}\sum_{j=1}^{G}\bigl(R(\tau_{i,j})-\mu_i\bigr)^2,}
\end{align}
}
{{where} ${s_i}$ {is} the {sample} standard deviation {used by the implementation and} ${\delta=10^{-6}}${. The same} ${A_{i,j}}$ {is assigned to the model-generated tokens} of {trajectory} ${\tau_{i,j}}${, with tool-observation tokens masked out. No additional batch-wide advantage normalization} is {enabled}.}

\noindent\textbf{Optimization-specific structure.} Optimization modeling exhibits properties uncommon in generic tool-use tasks. Solver nondeterminism and time limits are treated as environment noise. Problem difficulty varies sharply with context length and specification completeness, motivating a curriculum from short, fully specified instances to long-context and industrial-style inputs. Finally, {{execution errors,} solver {diagnostics,} and {checks performed by the generated code can help distinguish implementation failures from candidate} modeling errors}, enabling feedback-conditioned revision rather than undifferentiated terminal supervision.

\section{Experiments}
\label{sec:exp}

\noindent\textbf{Research questions.} Our experiments aim to answer the following questions: \textbf{RQ1.} Does modeling optimization as an interactive, agentic decision process improve verified solve rates over one-shot generation and heuristic tool-augmented baselines across diverse benchmarks? \textbf{RQ2.} How do key training choices influence performance and stability in agentic optimization modeling?    \textbf{RQ3.} How does agentic training alter the distribution of failure modes compared to non-agentic approaches?

\subsection{Experimental Setup}

\noindent\textbf{Benchmarks and tasks.} We evaluate on a range of established optimization modeling benchmarks, including NL4Opt~\citep{ramamonjison_nl4opt_2023}, MAMO EasyLP and ComplexLP~\citep{huang_mamo_2024}, NLP4LP~\citep{pmlr-v235-ahmaditeshnizi24a}, OptMATH~\citep{lu2025optmath}, IndustryOR~\citep{huang2025orlm}, ComplexOR~\citep{xiao_chain--experts_2024}, and Resocratic~\citep{yang_optibench_2024}. Across all benchmarks, the task is end-to-end optimization modeling: given a natural-language specification and associated data, the model must produce a mathematically correct formulation and executable solver code that yields a feasible (ideally optimal) solution and passes verification.

The suite spans linear, integer, mixed-integer, nonlinear, and second-order cone programs, but it does not exhaust real-world optimization modeling. Stochastic, robust, and highly domain-specific enterprise settings remain under-represented, so our claims should be read as evidence for the current public NL-to-Opt regime rather than universal coverage.

{For PEARL training, we use a mixed corpus of roughly 14k optimization-modeling instances: about 10k cleaned open-source instances drawn from OR-Instruct/ORLM~\citep{huang2025orlm}, OptMATH~\citep{lu2025optmath}, and OptiBench-style data~\citep{yang_optibench_2024}, plus about 3.8k newly constructed and manually curated instances. {Appendix}~{\ref{app:data_collection} describes how the training data were collected and prepared.}

\noindent\textbf{Compared methods.} Our primary system is \textsc{PEARL}-Qwen-3-4B-Instruct-2507. We compare it with baselines spanning different model scales, training regimes, and inference settings. \emph{Direct Prompt} baselines include open-weight and closed-source LLMs evaluated without task-specific training; for models that support different reasoning behaviors, either via distinct variants or inference-time configurations, we report both standard and reasoning-enabled results when available. These include Qwen-3-4B-Instruct/Think, Qwen-3-235B-A22B-Instruct/Think, DeepSeek-V3.2-685B/Think, and the GPT-5-family endpoints used during our method development and evaluation, GPT-5 and GPT-5-CodeX. \emph{Fine-Tuned} baselines are prior optimization-modeling systems adapted via supervised fine-tuning and/or reinforcement learning, including SIRL-Qwen2.5-7B/32B, ORLM-LLaMA-3-8B, and OptMATH-Qwen2.5-7B/32B. {For all methods in Table}~{\ref{tab:main_results_final}, we execute their final code and check its objective value against the reference. The baselines generate code in one response, while PEARL learns to revise its code using solver feedback. We also compare PEARL with the untrained Qwen3-4B-Instruct-2507 model using the same tools and turn limit for both (Appendix}~{\ref{app:subsec:tool_use_depth}; Table}~{\ref{tab:scaffold_control_tool_calls})}.}

\noindent\textbf{Evaluation metrics.} Evaluation is performed at both the instance and dataset levels. At the instance level, {a generated {program passes the objective-agreement criterion} if {it executes successfully and} its {output satisfies the normalized objective-error threshold}}, i.e., $\frac{|y_{\text{pred}} - y_{\text{label}}|}{|y_{\text{label}}| + 1} < 10^{-6}$. At the dataset level, we report pass@1 accuracy, defined as the fraction of instances for which a single model-generated solution satisfies this criterion. {{Each evaluation run produces} one {final candidate} per instance.} {By} default, we run each pass@1 evaluation three times and report mean $\pm$ standard deviation. Unless otherwise specified, inference is conducted with a sampling temperature of $0.6$. Additional evaluation details and parameter settings are provided in Appendix~\ref{app:experiments_setup}. {Code and datasets are provided at {\url{https://github.com/AuroraLHL/PEARL}.}}

{{For the main Qwen3-4B experiment, we train on four high-performance GPUs with a learning rate of} ${10^{-6}}${. Each rollout batch contains 16 problems, with eight interaction trajectories sampled for each problem. Full hyperparameters are provided in Table}~{\ref{tab:qwen3_4b_PEARL_hparams} (Appendix}~{\ref{app:design_details})}.}

\subsection{Experimental Results and Analysis}
\label{sec:results}

The main text answers RQ1 by measuring overall effectiveness against prompting and fine-tuned baselines (Sec.~\ref{sec:results:rq1}; setup details, baseline-selection caveats, and extended comparisons in Appendix~\ref{app:experiments_setup} and Appendix~\ref{app:subsec:8b_comparison}), RQ2 by analyzing training choices for stable agentic optimization modeling (Sec.~\ref{sec:results:rq2}; ablations and training dynamics in Appendix~\ref{app:ablation_sensitivity}), and RQ3 by attributing how failures change after PEARL training (Sec.~\ref{sec:results:rq3}; diagnostic definitions in Appendix~\ref{app:failure_mode_attribution}).
\subsubsection{RQ1: Overall Effectiveness of Agentic Optimization Modeling}
\label{sec:results:rq1}

Table~\ref{tab:main_results_final} answers RQ1 by comparing PEARL with direct prompting and prior task-specific fine-tuning baselines across all benchmarks. We draw the following takeaways.

\noindent\textbf{Takeaway 1: {{PEARL substantially improves performance at a fixed model} scale.}} PEARL substantially improves over same-scale direct prompting baselines: the 4B PEARL model reaches 69.71 macro average, compared with 22.20 for Qwen-3-4B-Instruct and 30.86 for Qwen-3-4B-Think. It also achieves the {highest} macro average among fine-tuned models {in Table}~{\ref{tab:main_results_final} under our Pyomo evaluation protocol. PEARL also has a higher macro average than its base model with the same tools and turn budget (Appendix}~{\ref{app:subsec:tool_use_depth}; Table}~{\ref{tab:scaffold_control_tool_calls}), and than SIRL with the same backbone and training data (Appendix}~{\ref{app:sirl_retraining}; Table}~{\ref{tab:sirl_controls}). Thus, tool access or extra data alone cannot explain the gain.}

\noindent\textbf{Takeaway 2: {{large} gains {appear on harder} modeling {benchmarks}}.} On simpler datasets such as NL4Opt and MAMO EasyLP, improvements are modest because many strong models already achieve high solve rates. In contrast, on harder benchmarks such as MAMO ComplexLP and Resocratic, PEARL achieves large absolute gains, suggesting improved robustness to {{complex modeling} and {implementation demands}}. This pattern {{is consistent with the benefits of learning} solve--debug--revise behavior}.

\providecommand{\accpm}[2]{#1{\scriptsize$\pm$#2}}

\begin{table*}[h]
 
    \centering 
    \caption{Pass@1 accuracy (\%) across benchmarks (higher is better). Each entry reports the mean $\pm$ standard deviation over three runs.} 
    \label{tab:main_results_final} 
    \resizebox{\linewidth}{!}{% 
    \begin{tabular}{l lcccccccccc} 
    \toprule 
    & \textbf{Method} 
    & \textbf{NL4Opt} 
    & \textbf{MAMO-E} 
    & \textbf{MAMO-C} 
    & \textbf{NLP4LP} 
    & \textbf{OptMATH} 
    & \textbf{IndustryOR} 
    & \textbf{ComplexOR} 
    & \textbf{Resocratic} 
    & Macro Avg 
    & Micro Avg\\ 
    \midrule 

    \multirow{8}{*}{\rotatebox{90}{\textbf{Direct Prompt}}} 
    & Qwen-3-4B-Instruct
     & \accpm{48.13}{1.40} & \accpm{2.20}{0.45} & \accpm{13.51}{2.20} & \accpm{55.62}{1.25} & \accpm{2.41}{0.50} & \accpm{16.66}{3.10} & \accpm{5.56}{2.78} & \accpm{33.50}{1.80} & \accpm{22.20}{0.75} & \accpm{22.42}{0.55} \\ 

    & Qwen-3-4B-Think
     & \accpm{48.60}{2.80} & \accpm{27.34}{2.50} & \accpm{6.31}{1.90} & \accpm{59.55}{2.10} & \accpm{1.81}{0.45} & \accpm{30.95}{4.10} & \accpm{22.22}{5.56} & \accpm{50.12}{2.50} & \accpm{30.86}{0.95} & \accpm{35.06}{0.80} \\ 

    & Qwen3-235B-A22B-Instruct-2507
     & \accpm{74.92}{1.77} & \accpm{91.62}{0.69} & \accpm{38.74}{5.41} & \accpm{88.95}{0.33} & \accpm{20.88}{3.09} & \accpm{58.73}{5.50} & \accpm{38.89}{0.00} & \accpm{70.64}{1.41} & \accpm{60.42}{1.13} & \accpm{72.26}{0.69} \\ 

    & Qwen3-235B-A22B-Think-2507
     & \accpm{79.59}{5.42} & \accpm{89.79}{3.07} & \accpm{40.84}{6.64} & \accpm{91.20}{5.54} & \accpm{30.72}{1.59} & \accpm{71.43}{4.12} & \accpm{50.00}{5.56} & \accpm{74.36}{4.50} & \accpm{65.99}{1.68} & \accpm{74.95}{1.39} \\

    & DeepSeek-V3.2-685B
     & \accpm{83.18}{3.07} & \accpm{94.07}{2.29} & \accpm{50.75}{9.38} & \accpm{92.70}{2.81} & \accpm{25.10}{3.03} & \accpm{58.73}{11.98} & \accpm{51.85}{6.42} & \accpm{75.68}{2.52} & \accpm{66.51}{1.88} & \accpm{77.07}{1.22} \\ 
    & DeepSeek-V3.2-Think-685B
     & \accpm{85.20}{1.18} & \accpm{95.53}{0.59} & \accpm{53.15}{3.12} & \accpm{97.38}{0.86} & \accpm{30.32}{1.25} & \accpm{64.29}{2.39} & \accpm{50.00}{5.56} & \accpm{77.67}{0.43} & \accpm{69.19}{0.65} & \accpm{79.58}{0.41} \\ 

     & GPT-5 & \accpm{83.80}{0.97} & \accpm{94.31}{0.32} & \accpm{59.16}{0.52} & \accpm{94.76}{0.65} & \accpm{31.33}{0.61} & \accpm{73.02}{4.95} & \accpm{53.70}{8.49} & \accpm{81.14}{1.29} & \textbf{\accpm{71.40}{1.55}} & \textbf{\accpm{80.31}{0.41}} \\ 
     & GPT-5-Codex & \accpm{80.37}{0.94} & \accpm{95.60}{0.19} & \accpm{61.56}{0.52} & \accpm{97.94}{0.32} & \accpm{32.13}{0.69} & \accpm{69.05}{2.38} & \accpm{55.56}{5.56} & \accpm{78.99}{0.38} & \textbf{\accpm{71.40}{0.37}} & \accpm{80.27}{0.15} \\

    \midrule 
    \multirow{6}{*}{\rotatebox{90}{\textbf{Fine-Tuned}}} 

    & \cellcolor{gray!0}SIRL-Qwen2.5-7B 
     & \cellcolor{gray!0}\accpm{76.17}{1.45} & \cellcolor{gray!0}\accpm{82.20}{1.20} & \cellcolor{gray!0}\accpm{2.70}{1.35} & \cellcolor{gray!0}\accpm{93.26}{0.75} & \cellcolor{gray!0}\accpm{6.02}{0.80} & \cellcolor{gray!0}\accpm{21.43}{3.60} & \cellcolor{gray!0}\accpm{11.11}{3.92} & \cellcolor{gray!0}\accpm{63.03}{1.60} & \accpm{36.81}{0.90} & \accpm{55.34}{0.70} \\ 

    & \cellcolor{gray!0}SIRL-Qwen2.5-32B 
     & \cellcolor{gray!0}\accpm{87.38}{1.10} & \cellcolor{gray!0}\accpm{98.90}{0.45} & \cellcolor{gray!0}\accpm{70.27}{3.20} & \cellcolor{gray!0}\accpm{97.75}{0.50} & \cellcolor{gray!0}\accpm{22.89}{1.40} & \cellcolor{gray!0}\accpm{59.52}{4.10} & \cellcolor{gray!0}\accpm{33.33}{5.56} & \cellcolor{gray!0}\accpm{78.16}{1.10} & \accpm{60.07}{0.75} & \accpm{73.65}{0.60} \\ 
    & ORLM-LLaMA-3-8B 
   & \accpm{78.50}{1.80}
   & \accpm{82.30}{1.60} 
   & \accpm{41.44}{3.80} 
   & \accpm{89.54}{1.20}
   & \accpm{6.02}{0.90} 
   & \accpm{42.86}{4.70}
   & \accpm{38.89}{5.56}
   & \accpm{70.22}{1.50}
   & \accpm{56.22}{0.95}
   & \accpm{67.97}{0.75}\\
    & OptMATH-Qwen2.5-7B 
   & \accpm{84.11}{1.30} 
   & \accpm{77.61}{1.80} 
   & \accpm{33.33}{3.40} 
   & \accpm{89.47}{1.30}
   & \accpm{22.29}{1.50} 
   & \accpm{30.95}{4.10}
   & \accpm{33.33}{5.56}
   & \accpm{91.32}{0.80}
   & \accpm{57.80}{0.85} 
   & \accpm{72.94}{0.65} 
   \\
    & OptMATH-Qwen2.5-32B 
   & \accpm{85.51}{1.05}
   & \accpm{95.05}{0.80}
   & \accpm{44.14}{3.60}
   & \accpm{95.51}{0.65}
   & \accpm{37.35}{1.70}
   & \accpm{45.24}{4.60}
   & \accpm{50.00}{5.56} 
   & \accpm{83.37}{0.95}
   & \accpm{67.02}{0.70} 
   & \accpm{80.26}{0.45} 
   \\
       & \textbf{PEARL-Qwen-3-4B-Instruct (Ours)}
     & \accpm{85.28}{0.99} & \accpm{91.65}{0.65} & \accpm{76.12}{3.19} & \accpm{93.25}{1.59} & \accpm{16.87}{1.70} & \accpm{67.86}{8.42} & \accpm{41.66}{3.92} & \accpm{84.99}{0.52} & \accpm{69.71}{0.13} & \accpm{79.84}{0.25} \\

    \bottomrule 
    \end{tabular}% 
    }% 
    \end{table*}

\noindent\textbf{Takeaway 3: the tool scaffold alone is insufficient.} Appendix~\ref{app:subsec:tool_use_depth} reports a controlled scaffold comparison in which the base Qwen3-4B-Instruct-2507 model uses the same tool interface, execution-feedback loop, and repair budget as PEARL. The base model can invoke tools, but it uses them more sparsely and achieves lower accuracy across all benchmarks, {so providing tools without training does not reproduce PEARL's performance. The untrained model achieves 36.32\% macro-averaged Pass@1, compared with 69.71\% for PEARL (Appendix}~{\ref{app:subsec:tool_use_depth}; Table}~{\ref{tab:scaffold_control_tool_calls}). The training gains also extend to another backbone: with Qwen3-8B, PEARL improves over the untrained model, reaching 67.73\% macro-averaged and 79.61\% micro-averaged Pass@1 (Appendix}~{\ref{app:subsec:8b_comparison}; Table}~{\ref{tab:appendix_8b_results})}.

\noindent\textbf{Takeaway 4: PEARL-4B is competitive with much larger general-purpose models, but not a universal winner.} We compare \textsc{PEARL} against substantially larger models, including Qwen-3-235B-A22B, DeepSeek-V3.2-685B, DeepSeek-V3.2-Think-685B, GPT-5, and GPT-5-CodeX. Despite being orders of magnitude smaller, \textsc{PEARL}-4B outperforms DeepSeek-V3.2-685B in both macro-averaged accuracy (69.71 vs. 66.51) and micro-averaged accuracy (79.84 vs. 77.07), with particularly strong results on MAMO ComplexLP and Resocratic. At the same time, GPT-5-family models and DeepSeek-V3.2-Think remain highly competitive or stronger on some aggregate metrics. The result is therefore not that small models always dominate, but that optimization-specific agentic training can close much of the scale gap for solver-grounded modeling tasks.

\noindent\textbf{Takeaway 5: compact backbones make repeated agentic inference more practical.} Agentic optimization modeling amplifies inference cost because each problem may require several policy calls interleaved with code execution and solver feedback, making serving memory and throughput constraints increasingly salient~\citep{kwon2023efficient}. Recent reports show that 3--4B models can be capable enough for local deployment~\citep{abdin2024phi3}, 2B on-device models can support function-calling agents with strong latency advantages~\citep{chen2024octopusv2}, and specialized repetitive agentic routines are natural targets for small language models~\citep{belcak2025small}. We therefore view \textsc{PEARL}-4B as evidence that optimization-specific agentic training can move solver-grounded modeling into a more deployable parameter regime, while not making a universal wall-clock or dollar-cost claim across different serving stacks.
{{PEARL averages 9,868.75 generated tokens per instance, compared with 1,465.51 for DeepSeek-V3.2 and 2,323.56 for DeepSeek-V3.2-Think (Appendix}~{\ref{app:data_token_usage}; Table}~{\ref{tab:token_cost}). Token count alone, however, is not a measure of serving cost. PEARL's 4B backbone supports local deployment with lower model-memory requirements and potentially lower per-token serving costs}~{\citep{abdin2024phi3,belcak2025small}. These advantages can help offset the extra generation needed for multi-turn modeling, especially for repeated local use.}}

\subsubsection{RQ2: Impact of Training Choices}
\label{sec:results:rq2}

\begin{figure}[h]
    \centering
    \includegraphics[width=1.0\linewidth]{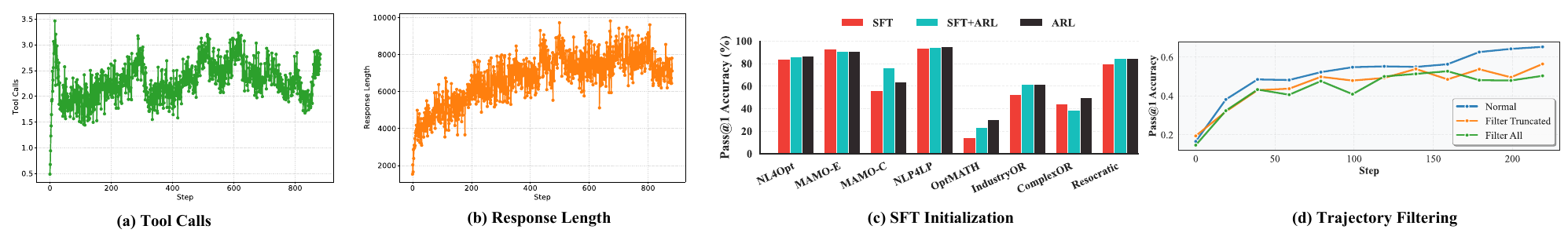}
    \caption{Main ablations for PEARL training choices. Tool use converges early, response length continues to grow under GRPO, SFT mainly serves as a warm-start for less aligned models, and aggressive trajectory filtering is not beneficial in this solver-feedback setting.}
    \label{fig:rq2_training_choices}
\end{figure}

We study training dynamics and design choices that affect the stability and efficiency of agentic optimization modeling; Figure~\ref{fig:rq2_training_choices} summarizes the main evidence, with additional details in Appendix~\ref{app:subsec:role_of_SFT}, Appendix~\ref{app:subsec:results_trajectory_filtering}, and Appendix~\ref{app:subsec:training_dynamics}. We make four observations: (1) \textbf{Tool-use behavior is learned early.} Figure~\ref{fig:rq2_training_choices}(a) shows that the average number of tool calls stabilizes early in training, suggesting that PEARL quickly learns when external execution and solver feedback are useful. (2) \textbf{Later training mainly refines solution traces.} Figure~\ref{fig:rq2_training_choices}(b) shows that response length continues to grow after tool usage has stabilized, indicating that GRPO training increasingly improves the detail and structure of the generated modeling trajectory rather than simply increasing tool frequency. (3) \textbf{SFT is a warm-start, not the final driver.} Figure~\ref{fig:rq2_training_choices}(c) indicates that supervised fine-tuning is important for less aligned starting models, but direct agentic RL reaches comparable performance to SFT+ARL when starting from Qwen3-4B-Instruct; thus, SFT helps initialize the policy, while solver-grounded agentic RL determines the final interaction behavior. (4) \textbf{Aggressive trajectory filtering is unnecessary in this solver-feedback setting.} Figure~\ref{fig:rq2_training_choices}(d) shows that, unlike some general tool-use domains~\citep{xue2026simpletir,shang_rstar2-agent_2025}, {filtering truncated or format-violating trajectories does not improve performance {in this setting} and can slightly {reduce} it{. The limited} tool{-}use {depth and capped} observations {may contribute to the observed stability}.}
\subsubsection{RQ3: Failure Mode Analysis}
\label{sec:results:rq3}

\begin{figure}[h]
    \centering
    \includegraphics[width=0.60\linewidth]{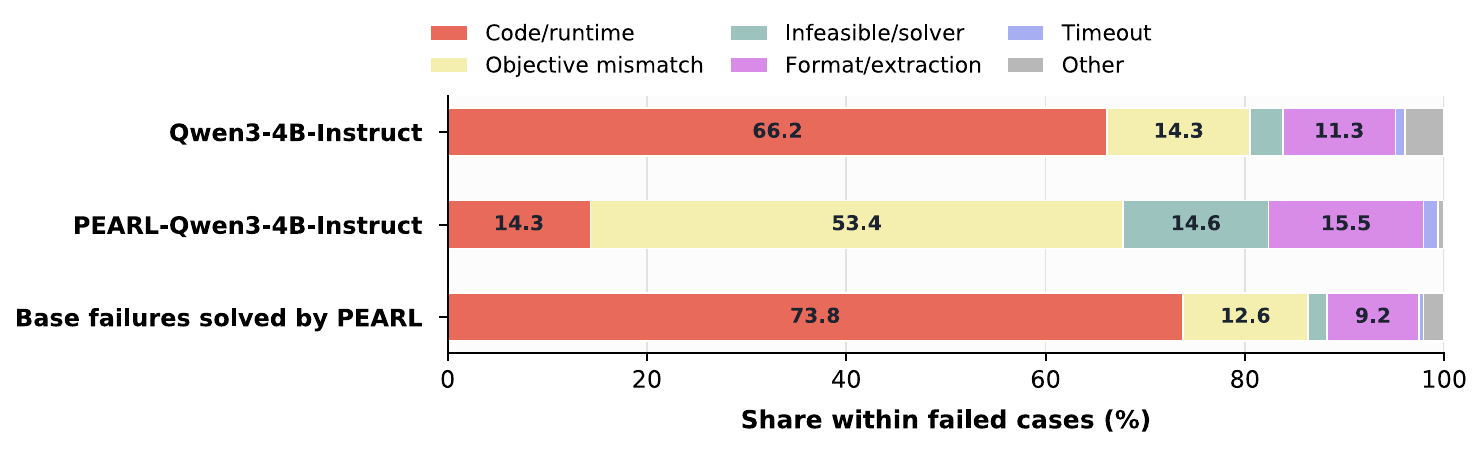}
    \caption{Failure attribution before and after PEARL training under the main pass@1 evaluation setting. Bars are normalized within failed cases; attribution setup, category definitions, and detailed interpretation are provided in Appendix~\ref{app:failure_mode_attribution}.}
    \label{fig:failure_mode_attribution}
\end{figure}

To understand how agentic training changes model behavior, we compare the failure attribution of the base model and PEARL under the main pass@1 evaluation setting. We make three observations: (1) \textbf{PEARL shifts failures away from execution and parsing.} Compared with the base model, PEARL fixes many {code execution errors} and {formatting} or extraction failures, indicating improved ability to emit executable solver programs, follow the required output protocol, and recover from tool feedback. (2) \textbf{The corrected base-model failures are mostly repairable implementation failures.} Among base-model failures that PEARL later solves, most are {code execution} failures, suggesting that the learned policy mainly teaches the model to execute, inspect, and repair solver programs rather than merely produce longer answers. (3) \textbf{The remaining failures are more modeling-centric and expose the limits of solver feedback.} After PEARL removes many basic execution failures, a larger share of the remaining errors comes from objective mismatch and solver-side issues. These residual cases are more likely to reflect formulation quality, ambiguous specifications, or optimization difficulty. Solver and compiler messages are useful but incomplete evidence for structural correctness, so PEARL combines execution feedback with deterministic validation and, when available, objective-value agreement against references; nevertheless, these checks remain partial oracles.

\subsubsection{{Additional Comparisons and Data Ablation}}
\label{sec:results:additional}

\noindent\textbf{{Training on the same problems.}} {Using the same Qwen2.5-7B-Instruct backbone, we retrain SIRL on PEARL's 13.8k training problems. Both use Pyomo and the same eight benchmarks, evaluation prompt, and objective check. PEARL reaches 67.86\% macro and 77.82\% micro Pass@1, versus 57.22\% and 73.03\% for SIRL (Appendix}~{\ref{app:sirl_retraining}; Table}~{\ref{tab:sirl_controls})}.

\noindent\textbf{{Testing SIRL with Gurobi.}} {The released SIRL-Qwen2.5-7B checkpoint, evaluated with its native Gurobi API and our evaluation-prompt template, achieves 66.84\% macro and 82.29\% micro Pass@1 (Appendix}~{\ref{app:sirl_retraining}; Table}~{\ref{tab:sirl_controls}). It is competitive with PEARL in macro accuracy and stronger in micro accuracy, but uses different training data.}

\noindent\textbf{{Contribution of the curated data.}} {In a separate Qwen3-4B experiment, we keep the training recipe unchanged and add 3.8k curated problems to the public 10k pool. Macro Pass@1 rises from 64.12\% to 70.40\% and micro Pass@1 from 78.95\% to 80.38\% (Appendix}~{\ref{app:data_token_usage}; Table}~{\ref{tab:data_ablation})}.

{\section{Conclusion}
\label{sec:conclusion}
We presented \textsc{PEARL}, an agentic RL framework that treats optimization modeling from natural language as {a} multi-turn {solver-interaction} process. By integrating solver execution and feedback into {rollouts}, \textsc{PEARL} learns solver-aware behaviors such as when to test, how to revise formulations, and when to terminate. \textsc{PEARL} substantially improves over {the same-scale base model} across {the evaluated} benchmarks. Notably, compact \textsc{PEARL} models {are competitive with the} much larger general-purpose {baselines evaluated in this study}, {showing} that learned interaction and feedback utilization can {complement} model scale in {optimization modeling}. Our current evidence is still limited by bounded interaction budgets, benchmark coverage, and the partial nature of solver-based validation. Future work should extend agentic optimization modeling to stochastic, robust, and domain-specific settings with richer validators and more adaptive interaction policies. Limitations and broader impact are discussed in Appendix~\ref{app:limitations_impact}{.}}

{
\begin{ack}
The computational resources were supported by the Center for Intelligent Computing and Song-Shan Lake HPC Center (SSL-HPC) in Great Bay University, Dongguan, China. This work was supported in part by National Key Research and Development Program of China under the grant number 2024YFA1012900, and the National Natural Science Foundation of China under the grant numbers 12331010 and 12288101. We also thank the anonymous reviewers for their valuable feedback.

\noindent\textbf{Competing interests.} The authors declare no competing interests.
\end{ack}
}

\bibliographystyle{plainnat}
\bibliography{ref}

\begin{thebibliography}{44}
\providecommand{\natexlab}[1]{#1}
\providecommand{\url}[1]{\texttt{#1}}
\expandafter\ifx\csname urlstyle\endcsname\relax
  \providecommand{\doi}[1]{doi: #1}\else
  \providecommand{\doi}{doi: \begingroup \urlstyle{rm}\Url}\fi

\bibitem[Abdin et~al.(2024)Abdin, Aneja, Awadalla, Awadallah, Awan, Bach, Bahree, Bakhtiari, Bao, Behl, Benhaim, Bilenko, Bjorck, Bubeck, Cai, Cai, Chaudhary, Chen, Chen, Chen, Chen, Chen, Cheng, Chopra, Dai, Dixon, Eldan, Fragoso, Gao, Gao, Gao, Garg, Giorno, Goswami, Gunasekar, Haider, Hao, Hewett, Hu, Huynh, Iter, Jacobs, Javaheripi, Jin, Karampatziakis, Kauffmann, Khademi, Kim, Kim, Kurilenko, Lee, Lee, Li, Li, Liang, Liden, Lin, Lin, Liu, Liu, Liu, Liu, Liu, Luo, Madan, Mahmoudzadeh, Majercak, Mazzola, Mendes, Mitra, Modi, Nguyen, Norick, Patra, Perez-Becker, Portet, Pryzant, Qin, Radmilac, Ren, de~Rosa, Rosset, Roy, Ruwase, Saarikivi, Saied, Salim, Santacroce, Shah, Shang, Sharma, Shen, Shukla, Song, Tanaka, Tupini, Vaddamanu, Wang, Wang, Wang, Wang, Wang, Wang, Ward, Wen, Witte, Wu, Wu, Wyatt, Xiao, Xu, Xu, Xu, Xue, Yadav, Yang, Yang, Yang, Yang, Yu, Yuan, Zhang, Zhang, Zhang, Zhang, Zhang, Zhang, Zhang, and Zhou]{abdin2024phi3}
Marah Abdin, Jyoti Aneja, Hany Awadalla, Ahmed Awadallah, Ammar~Ahmad Awan, Nguyen Bach, Amit Bahree, Arash Bakhtiari, Jianmin Bao, Harkirat Behl, Alon Benhaim, Misha Bilenko, Johan Bjorck, Sébastien Bubeck, Martin Cai, Qin Cai, Vishrav Chaudhary, Dong Chen, Dongdong Chen, Weizhu Chen, Yen-Chun Chen, Yi-Ling Chen, Hao Cheng, Parul Chopra, Xiyang Dai, Matthew Dixon, Ronen Eldan, Victor Fragoso, Jianfeng Gao, Mei Gao, Min Gao, Amit Garg, Allie~Del Giorno, Abhishek Goswami, Suriya Gunasekar, Emman Haider, Junheng Hao, Russell~J. Hewett, Wenxiang Hu, Jamie Huynh, Dan Iter, Sam~Ade Jacobs, Mojan Javaheripi, Xin Jin, Nikos Karampatziakis, Piero Kauffmann, Mahoud Khademi, Dongwoo Kim, Young~Jin Kim, Lev Kurilenko, James~R. Lee, Yin~Tat Lee, Yuanzhi Li, Yunsheng Li, Chen Liang, Lars Liden, Xihui Lin, Zeqi Lin, Ce~Liu, Liyuan Liu, Mengchen Liu, Weishung Liu, Xiaodong Liu, Chong Luo, Piyush Madan, Ali Mahmoudzadeh, David Majercak, Matt Mazzola, Caio César~Teodoro Mendes, Arindam Mitra, Hardik Modi, Anh Nguyen, Brandon Norick, Barun Patra, Daniel Perez-Becker, Thomas Portet, Reid Pryzant, Heyang Qin, Marko Radmilac, Liliang Ren, Gustavo de~Rosa, Corby Rosset, Sambudha Roy, Olatunji Ruwase, Olli Saarikivi, Amin Saied, Adil Salim, Michael Santacroce, Shital Shah, Ning Shang, Hiteshi Sharma, Yelong Shen, Swadheen Shukla, Xia Song, Masahiro Tanaka, Andrea Tupini, Praneetha Vaddamanu, Chunyu Wang, Guanhua Wang, Lijuan Wang, Shuohang Wang, Xin Wang, Yu~Wang, Rachel Ward, Wen Wen, Philipp Witte, Haiping Wu, Xiaoxia Wu, Michael Wyatt, Bin Xiao, Can Xu, Jiahang Xu, Weijian Xu, Jilong Xue, Sonali Yadav, Fan Yang, Jianwei Yang, Yifan Yang, Ziyi Yang, Donghan Yu, Lu~Yuan, Chenruidong Zhang, Cyril Zhang, Jianwen Zhang, Li~Lyna Zhang, Yi~Zhang, Yue Zhang, Yunan Zhang, and Xiren Zhou.
\newblock Phi-3 technical report: A highly capable language model locally on your phone, 2024.
\newblock URL \url{https://arxiv.org/abs/2404.14219}.

\bibitem[Ahmaditeshnizi et~al.(2024)Ahmaditeshnizi, Gao, and Udell]{pmlr-v235-ahmaditeshnizi24a}
Ali Ahmaditeshnizi, Wenzhi Gao, and Madeleine Udell.
\newblock {O}pti{MUS}: Scalable optimization modeling with ({MI}){LP} solvers and large language models.
\newblock In Ruslan Salakhutdinov, Zico Kolter, Katherine Heller, Adrian Weller, Nuria Oliver, Jonathan Scarlett, and Felix Berkenkamp, editors, \emph{Proceedings of the 41st International Conference on Machine Learning}, volume 235 of \emph{Proceedings of Machine Learning Research}, pages 577--596. PMLR, 21--27 Jul 2024.
\newblock URL \url{https://proceedings.mlr.press/v235/ahmaditeshnizi24a.html}.

\bibitem[Anand et~al.(2017)Anand, Aggarwal, and Kumar]{anand2017comparative}
Rimmi Anand, Divya Aggarwal, and Vijay Kumar.
\newblock A comparative analysis of optimization solvers.
\newblock \emph{Journal of Statistics and Management Systems}, 20\penalty0 (4):\penalty0 623--635, 2017.

\bibitem[Antoniou and Lu(2007)]{antoniou2007practical}
Andreas Antoniou and Wu-Sheng Lu.
\newblock \emph{Practical optimization: algorithms and engineering applications}.
\newblock Springer, 2007.

\bibitem[Astorga et~al.(2025)Astorga, Liu, Xiao, and van~der Schaar]{astorga2025autoformulation}
Nicol{\'a}s Astorga, Tennison Liu, Yuanzhang Xiao, and Mihaela van~der Schaar.
\newblock Autoformulation of mathematical optimization models using {LLM}s.
\newblock In \emph{Forty-second International Conference on Machine Learning}, 2025.
\newblock URL \url{https://openreview.net/forum?id=33YrT1j0O0}.

\bibitem[Belcak et~al.(2025)Belcak, Heinrich, Diao, Fu, Dong, Muralidharan, Lin, and Molchanov]{belcak2025small}
Peter Belcak, Greg Heinrich, Shizhe Diao, Yonggan Fu, Xin Dong, Saurav Muralidharan, Yingyan~Celine Lin, and Pavlo Molchanov.
\newblock Small language models are the future of agentic ai.
\newblock \emph{arXiv preprint arXiv:2506.02153}, 2025.

\bibitem[Bynum et~al.(2021)Bynum, Hackebeil, Hart, Laird, Nicholson, Siirola, Watson, Woodruff, et~al.]{bynum2021pyomo}
Michael~L Bynum, Gabriel~A Hackebeil, William~E Hart, Carl~D Laird, Bethany~L Nicholson, John~D Siirola, Jean-Paul Watson, David~L Woodruff, et~al.
\newblock \emph{Pyomo-optimization modeling in python}, volume~67 of \emph{Springer Optimization and Its Applications}.
\newblock Springer, 2021.

\bibitem[Chen and Li(2024)]{chen2024octopusv2}
Wei Chen and Zhiyuan Li.
\newblock Octopus v2: On-device language model for super agent, 2024.
\newblock URL \url{https://arxiv.org/abs/2404.01744}.

\bibitem[Chen et~al.(2025{\natexlab{a}})Chen, Xia, Shao, Ge, and Ye]{chen2025solverinformed}
Yitian Chen, Jingfan Xia, Siyu Shao, Dongdong Ge, and Yinyu Ye.
\newblock Solver-informed {RL}: Grounding large language models for authentic optimization modeling.
\newblock In \emph{The Thirty-ninth Annual Conference on Neural Information Processing Systems}, 2025{\natexlab{a}}.
\newblock URL \url{https://openreview.net/forum?id=80L235oVBe}.

\bibitem[Chen et~al.(2025{\natexlab{b}})Chen, Zhang, Zope, Barbalho, Mellou, Molinaro, Kulkarni, Menache, and Li]{chen_optimind_2025}
Zeyi Chen, Xinzhi Zhang, Humishka Zope, Hugo Barbalho, Konstantina Mellou, Marco Molinaro, Janardhan Kulkarni, Ishai Menache, and Sirui Li.
\newblock {OptiMind}: {Teaching} {LLMs} to {Think} {Like} {Optimization} {Experts}, September 2025{\natexlab{b}}.
\newblock URL \url{http://arxiv.org/abs/2509.22979}.
\newblock arXiv:2509.22979 [cs].

\bibitem[Cheng et~al.(2025)Cheng, Ouyang, Yu, Yan, Luo, Liu, Wang, Liu, and Chen]{cheng_agent-r1_2025}
Mingyue Cheng, Jie Ouyang, Shuo Yu, Ruiran Yan, Yucong Luo, Zirui Liu, Daoyu Wang, Qi~Liu, and Enhong Chen.
\newblock Agent-{R1}: {Training} {Powerful} {LLM} {Agents} with {End}-to-{End} {Reinforcement} {Learning}, November 2025.
\newblock URL \url{http://arxiv.org/abs/2511.14460}.
\newblock arXiv:2511.14460 [cs].

\bibitem[Ding et~al.(2026)Ding, Tan, Zhang, and Chen]{ding2026or}
Zezhen Ding, Zhen Tan, Jiheng Zhang, and Tianlong Chen.
\newblock Or-r1: Automating modeling and solving of operations research optimization problem via test-time reinforcement learning.
\newblock In \emph{Proceedings of the AAAI Conference on Artificial Intelligence}, volume~40, pages 228--236, 2026.

\bibitem[Feng et~al.(2026)Feng, Huang, Qu, Zhang, Qin, Zhong, Jiang, Chi, and Zhong]{feng2026retool}
Jiazhan Feng, Shijue Huang, Xingwei Qu, Ge~Zhang, Yujia Qin, Baoquan Zhong, Chengquan Jiang, Jinxin Chi, and Wanjun Zhong.
\newblock Retool: Reinforcement learning for strategic tool use in {LLM}s.
\newblock In \emph{The Fourteenth International Conference on Learning Representations}, 2026.
\newblock URL \url{https://openreview.net/forum?id=tRk1nofSmz}.

\bibitem[Huang et~al.(2025)Huang, Tang, Hu, Jiang, Zheng, Ge, Wang, and Wang]{huang2025orlm}
Chenyu Huang, Zhengyang Tang, Shixi Hu, Ruoqing Jiang, Xin Zheng, Dongdong Ge, Benyou Wang, and Zizhuo Wang.
\newblock Orlm: A customizable framework in training large models for automated optimization modeling.
\newblock \emph{Operations Research}, 2025.

\bibitem[Huang et~al.(2024)Huang, Shen, Hu, Gao, and Wang]{huang_mamo_2024}
Xuhan Huang, Qingning Shen, Yan Hu, Anningzhe Gao, and Benyou Wang.
\newblock Mamo: a {Mathematical} {Modeling} {Benchmark} with {Solvers}, 2024.
\newblock URL \url{https://arxiv.org/abs/2405.13144}.
\newblock Publisher: arXiv Version Number: 2.

\bibitem[Jiang et~al.(2025)Jiang, Shu, Qian, Lu, ZHOU, Zhou, and Yu]{jiang2025llmopt}
Caigao Jiang, Xiang Shu, Hong Qian, Xingyu Lu, JUN ZHOU, Aimin Zhou, and Yang Yu.
\newblock {LLMOPT}: Learning to define and solve general optimization problems from scratch.
\newblock In \emph{The Thirteenth International Conference on Learning Representations}, 2025.
\newblock URL \url{https://openreview.net/forum?id=9OMvtboTJg}.

\bibitem[Kwon et~al.(2023)Kwon, Li, Zhuang, Sheng, Zheng, Yu, Gonzalez, Zhang, and Stoica]{kwon2023efficient}
Woosuk Kwon, Zhuohan Li, Siyuan Zhuang, Ying Sheng, Lianmin Zheng, Cody~Hao Yu, Joseph Gonzalez, Hao Zhang, and Ion Stoica.
\newblock Efficient memory management for large language model serving with pagedattention.
\newblock In \emph{Proceedings of the 29th symposium on operating systems principles}, pages 611--626, 2023.

\bibitem[Li et~al.(2026{\natexlab{a}})Li, Guo, Lu, Wang, Song, Shen, Wu, and Sun]{li2026optargus}
Zhong Li, Zihan Guo, Xiaohan Lu, Juntao Wang, Jie Song, Chao Shen, Jiageng Wu, and Mingyang Sun.
\newblock {OptArgus}: A multi-agent system to detect hallucinations in {LLM}-based optimization modeling, 2026{\natexlab{a}}.
\newblock URL \url{https://arxiv.org/abs/2605.11738}.
\newblock arXiv:2605.11738.

\bibitem[Li et~al.(2026{\natexlab{b}})Li, Huang, Wan, Wang, Zhang, Chen, Liu, Wen, and Luo]{li2026costly}
Zhong Li, Xin Huang, Jinhui Wan, Xiangyi Wang, Shenkai Zhang, Ruiqi Chen, Wenyu Liu, Zaiwen Wen, and Ziyan Luo.
\newblock Right answers, costly models: The efficiency gap in {LLM}-based optimization modeling, 2026{\natexlab{b}}.
\newblock URL \url{https://arxiv.org/abs/2609.38884}.
\newblock arXiv:2609.38884.

\bibitem[Li et~al.(2026{\natexlab{c}})Li, Lu, Wei, Chen, Liu, Lan, Zhang, and Wen]{li2026industrial}
Zhong Li, Hongliang Lu, Tao Wei, Yuxuan Chen, Wenyu Liu, Yuan Lan, Fan Zhang, and Zaiwen Wen.
\newblock Constructing industrial-scale optimization modeling benchmark.
\newblock In \emph{Proceedings of the 43rd International Conference on Machine Learning}, volume 306 of \emph{Proceedings of Machine Learning Research}, pages 68993--69036. PMLR, 2026{\natexlab{c}}.
\newblock URL \url{https://proceedings.mlr.press/v306/li26dz.html}.

\bibitem[Liu et~al.(2025)Liu, Mei, Lin, Xue, Wang, Xu, Wu, Zhang, Lin, Dong, et~al.]{liu2025deepseek}
Aixin Liu, Aoxue Mei, Bangcai Lin, Bing Xue, Bingxuan Wang, Bingzheng Xu, Bochao Wu, Bowei Zhang, Chaofan Lin, Chen Dong, et~al.
\newblock Deepseek-v3. 2: Pushing the frontier of open large language models.
\newblock \emph{arXiv preprint arXiv:2512.02556}, 2025.

\bibitem[Liu et~al.(2026{\natexlab{a}})Liu, Yang, Liu, SONG, Gao, and Liu]{liu2026mmagent}
Fan Liu, Zhe-Rui Yang, Cancheng Liu, Tianrui SONG, Xiaofeng Gao, and Hao Liu.
\newblock {MM}-agent: {LLM} as agents for real-world mathematical modeling problem.
\newblock In \emph{The Thirty-ninth Annual Conference on Neural Information Processing Systems}, 2026{\natexlab{a}}.
\newblock URL \url{https://openreview.net/forum?id=o8n5oNDsiq}.

\bibitem[Liu et~al.(2026{\natexlab{b}})Liu, Wang, Cai, Han, Kuang, and HAO]{liu2026optitree}
Haoyang Liu, Jie Wang, Yuyang Cai, Xiongwei Han, Yufei Kuang, and Jianye HAO.
\newblock Optitree: Hierarchical thoughts generation with tree search for {LLM} optimization modeling.
\newblock In \emph{The Thirty-ninth Annual Conference on Neural Information Processing Systems}, 2026{\natexlab{b}}.
\newblock URL \url{https://openreview.net/forum?id=Ej20yjWMCj}.

\bibitem[Liu et~al.(2026{\natexlab{c}})Liu, Wu, Kuang, Han, Zhong, Feng, and Lu]{liu2026localizable}
Weiting Liu, Han Wu, Yufei Kuang, Xiongwei Han, Tao Zhong, Jianfeng Feng, and Wenlian Lu.
\newblock Automated optimization modeling via a localizable error-driven perspective, 2026{\natexlab{c}}.
\newblock URL \url{https://arxiv.org/abs/2602.11164}.
\newblock arXiv:2602.11164.

\bibitem[Lu et~al.(2025)Lu, Xie, Wu, Ren, Chen, and Wen]{lu2025optmath}
Hongliang Lu, Zhonglin Xie, Yaoyu Wu, Can Ren, Yuxuan Chen, and Zaiwen Wen.
\newblock Optmath: A scalable bidirectional data synthesis framework for optimization modeling.
\newblock In \emph{Forty-second International Conference on Machine Learning}, 2025.

\bibitem[Prasath and Karande(2023)]{prasath2023synthesis}
Ganesh Prasath and Shirish Karande.
\newblock Synthesis of mathematical programs from natural language specifications.
\newblock \emph{arXiv preprint arXiv:2304.03287}, 2023.

\bibitem[Ramamonjison et~al.(2022)Ramamonjison, Yu, Li, Li, Carenini, Ghaddar, He, Mostajabdaveh, Banitalebi-Dehkordi, Zhou, and Zhang]{ramamonjison_nl4opt_2023}
Rindranirina Ramamonjison, Timothy Yu, Raymond Li, Haley Li, Giuseppe Carenini, Bissan Ghaddar, Shiqi He, Mahdi Mostajabdaveh, Amin Banitalebi-Dehkordi, Zirui Zhou, and Yong Zhang.
\newblock {NL4Opt} {Competition}: {Formulating} {Optimization} {Problems} {Based} on {Their} {Natural} {Language} {Descriptions}.
\newblock In \emph{Proceedings of the NeurIPS 2022 Competitions Track}, volume 220 of \emph{Proceedings of Machine Learning Research}, pages 189--203. PMLR, 2022.
\newblock URL \url{https://proceedings.mlr.press/v220/ramamonjison23a.html}.

\bibitem[Shang et~al.(2025)Shang, Liu, Zhu, Zhang, Xu, Guan, Zhang, Dong, Zhou, Zhang, Xin, Miao, Li, Yang, and Yang]{shang_rstar2-agent_2025}
Ning Shang, Yifei Liu, Yi~Zhu, Li~Lyna Zhang, Weijiang Xu, Xinyu Guan, Buze Zhang, Bingcheng Dong, Xudong Zhou, Bowen Zhang, Ying Xin, Ziming Miao, Scarlett Li, Fan Yang, and Mao Yang.
\newblock {rStar2}-{Agent}: {Agentic} {Reasoning} {Technical} {Report}, August 2025.
\newblock URL \url{http://arxiv.org/abs/2508.20722}.
\newblock arXiv:2508.20722 [cs].

\bibitem[Shao et~al.(2024)Shao, Wang, Zhu, Xu, Song, Bi, Zhang, Zhang, Li, Wu, et~al.]{shao2024deepseekmath}
Zhihong Shao, Peiyi Wang, Qihao Zhu, Runxin Xu, Junxiao Song, Xiao Bi, Haowei Zhang, Mingchuan Zhang, YK~Li, Yang Wu, et~al.
\newblock Deepseekmath: Pushing the limits of mathematical reasoning in open language models.
\newblock \emph{arXiv preprint arXiv:2402.03300}, 2024.

\bibitem[Singh(2012)]{singh2012overview}
Ajay Singh.
\newblock An overview of the optimization modelling applications.
\newblock \emph{Journal of Hydrology}, 466:\penalty0 167--182, 2012.

\bibitem[Tang et~al.(2025)Tang, Ye, Huang, Huang, Li, Li, Chen, Yan, Wang, Zha, Liu, and Wang]{tang_calm_2025}
Zhengyang Tang, Zihan Ye, Chenyu Huang, Xuhan Huang, Chengpeng Li, Sihang Li, Guanhua Chen, Ming Yan, Zizhuo Wang, Hongyuan Zha, Dayiheng Liu, and Benyou Wang.
\newblock {CALM} {Before} the {STORM}: {Unlocking} {Native} {Reasoning} for {Optimization} {Modeling}.
\newblock In \emph{Submitted to The Fourteenth International Conference on Learning Representations}, 2025.
\newblock URL \url{https://openreview.net/forum?id=DILQqCQIJ3}.

\bibitem[Wu et~al.(2026)Wu, Shen, Guo, Shao, Zhang, Song, Li, and Sun]{wu2026fix}
Jiageng Wu, Chao Shen, Zihan Guo, Junyan Shao, Zongyan Zhang, Jie Song, Zhong Li, and Mingyang Sun.
\newblock Fix what's wrong, keep what's right: Hallucination repair for {LLM}-based optimization modeling, 2026.
\newblock URL \url{https://arxiv.org/abs/2610.12529}.
\newblock arXiv:2610.12529.

\bibitem[Xiao et~al.(2024)Xiao, Zhang, Wu, Xu, Wang, Han, Fu, Zhong, Zeng, Song, and Chen]{xiao_chain--experts_2024}
Ziyang Xiao, Dongxiang Zhang, Yangjun Wu, Lilin Xu, Yuan Wang, Xiongwei Han, Xiaojin Fu, Tao Zhong, Jia Zeng, Mingli Song, and Gang Chen.
\newblock {CHAIN}-{OF}-{EXPERTS}: {WHEN} {LLMS} {MEET} {COMPLEX} {OPERATIONS} {RESEARCH} {PROBLEMS}.
\newblock In \emph{The Twelfth International Conference on Learning Representations}, 2024.
\newblock URL \url{https://openreview.net/forum?id=HobyL1B9CZ}.

\bibitem[Xiao et~al.(2025)Xiao, Xie, Xu, Guan, Zhu, Han, Fu, Yu, Wu, Shi, Kang, Duan, Zhong, Yuan, Zeng, Wang, Chen, and Zhang]{ijcai2025p1192}
Ziyang Xiao, Jingrong Xie, Lilin Xu, Shisi Guan, Jingyan Zhu, Xiongwei Han, Xiaojin Fu, WingYin Yu, Han Wu, Wei Shi, Qingcan Kang, Jiahui Duan, Tao Zhong, Mingxuan Yuan, Jia Zeng, Yuan Wang, Gang Chen, and Dongxiang Zhang.
\newblock A survey of optimization modeling meets {LLMs}: Progress and future directions.
\newblock In James Kwok, editor, \emph{Proceedings of the Thirty-Fourth International Joint Conference on Artificial Intelligence, {IJCAI-25}}, pages 10742--10750. International Joint Conferences on Artificial Intelligence Organization, 8 2025.
\newblock \doi{10.24963/ijcai.2025/1192}.
\newblock URL \url{https://doi.org/10.24963/ijcai.2025/1192}.
\newblock Survey Track.

\bibitem[Xie et~al.(2025)Xie, Hu, Xu, Wu, and Li]{xie2025murka}
Wantong Xie, Yi-Xiang Hu, Jieyang Xu, Feng Wu, and Xiangyang Li.
\newblock Murka: Multi-reward reinforcement learning with knowledge alignment for optimization tasks.
\newblock In \emph{The Thirty-ninth Annual Conference on Neural Information Processing Systems}, 2025.

\bibitem[Xue et~al.(2026)Xue, Zheng, Liu, Li, Zheng, MA, and An]{xue2026simpletir}
Zhenghai Xue, Longtao Zheng, Qian Liu, Yingru Li, Xiaosen Zheng, Zejun MA, and Bo~An.
\newblock Simple{TIR}: End-to-end reinforcement learning for multi-turn tool-integrated reasoning.
\newblock In \emph{The Fourteenth International Conference on Learning Representations}, 2026.
\newblock URL \url{https://openreview.net/forum?id=EplNy91Xqh}.

\bibitem[Yang et~al.(2026)Yang, Jiang, Chen, Yang, Wong, Gao, Kooi, Li, Shi, Qiu, Huang, Zu, Yang, Zhang, Wong, Ilievski, Yu, Plaat, Ren, Hoogendoorn, and Fran{\c{c}}ois-Lavet]{yang2026modularized}
Zhao Yang, Yuxuan Jiang, Ting-Chih Chen, Lincen Yang, Annie Wong, Chao Gao, Jacob~E. Kooi, Zhong Li, Jiayang Shi, Kevin Qiu, Qi~Huang, Xinrui Zu, Shiping Yang, Hengyuan Zhang, Ngai Wong, Filip Ilievski, Shujian Yu, Aske Plaat, Zhaochun Ren, Mark Hoogendoorn, and Vincent Fran{\c{c}}ois-Lavet.
\newblock Modularized reinforcement learning on {LLMs}: From {MDP} creation to exploration and learning, 2026.
\newblock URL \url{https://arxiv.org/abs/2606.21943}.
\newblock arXiv:2606.21943.

\bibitem[Yang et~al.(2024)Yang, Wang, Huang, Guo, Shi, Han, Feng, Song, Liang, and Tang]{yang_optibench_2024}
Zhicheng Yang, Yiwei Wang, Yinya Huang, Zhijiang Guo, Wei Shi, Xiongwei Han, Liang Feng, Linqi Song, Xiaodan Liang, and Jing Tang.
\newblock {OptiBench} {Meets} {ReSocratic}: {Measure} and {Improve} {LLMs} for {Optimization} {Modeling}, October 2024.
\newblock URL \url{http://arxiv.org/abs/2407.09887}.
\newblock arXiv:2407.09887 [cs].

\bibitem[Yu et~al.(2025)Yu, Zhang, Zhu, Yuan, Zuo, Yue, Dai, Fan, Liu, Liu, Liu, Lin, Lin, Ma, Sheng, Tong, Zhang, Zhang, Zhang, Zhu, Zhu, Chen, Chen, Wang, Yu, Song, Wei, Zhou, Liu, Ma, Zhang, Yan, Qiao, Wu, and Wang]{yu_dapo_2025}
Qiying Yu, Zheng Zhang, Ruofei Zhu, Yufeng Yuan, Xiaochen Zuo, Yu~Yue, Weinan Dai, Tiantian Fan, Gaohong Liu, Lingjun Liu, Xin Liu, Haibin Lin, Zhiqi Lin, Bole Ma, Guangming Sheng, Yuxuan Tong, Chi Zhang, Mofan Zhang, Wang Zhang, Hang Zhu, Jinhua Zhu, Jiaze Chen, Jiangjie Chen, Chengyi Wang, Hongli Yu, Yuxuan Song, Xiangpeng Wei, Hao Zhou, Jingjing Liu, Wei-Ying Ma, Ya-Qin Zhang, Lin Yan, Mu~Qiao, Yonghui Wu, and Mingxuan Wang.
\newblock {DAPO}: {An} {Open}-{Source} {LLM} {Reinforcement} {Learning} {System} at {Scale}, May 2025.
\newblock URL \url{http://arxiv.org/abs/2503.14476}.
\newblock arXiv:2503.14476 [cs].

\bibitem[Zhang et~al.(2026)Zhang, Geng, Yu, Yin, Zhang, Tan, Zhou, Li, Xue, Li, Zhou, Chen, Zhang, Fan, Wang, Huang, Velez, Liao, WANG, Yang, Ji, Wang, YAN, Torr, and BAI]{zhang2026the}
Guibin Zhang, Hejia Geng, Xiaohang Yu, Zhenfei Yin, Zaibin Zhang, Zelin Tan, Heng Zhou, Zhong-Zhi Li, Xiangyuan Xue, Yijiang Li, Yifan Zhou, Yang Chen, Chen Zhang, Yutao Fan, Zihu Wang, Songtao Huang, Francisco~Piedrahita Velez, Yue Liao, Hongru WANG, Mengyue Yang, Heng Ji, Jun Wang, Shuicheng YAN, Philip Torr, and LEI BAI.
\newblock The landscape of agentic reinforcement learning for {LLM}s: A survey.
\newblock \emph{Transactions on Machine Learning Research}, 2026.
\newblock ISSN 2835-8856.
\newblock URL \url{https://openreview.net/forum?id=RY19y2RI1O}.
\newblock Survey Certification.

\bibitem[Zhang et~al.(2025)Zhang, Liu, Lv, Sun, Jing, Iong, Hou, Qi, Lai, Xu, Lu, Wang, Tang, and Dong]{zhang_agentrl_2025}
Hanchen Zhang, Xiao Liu, Bowen Lv, Xueqiao Sun, Bohao Jing, Iat~Long Iong, Zhenyu Hou, Zehan Qi, Hanyu Lai, Yifan Xu, Rui Lu, Hongning Wang, Jie Tang, and Yuxiao Dong.
\newblock {AgentRL}: {Scaling} {Agentic} {Reinforcement} {Learning} with a {Multi}-{Turn}, {Multi}-{Task} {Framework}, October 2025.
\newblock URL \url{https://openreview.net/forum?id=zq3vAmuUk9}.

\bibitem[Zhang et~al.(2024)Zhang, Wang, Guo, Wang, Lin, Yang, and Yin]{zhang-etal-2024-solving}
Jihai Zhang, Wei Wang, Siyan Guo, Li~Wang, Fangquan Lin, Cheng Yang, and Wotao Yin.
\newblock Solving {General} {Natural}-{Language}-{Description} {Optimization} {Problems} with {Large} {Language} {Models}.
\newblock In Yi~Yang, Aida Davani, Avi Sil, and Anoop Kumar, editors, \emph{Proceedings of the 2024 {Conference} of the {North} {American} {Chapter} of the {Association} for {Computational} {Linguistics}: {Human} {Language} {Technologies} ({Volume} 6: {Industry} {Track})}, pages 483--490, Mexico City, Mexico, June 2024. Association for Computational Linguistics.
\newblock URL \url{https://aclanthology.org/2024.naacl-industry.42}.

\bibitem[Zhou et~al.(2026)Zhou, Xu, Lin, and Ge]{zhou_steporlm_2025}
Chenyu Zhou, Tianyi Xu, Jianghao Lin, and Dongdong Ge.
\newblock {StepORLM}: {A} {Self}-{Evolving} {Framework} {With} {Generative} {Process} {Supervision} {For} {Operations} {Research} {Language} {Models}.
\newblock In \emph{The Fourteenth International Conference on Learning Representations}, 2026.
\newblock URL \url{https://openreview.net/forum?id=ZrgxU8WMmG}.

\bibitem[Zhu et~al.(2025)Zhu, Xie, Lv, and slime Contributors]{slime_github}
Zilin Zhu, Chengxing Xie, Xin Lv, and slime Contributors.
\newblock slime: An llm post-training framework for rl scaling.
\newblock \url{https://github.com/THUDM/slime}, 2025.
\newblock GitHub repository. Corresponding author: Xin Lv.

\end{thebibliography}

\newpage
\appendix

{\large{\textbf{Appendix for \textit{PEARL: Solver-in-the-Loop Interactive Optimization Modeling from Natural Language}}}}

\section*{\large Table of Contents}
{\footnotesize
{
\begin{itemize}
    \item[A] \hyperref[app:design_details]{Design Details on PEARL}
    \begin{itemize}
        \item[A.1] \hyperref[app:subsec:rl_comparison]{Comparison with RL-Based Methods}
        \item[A.2] \hyperref[app:subsec:reward_func]{Reward Function Design}
        \item[A.3] \hyperref[app:subsec:training_algorithm]{PEARL Training Algorithm}
        \item[A.4] \hyperref[fig:pearl_overview]{Training and Inference Loop}
        \item[A.5] \hyperref[tab:qwen3_4b_PEARL_hparams]{Training Hyperparameters}
    \end{itemize}
    \item[B] \hyperref[app:experiments_setup]{Experiment Setup}
    \begin{itemize}
        \item[B.1] \hyperref[app:subsec:workflow_overview]{Workflow Overview}
        \item[B.2] \hyperref[app:subsec:training_setup]{Training Setup and Loss Masking}
        \item[B.3] \hyperref[app:subsec:inference_setup]{Inference Setup}
        \item[B.4] \hyperref[app:subsec:8b_comparison]{8B Model Comparison}
    \end{itemize}
    \item[C] \hyperref[app:failure_mode_attribution]{Failure Mode Attribution}
    \item[D] \hyperref[app:ablation_sensitivity]{Ablation Study and Sensitivity Analysis}
    \begin{itemize}
        \item[D.1] \hyperref[app:subsec:sft]{Supervised Fine-Tuning Initialization}
        \begin{itemize}
            \item[D.1.1] \hyperref[app:subsubsec:sft_theory]{Theoretical Analysis}
            \item[D.1.2] \hyperref[app:subsec:role_of_SFT]{Experiments}
        \end{itemize}
        \item[D.2] \hyperref[app:subsec:trajectory_filtering]{Adaptive Sampling with Trajectory Filtering}
        \begin{itemize}
            \item[D.2.1] \hyperref[app:subsec:trajectory_filtering_theory]{Theoretical Analysis}
            \item[D.2.2] \hyperref[app:subsec:results_trajectory_filtering]{Experiments}
        \end{itemize}
        \item[D.3] \hyperref[app:subsec:training_dynamics]{Training Dynamics}
        \item[D.4] \hyperref[app:subsec:tool_use_depth]{Tool-Use Depth and Termination}
    \end{itemize}
    \item[E] \hyperref[app:case_study]{PEARL Rollout Example}
    \item[F] \hyperref[app:prompt_templates]{Prompt Templates}
    \begin{itemize}
        \item[F.1] \hyperref[app:training_prompt]{Training Prompt}
        \item[F.2] \hyperref[app:direct_prompt]{Direct Testing Prompt}
    \end{itemize}
    \item[G] \hyperref[app:data_collection]{{Training Data and Benchmark Documentation}}
    \item[{H}] \hyperref[app:comparison_protocols]{{Additional Experiments and Protocols}}
    \item[{I}] \hyperref[app:limitations_impact]{Limitations and Broader Impact}
\end{itemize}
}
}

\section{Design Details on PEARL}
\label{app:design_details}

{\subsection{Comparison of PEARL and other RL-based Methods}
\label{app:subsec:rl_comparison}
Table~\ref{tab:agentic_rl_comparison} compares PEARL with recent reinforcement learning--based methods for optimization modeling, including SIRL~\citep{chen2025solverinformed}, OR-R1~\citep{ding2026or}, MURKA~\citep{xie2025murka}, StepORLM~\citep{zhou_steporlm_2025}, and STORM~\citep{tang_calm_2025}. These methods make important progress by replacing purely supervised imitation with solver- or preference-grounded learning signals. {Their training targets and interaction architectures differ: for example}, {MURKA combines GRPO-trained extraction}, a {distilled Solver}, and {an iterative Checker. Thus,} the {distinction is not that prior systems never perform iterative verification, but how} solver {interaction enters the learned policy}.}

{\noindent\textbf{The key distinction is where RL is applied.}
{Many prior} RL-based methods optimize \emph{outputs} {or individual workflow components}: the learned behavior is to make {a} response more likely to be correct, executable, or process-consistent. PEARL instead applies RL to the \emph{modeling process}. The trajectory consists of multiple interaction steps, where the policy can decide to draft, revise, execute code, inspect solver feedback, and terminate. Thus, PEARL does not merely learn what a good final solver program looks like; it learns a control policy for navigating the solve--debug--revise loop.}

{\noindent\textbf{PEARL treats solver interaction as part of the environment.}
In {SIRL}~{\citep{chen2025solverinformed}}, solver {results provide rewards for completed programs during training. In MURKA}~{\citep{xie2025murka}, a dedicated Checker tests candidate solutions and sends corrective feedback to the Solver.} In PEARL, {multi-turn RL trains the model to manage the interaction itself: it chooses when to run code and uses the output to decide whether to revise or finish. The model receives execution output and} errors, {including solver diagnostics or checks printed by its code. Only the external scorer compares the final objective value with the reference; the model does not receive the reference or the comparison result.}}

\noindent\textbf{This agentic formulation gives PEARL three practical advantages.}
First, it supports \emph{adaptive computation}: the policy can stop early on easy instances and spend additional tool calls on harder ones. Second, it supports \emph{solver-aware repair}: errors are not only penalized after the fact, but are observed and acted upon during the trajectory. Third, it supports \emph{long-horizon credit assignment}: choices such as indexing schemes, data parsing, or early execution can receive credit based on their downstream effect on final solver-verified correctness. These advantages are especially relevant for optimization modeling, where many failures are not visible from the generated text alone and only emerge through execution.

The comparison in Table~\ref{tab:agentic_rl_comparison} should therefore be read as a methodological distinction rather than a claim that prior RL methods are ineffective. Prior methods improve the quality of optimization-modeling outputs under static or weakly interactive generation. PEARL targets a different training problem: learning an agent that can use solver feedback during generation as an observable signal for deciding when to execute, how to revise, and when to terminate.

\begin{table}[htbp]

\centering
\caption{{Comparison of RL-based} methods {for optimization modeling}. The {table summarizes what each method learns and how it uses solver feedback during training or inference.}}
\label{tab:agentic_rl_comparison}
\small
\setlength{\tabcolsep}{4pt}
\renewcommand{\arraystretch}{1.1}
\begin{tabularx}{\linewidth}{@{}>{\raggedright\arraybackslash\leavevmode}p{0.14\linewidth}>{\raggedright\arraybackslash\leavevmode}p{0.34\linewidth}>{\raggedright\arraybackslash\leavevmode}X@{}}
\toprule
\textbf{{Method}} & \textbf{{Learning target}} & \textbf{{Role of solver feedback}}\\
\midrule
{SIRL} & {Outcome-based learning of complete generations} & {Terminal execution}, {feasibility}, and {objective-based scoring}.\\
\addlinespace[0.45em]
OR-R1 & {Output-level RL} & {Solver outcomes score generated solutions.}\\
\addlinespace[0.45em]
MURKA & {GRPO-trained Extractor and distilled Solver} & {A separate Checker iteratively verifies solutions through solver feedback.}\\
\addlinespace[0.45em]
StepORLM & {Generative process supervision and preference learning} & {Solver outcomes and process feedback guide reasoning-trajectory learning.}\\
\addlinespace[0.45em]
STORM & {Curriculum-guided output learning} & {Outcome signals support curriculum-shaped alignment.}\\
\midrule
\textbf{PEARL} & Multi-turn {policy for execution, revision, and termination} & {Solver} output {conditions later actions; trajectory rewards combine format and objective agreement.}\\
\bottomrule
\end{tabularx}
\end{table}

\subsection{Details on Reward Function Design}
\label{app:subsec:reward_func}
The reward function is designed to encourage correct formatting, proper tool usage, and accurate optimization modeling. The total reward is decomposed into {format rewards and {an objective-agreement reward}}:

\begin{itemize}
    \item \textbf{Format Reward ($r_{\text{fmt}}$):}
    \begin{itemize}
        \item {A reward of $0.1$ is assigned {once} if the {complete response contains both} \texttt{{<interpreter>}} {and} \texttt{{</interpreter>}}{. This is} a {tag-presence check, not a check of execution success or a per-call bonus}.}
        \item {A reward of $0.2$ is assigned {when} the {parser extracts a nonempty} \texttt{'''python ... '''} {block from the last complete} \texttt{{<answer>}.{..</answer>}} {block. Nested answer tags and interaction tags inside the extracted code are rejected. The implementation uses triple single quotes for this format.}}
    \end{itemize}

    \item \textbf{{{Objective-Agreement} Reward} ($r_{\text{corr}}$):}
    {{We evaluate objective agreement} by executing the extracted code and comparing the {resulting} objective value {with} the {reference value}.} Specifically, the system appends a value probe to the generated code to capture the solver's output.
    \begin{itemize}
        \item {If the {extracted} code executes successfully (return code 0) and {the appended probe yields} an {objective} value $v_{\text{pred}}$, we compare it with ${v_{\text{label}}}${. The probe reads a Pyomo objective value or a Gurobi incumbent objective;} the {scorer does not independently require an optimal termination status}.}
        \item If the absolute difference $|v_{\text{pred}} - v_{\text{label}}| \le 10^{-3}$, the solution {{passes the objective-agreement criterion}}, and a reward of $1.0$ is granted.
    \end{itemize}
\end{itemize}

{The total reward {is at most} ${1.3}${. A valid final answer with an accepted objective can receive} ${1.2}$ {without any tool call. If the final code cannot be extracted, both the} ${0.2}$ {final-format bonus and the objective-agreement reward are zero; the interpreter-tag bonus is evaluated independently. Thus, although the components are additive, objective scoring depends on successful final-code extraction.}}

{\paragraph{{Multiple optima and missing references.}}
{The scorer compares objective values, not} a {canonical decision vector, so distinct optimal solutions (including degenerate optima) are accepted when they return} the {reference objective. Any internal transformation} of the {objective must be converted back} to the problem{'s original objective definition before comparison. This does not check every intended constraint. Without a reference objective,} the {current training scorer can award only format bonuses; such instances are excluded from the reported training corpus. Pure constraint-satisfaction problems require a separate assignment verifier and are not evaluated here}.}

\subsection{PEARL Training Algorithm}
\label{app:subsec:training_algorithm}

Algorithm~\ref{alg:pearl} gives the training loop corresponding to the objective in Section~\ref{subsec:training}. The algorithm starts from a reference policy $\pi_{\mathrm{ref}}$, which is either an instruction-tuned backbone or an SFT-warm-started model depending on the experiment. Each training iteration samples optimization-modeling instances from $\mathcal{D}$ and rolls out the current policy in the PEARL environment. A rollout is not a single completion: it is a multi-turn trajectory in which the agent may draft a formulation, invoke Python or solver tools, observe execution feedback, revise the formulation or implementation, and eventually emit a final answer.

The return {$R({\tau_{i,j}})$} is computed at the trajectory level using the reward components described in Appendix~\ref{app:subsec:reward_func}: {format {rewards check interpreter-tag presence} and {final-code extraction}}, while {{the terminal} objective-agreement {reward requires successful execution and an accepted objective; execution and feasibility} are {not separately rewarded}}. We then convert returns into normalized advantages, so credit is assigned to complete solver-grounded interaction traces rather than to isolated text spans. This design is important in optimization modeling because early actions, such as choosing an indexing scheme or deciding when to execute, may only reveal their value after several tool calls.

Finally, the policy is updated with the GRPO-style objective described in Section~\ref{subsec:training}, using DAPO-inspired decoupled clipping with thresholds $\epsilon_+$ and $\epsilon_-$. In the main PEARL runs, KL and entropy coefficients are set to zero after initialization; the update is therefore driven by solver-grounded returns and stabilized by clipping and {{within-problem group} normalization}. The high-level loop in Algorithm~\ref{alg:pearl} omits engineering details such as loss masking over tool observations, execution timeouts, and distributed rollout scheduling, which are summarized in Appendix~\ref{app:subsec:training_setup} and Table~\ref{tab:qwen3_4b_PEARL_hparams}.

\begin{algorithm}[h]
\caption{\textsc{PEARL}: NL-to-Opt via Agentic RL}
\label{alg:pearl}
\DontPrintSemicolon
\small

\KwIn{Dataset $\mathcal{D}$ of optimization modeling instances;
reference policy $\pi_{\mathrm{ref}}$;
clipping thresholds $\epsilon_{+}, \epsilon_{-}$}

Initialize policy $\pi_\theta \leftarrow \pi_{\mathrm{ref}}$\;

\While{not converged}{
  \ForEach{$x_i \sim \mathcal{D}$}{
    {Sample ${G=8}$ multi-turn, tool-augmented {trajectories}
    ${\{\tau_{i,j}\}_{j=1}^{G}} \sim \pi_\theta(\cdot \mid x_i)$}\;
    {\hfill{\footnotesize\color{gray}// NL modeling + execution + solver calls}}\;

    {Compute {each} trajectory return $R({\tau_{i,j}})$}\;
    {\hfill{\footnotesize{// {Format and terminal objective agreement}}}}\;

    {Compute ${A_{i,j}}$ {by normalizing returns within group} ${i}$}\;
    {\hfill{\footnotesize\color{gray}// Credit assignment for optimization modeling}}\;
  }

  Update $\pi_\theta$ via decoupled-clipped policy gradient\;
  {\hfill{\footnotesize\color{gray}// Agentic RL over multi-turn optimization modeling}}\;
}
\end{algorithm}

\begin{figure}[h]
\centering
\includegraphics[width=0.78\linewidth]{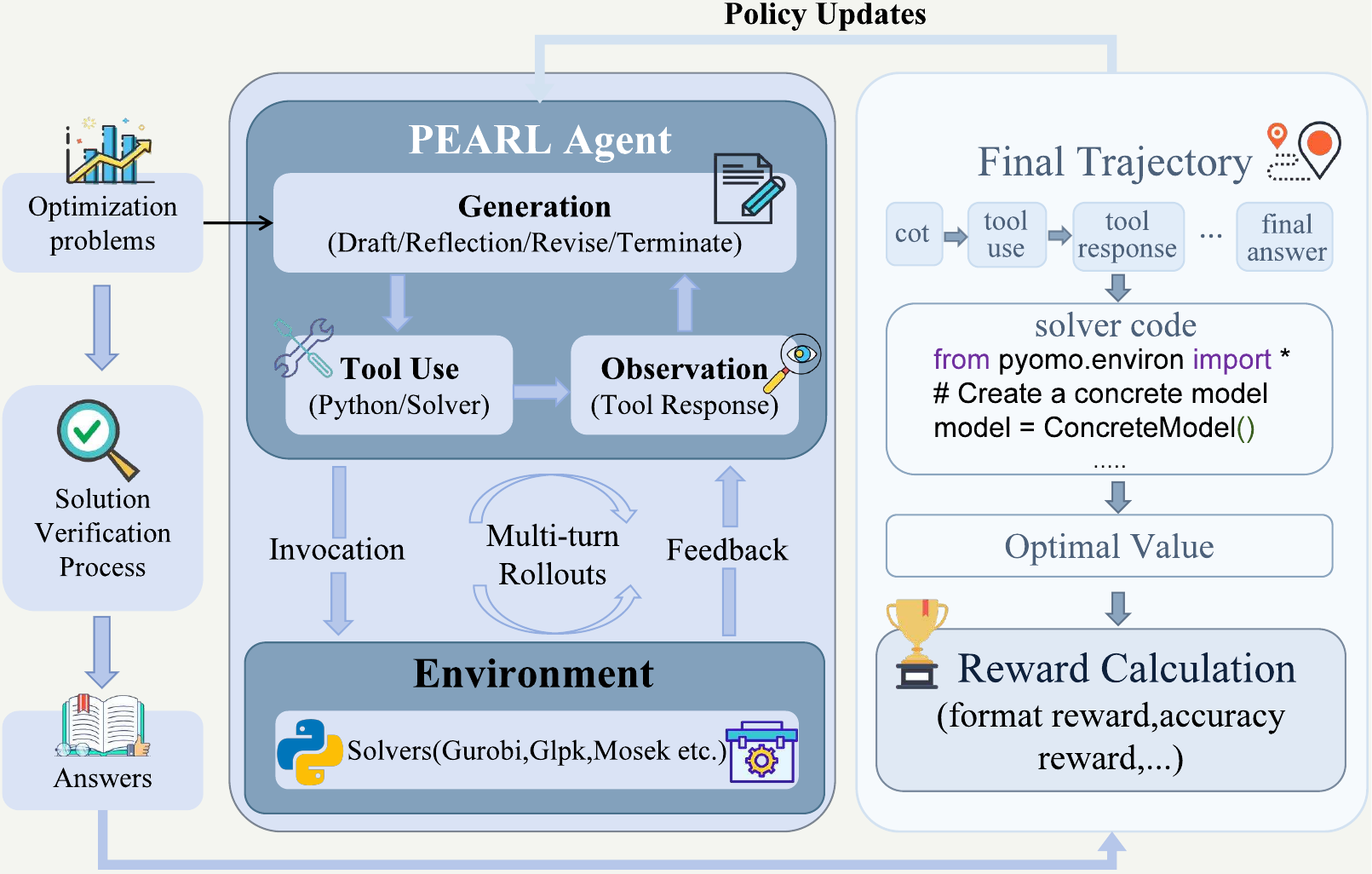}
\caption{Overview of the PEARL training and inference loop for optimization modeling with tool use{.} {{The right-hand branch depicts external training-time scoring; its objective-value probe does not certify optimal solver termination}.}}
\label{fig:pearl_overview}
\end{figure}

{Figure~\ref{fig:pearl_overview} complements the pseudo-code{: both} training and inference {expose execution output to} the {policy} during {generation. After generation, the} training {scorer separately executes} the {final code and compares its objective with} the {reference} to {compute the terminal reward}.}

\begin{table}[h]
\centering
\caption{Main hyperparameters for Qwen3-4B-Instruct-2507 PEARL training.}
\label{tab:qwen3_4b_PEARL_hparams}
\resizebox{0.68\linewidth}{!}{%
\begin{tabular}{ll}
\toprule
\textbf{Hyperparameter} & \textbf{Value} \\
\midrule
Base model & Qwen3-4B-Instruct-2507 \\
Hardware & 1 node, 4 {high-performance GPUs}, Ray-based training \\
Rollout engine & SGLang tool-call rollout (colocated), 4 GPUs per engine \\
Algorithm & GRPO with decoupled clipping ($\epsilon{=}0.2$, $\epsilon_{\text{high}}{=}0.28$) \\
KL regularization & disabled in main runs, coefficient $0.0$ \\
Entropy regularization & coefficient $0.0$ \\
{{Configured rollout iterations}} & {3000, shuffle enabled} \\
Samples per prompt & 8 \\
Rollout batch size & 16 prompts per rollout batch \\
Global batch size & 128 \\
Max response length & 16384 tokens \\
Sampling temperature & 0.8 \\
Optimizer & Adam, constant LR \\
Learning rate & $1\times 10^{-6}$ \\
Adam betas & $(0.9, 0.98)$ \\
Weight decay & 0.1 \\
Parallelism & tensor MP=2, context MP=2, pipeline MP=1, sequence parallel on \\
Memory / efficiency & dynamic batch size, max tokens per GPU 9216, activation recompute enabled \\
Checkpointing & save interval 20 \\
\midrule
\multicolumn{2}{l}{\textit{PEARL-specific tool-use settings}} \\
\midrule
{Max {generation} turns per rollout} & {8{, including the final-answer} turn} \\
{{Tool-output truncation}} & {{First} 2000 {characters, plus truncation notice}} \\
Tool concurrency & 64 \\
{{Training} Python {execution} timeout} & 120 seconds \\
Python tool memory limit & 4 GB \\
Return log probabilities & enabled \\
\bottomrule
\end{tabular}%
}
\end{table}

\section{Experiment Setup}
\label{app:experiments_setup}

\subsection{Workflow overview}
\label{app:subsec:workflow_overview}

As shown in Figure~\ref{fig:pearl_overview}, we cast optimization modeling from natural language as an interactive decision process. Given a problem description, the PEARL agent incrementally constructs the final solution via staged generation (Draft/Reflection/Revise/Terminate) and queries external tools (Python / solvers) to obtain verifiable feedback such as execution traces, solver status, and objective values. Each interaction yields a multi-turn trajectory that contains intermediate reasoning, tool calls, tool observations, and the final modeling write-up with runnable code. {{After generation, the external scorer extracts} and {executes} the final {code and compares} the {extracted objective} value {with the reference. The trajectory-level format bonuses} and {objective-agreement} reward {are then combined} for policy updates{;} the {reference comparison is not returned as an in-loop observation}.}

The maximum turn budget is only an upper bound: the policy can terminate early by emitting the final \texttt{<answer>} action. In practice, tool use stabilizes at a small number of calls for most instances, supporting a bounded-horizon interaction policy.

\subsection{Training setup}
\label{app:subsec:training_setup}

\noindent\textbf{Implementation and tool environment.} We implement training with the \textsc{slime} framework~\citep{slime_github}, using its tool-augmented rollout engine and distributed training stack. Rollouts are generated with an engine that supports multi-turn tool calling, where the model can emit \texttt{<code>...</code>} blocks to run Python in a sandbox and observe results in \texttt{<interpreter>...</interpreter>} blocks, and finally return a single \texttt{<answer>...</answer>} block containing the finalized formulation and verified runnable code. We use Pyomo~\cite{bynum2021pyomo} as the {modeling {interface}} for optimization formulations. An example of PEARL Agent solving an optimization problem can be found in Appendix \ref{app:case_study}.

We train the policy with \textsc{GRPO}\cite{shao2024deepseekmath} .
Table~\ref{tab:qwen3_4b_PEARL_hparams} summarizes the main hyperparameters for our Qwen3-4B-Instruct-2507 experiments{.} 

{{The configured 3,000 rollout iterations are not 3,000 individual trajectories: a complete iteration contains 16 problem groups of eight trajectories. Returns are normalized within each group. The implementation has a batch-wide fallback when the sample count differs from the configured full batch; this fallback is distinct from the main full-batch procedure. Online evaluation during training uses the absolute objective-error tolerance} ${10^{-3}}$ {without format bonuses. Its accuracy curves should not be treated as evaluations under the final benchmark threshold in Section}~{\ref{sec:exp}}.}

\noindent\textbf{Loss masking for observation tokens.}
In our policy optimization procedure, the token-level losses are computed over multi-turn rollout sequences. In PEARL, each rollout sequence consists of both agent-generated tokens (modeling steps, reasoning, tool invocations in \texttt{<code>...</code>} blocks) and observation tokens returned by external tools (execution traces, solver outputs, error messages in \texttt{<interpreter>...</interpreter>} blocks). While optimizing agent-generated tokens enhances the model's ability to formulate optimization problems, interact with solvers, and perform multi-turn reasoning, applying the same optimization to observation tokens can lead to unintended learning dynamics, as these tokens represent deterministic tool outputs rather than learned agent behavior. To address this, we introduce loss masking for observation tokens, ensuring the policy gradient objective is computed only over agent-generated tokens, excluding tool-returned observations from the optimization process. This approach stabilizes training while preserving the agent's ability to leverage solver feedback for iterative model refinement.

\subsection{Inference setup}
\label{app:subsec:inference_setup}
For evaluation, we deploy the trained PEARL model using vLLM for efficient inference. Table~\ref{tab:qwen3_4b_inference_params} summarizes the inference configuration.

\begin{table}[h]
\centering
\caption{Inference parameters for Qwen3-4B-Instruct-2507 PEARL evaluation.}
\label{tab:qwen3_4b_inference_params}
 \resizebox{0.5\linewidth}{!}{% 
\begin{tabular}{ll}
\toprule
\textbf{Parameter} & \textbf{Value} \\
\midrule
Provider & vLLM (optrl\_vllm) \\
Model & Qwen/Qwen3-4B-Instruct-2507 \\
Tokenizer & Qwen/Qwen3-4B-Instruct-2507 \\
Max model length & 16384 tokens \\
Max output tokens & 16384 tokens \\
{{LLM configuration timeout}} & {300 seconds {(see note below)}} \\
\midrule
\multicolumn{2}{l}{\textit{Engine configuration}} \\
\midrule
Tensor parallel size & 1 \\
GPU memory utilization & 0.8 \\
\midrule
\multicolumn{2}{l}{\textit{Sampling parameters}} \\
\midrule
Temperature & 0.6 \\
Top-p & 0.95 \\
Top-k & 50 \\
\bottomrule
\end{tabular}
}
\end{table}

{{The 300-second entry is an LLM configuration field, not a whole-trajectory time budget. The local} \texttt{{optrl\_vllm}} {client does not enforce this field around generation. Python tool execution has a separate timeout; the inference tool configuration sets it to 120 seconds per call.}}

\subsection{8B Model Comparison}
\label{app:subsec:8b_comparison}

\begin{table*}[h]
    \centering
    \caption{Pass@1 accuracy (\%) for 8B-related comparisons. The 8B rows are reported in the appendix to keep the main table focused on the primary 4B PEARL model and representative baselines.}
    \label{tab:appendix_8b_results}
    \resizebox{\linewidth}{!}{%
    \begin{tabular}{lcccccccccc}
    \toprule
    \textbf{Method}
    & \textbf{NL4Opt}
    & \textbf{MAMO-E}
    & \textbf{MAMO-C}
    & \textbf{NLP4LP}
    & \textbf{OptMATH*}
    & \textbf{IndustryOR}
    & \textbf{ComplexOR}
    & \textbf{Resocratic}
    & Macro Avg
    & Micro Avg\\
    \midrule
    Qwen-3-8B
     & \accpm{38.32}{1.60} & \accpm{38.53}{2.20} & \accpm{12.61}{2.30} & \accpm{46.07}{1.90} & \accpm{3.01}{0.55} & \accpm{2.38}{1.35} & \accpm{5.56}{2.78} & \accpm{37.47}{1.70} & \accpm{22.99}{0.70} & \accpm{32.56}{0.60} \\
    PEARL-Qwen-3-8B
     & \accpm{86.45}{1.10} & \accpm{94.68}{0.75} & \accpm{57.66}{3.40} & \accpm{94.38}{1.20} & \accpm{24.10}{1.65} & \accpm{64.29}{6.20} & \accpm{38.89}{4.80} & \accpm{81.39}{0.90} & \accpm{67.73}{0.55} & \accpm{79.61}{0.30}\\
    PEARL-Qwen-3-4B-Instruct-2507
     & \accpm{85.28}{0.99} & \accpm{91.65}{0.65} & \accpm{76.12}{3.19} & \accpm{93.25}{1.59} & \accpm{16.87}{1.70} & \accpm{67.86}{8.42} & \accpm{41.66}{3.92} & \accpm{84.99}{0.52} & \accpm{69.71}{0.13} & \accpm{79.84}{0.25} \\
    \bottomrule
    \end{tabular}%
    }
\end{table*}

{Table~\ref{tab:appendix_8b_results} shows that PEARL training yields a large improvement over the raw Qwen-3-8B baseline, confirming that the learned interaction policy is beneficial beyond the 4B backbone. Comparing PEARL-Qwen-3-8B with PEARL-Qwen-3-4B-Instruct-2507, the 8B model improves on some easier or scale-sensitive benchmarks but does not dominate the 4B model overall, suggesting that both backbone choice and training data/interface stability matter. {These backbone comparisons} do not {isolate model size from differences} in {initialization and training behavior}.}

\section{Failure Mode Attribution}
\label{app:failure_mode_attribution}

Figure~\ref{fig:failure_mode_attribution} in the main text summarizes the failure attribution analysis. We include the detailed setup and interpretation here to avoid overloading the compact main-text figure.

\noindent\textbf{Attribution setup.}
We analyze failures from the Qwen3-4B-Instruct base model and the PEARL-Qwen3-4B-Instruct model under the main pass@1 evaluation setting. For each instance that fails the pass@1 correctness criterion, we assign one primary failure category using the execution-and-verification pipeline. The categories are: \emph{code/runtime}, where the generated program crashes, imports fail, variables are undefined, parsing fails, or no executable solver result can be obtained; \emph{format/extraction}, where the response violates the required answer or code format or the evaluator cannot reliably extract the intended implementation; \emph{timeout}, where execution exceeds the configured budget; \emph{infeasible/solver}, where the generated model reaches the solver but produces an infeasible, unbounded, or otherwise solver-side invalid status; \emph{objective mismatch}, where the code executes and produces a candidate solution but the objective value does not match the reference within tolerance; and \emph{other}, which covers residual failures that do not cleanly fit the preceding categories. Each bar in Figure~\ref{fig:failure_mode_attribution} is normalized within the corresponding set of failed cases, so the figure describes the composition of failures rather than the overall pass rate.

\noindent\textbf{Rows in the figure.}
The first two rows compare the failure composition before and after PEARL training. The first row aggregates failures of Qwen3-4B-Instruct under the PEARL evaluation pipeline, while the second row aggregates failures of PEARL-Qwen3-4B-Instruct. The third row is a paired subset analysis: it considers only instances that the base model failed but PEARL later solved, and attributes those solved cases according to the base model's original failure mode. This row therefore answers a different question from the first two rows: instead of describing what PEARL still gets wrong, it identifies which base-model failure modes PEARL most often repairs.

\noindent\textbf{Main interpretation.}
The base model's failures are dominated by {code execution errors}, which account for 66.2\% of its failed cases. This means that many errors occur before the evaluator can even assess the mathematical quality of the generated optimization formulation: the model may emit non-runnable code, inconsistent variable names, invalid data handling, or solver calls that fail at runtime. After PEARL training, the {share of code execution failures} drops to 14.3\%, indicating that interactive execution and solver feedback teach the model to produce substantially more executable solver programs.

The PEARL failure distribution is instead dominated by objective mismatch, which accounts for 53.4\% of PEARL's remaining failed cases. This increase should be interpreted as a conditional shift among failures, not as evidence that PEARL creates more objective errors overall. Because PEARL fixes many low-level execution and formatting failures, the residual error mass becomes concentrated in harder cases where the code runs but the formulation, objective, indexing, or constraints still do not match the reference semantics. In other words, agentic training moves the dominant failure mode from ``cannot execute or parse'' toward ``executes but is mathematically wrong,'' which is a more substantive optimization-modeling failure.

The paired subset in the third row further clarifies what PEARL repairs. Among base-model failures that PEARL later solves, 73.8\% were originally {code execution} failures, followed by objective mismatch (12.6\%) and {formatting or extraction} errors (9.2\%). This supports the claim that the learned interaction policy is especially effective at implementation repair: PEARL learns to run partial code, inspect exceptions or solver messages, and revise the program until it reaches a valid execution path. It also repairs some modeling-level failures, but the dominant improvement comes from making the solver-grounded workflow operational.

Finally, the remaining PEARL failures show the limits of solver feedback as an oracle. Solver logs and compiler errors are highly useful for detecting runtime, feasibility, and formatting problems, but they do not fully identify all semantic mismatches between the intended optimization problem and the generated formulation. Objective-value checks catch many such mismatches when references are available, yet they remain partial: different formulations can agree on an objective for one instance but differ structurally, while ambiguous problem statements may admit multiple plausible interpretations. This motivates the broader validation agenda discussed in Appendix~\ref{app:limitations_impact}.

\section{Ablation Study and Sensitivity Analysis}
\label{app:ablation_sensitivity}

\subsection{Supervised Fine-Tuning Initialization}
\label{app:subsec:sft}

\subsubsection{Theoretical Analysis}
\label{app:subsubsec:sft_theory}
Alternatively, before agentic reinforcement learning, we can initialize the policy using supervised fine-tuning (SFT) to obtain a stable and competent reference model for subsequent multi-turn optimization. This initialization follows standard practice in optimization modeling and tool-augmented language modeling, and is not intended to encode interactive decision-making or solver-driven control behaviors.

\noindent\textbf{Training data.}
The SFT stage uses a mixture of publicly available optimization modeling datasets and curated instruction-style data, including OR-Instruct from ORLM~\citep{huang2025orlm}, OptMATH~\citep{lu2025optmath}, and OptiBench-style modeling instances~\citep{yang_optibench_2024}. These datasets consist of paired natural-language problem descriptions and corresponding solver-ready formulations or executable code, covering a broad range of optimization domains, constraint structures, and problem scales.

\noindent\textbf{Training objective.}
Supervised fine-tuning is performed using standard autoregressive next-token prediction over \emph{complete} solution traces, which may include natural-language reasoning, mathematical formulations, and solver code. Formally, let $\mathcal{D}_{\mathrm{SFT}}$ denote the supervised dataset of input--output pairs $(x, \mathbf{y})$, where $x$ is a natural-language optimization prompt and $\mathbf{y} = (y_1, \ldots, y_{|\mathbf{y}|})$ is the target output sequence. The SFT objective is defined as:
\begin{equation}
\mathcal{L}_{\mathrm{SFT}}(\theta)
=
- \mathbb{E}_{(x,\mathbf{y})\sim\mathcal{D}_{\mathrm{SFT}}}
\left[
\sum_{t=1}^{|\mathbf{y}|}
\log p_\theta(y_t \mid x, \mathbf{y}_{<t})
\right],
\label{eq:sft}
\end{equation}
where $\theta$ denotes the model parameters and $p_\theta(y_t \mid x, \mathbf{y}_{<t})$ is the conditional probability of generating the $t$-th token given the input prompt and all preceding tokens.

Importantly, SFT training is restricted to \emph{single-pass generation}: the model is not exposed to multi-turn interaction, intermediate solver feedback, partial execution results, or revision behavior during this stage.

\noindent\textbf{Role of SFT in PEARL.}
The purpose of SFT is to endow the base model with fundamental optimization modeling competence and syntactic validity, ensuring that generated formulations and solver code are executable at the outset of reinforcement learning. The resulting model defines the reference policy $\pi_{\mathrm{ref}}$, which serves solely as an initialization for agentic reinforcement learning. All interactive behaviors in PEARL—such as deciding when to invoke solvers, how to interpret solver diagnostics, and whether to revise or terminate—are learned exclusively during the agentic reinforcement learning stage described in Section~\ref{subsec:training}.

\subsubsection{Experiments}
\label{app:subsec:role_of_SFT}
\begin{figure}[h]
    \centering
    \includegraphics[width=0.5\textwidth]{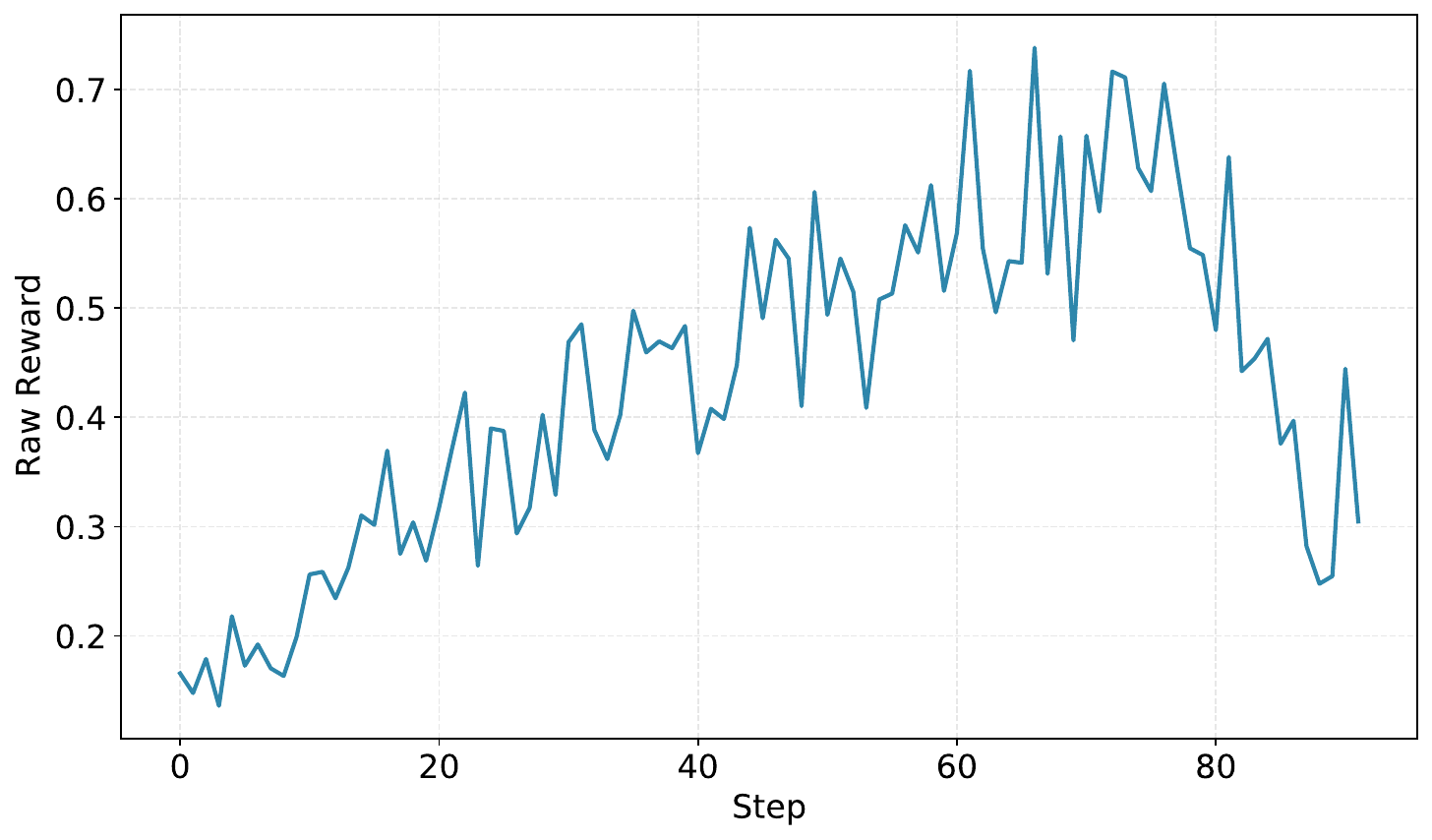}
    \caption{Training reward curve when directly applying PEARL to Qwen3-8B-Base without prior instruction tuning. The reward begins to decline after 60 steps, indicating training collapse due to the model's inability to consistently follow the required output format.}
    \label{fig:base_model_collapse}
\end{figure}

In our main experiments, we directly apply reinforcement learning to instruction-tuned models. However, when we attempted to train Qwen3-8B-Base (a base model without instruction tuning), we observed poor performance and training instability. As shown in Figure~\ref{fig:base_model_collapse}, the model struggles to follow the required output format in early training stages, leading to training collapse with rewards declining after approximately 60 steps. This suggests that for base models without instruction tuning, a supervised fine-tuning (SFT) warm-start is necessary to establish basic format-following capabilities.

Recall that, to obtain SFT data, we use a curated set of supervised solution traces. We conduct an ablation study to evaluate whether performing SFT before RL benefits the PEARL agent. As shown in Table~\ref{tab:sft_ablation}, SFT alone (SFT-Qwen-3-4B-Instruct) substantially improves the model's capabilities compared to the base model. However, the SFT-only model does not reach the performance ceiling of PEARL. Under this controlled ablation setting, adding PEARL after SFT improves MAMO-C from 55.86 to 76.57, while direct PEARL from the instruction-tuned model reaches 63.96 on MAMO-C but remains comparable in Macro/Micro average (70.40/80.38 versus 69.63/80.32 for PEARL after SFT). This appendix ablation is separate from the final main-result checkpoint reported in Table~\ref{tab:main_results_final}, where the PEARL MAMO-C result is 76.12$\pm$3.19. The results indicate that while SFT provides a useful warm-start for base models, instruction-tuned models can achieve similar final performance through direct RL training, bypassing the need for an intermediate SFT stage.

\begin{figure}[h]
    \centering
    \includegraphics[width=0.70\linewidth]{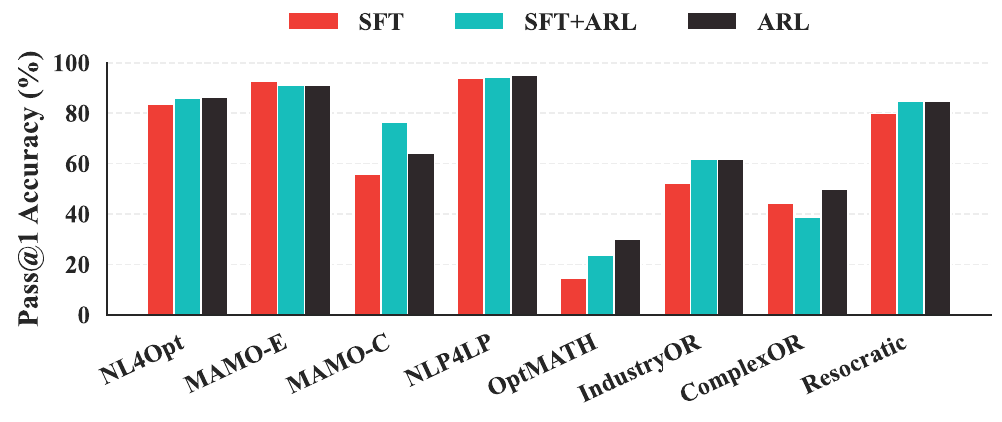}
    \caption{Analyzing the role of SFT prior to agentic RL. Results are obtained using the same base model, Qwen3-4B-Instruct, under three training regimes: (i) SFT only, (ii) SFT followed by agentic RL (SFT+ARL), and (iii) direct agentic RL without SFT (ARL).}
    \label{fig:sft_ablation}
\end{figure}

\begin{table}[h] 
 \centering 
 \caption{Ablation study on the role of supervised fine-tuning (SFT) before PEARL training. We compare three controlled ablation settings using Qwen3-4B-Instruct as the base: (1) SFT only, (2) SFT followed by PEARL training, and (3) direct PEARL training without SFT. This ablation is separate from the final main-result checkpoint in Table~\ref{tab:main_results_final}. Results show that while SFT provides substantial gains, the ultimate performance ceiling is similar whether SFT is used as an intermediate step or not.} 

 \label{tab:sft_ablation} 
 \resizebox{\linewidth}{!}{% 
 \begin{tabular}{l lcccccccccc} 
 \toprule 
 & \textbf{Method} 
 & \textbf{NL4Opt} 
 & \textbf{MAMO-E} 
 & \textbf{MAMO-C} 
 & \textbf{NLP4LP} 
 & \textbf{OptMATH} 
 & \textbf{IndustryOR} 
 & \textbf{ComplexOR} 
 & \textbf{Resocratic} 
 & Macro Avg 
 & Micro Avg\\ 
 \midrule 

  & SFT-Qwen-3-4B-Instruct
  & 83.64 & \textbf{92.66} & 55.86 & 93.82 & 14.46 & 52.38 & 44.44 & 79.90 & 64.64 & 76.86 \\ 
 & PEARL-SFT-Qwen-3-4B-Instruct
  & 85.98 & 91.19 & \textbf{76.57} & 94.38 & 23.49 & \textbf{61.90} & 38.89 & 84.62 & 69.63 & 80.32 \\ 
  & PEARL-Qwen-3-4B-Instruct
  & \textbf{86.45} & 91.01 & 63.96 & \textbf{94.93} & \textbf{30.12} & \textbf{61.90} & \textbf{50 }& \textbf{84.86} & \textbf{70.40} & \textbf{80.38} \\ 

 \bottomrule 
 \end{tabular}% 
 }% 
 \end{table}

\subsection{Adaptive Sampling with Trajectory Filtering}
\label{app:subsec:trajectory_filtering}

\subsubsection{Theoretical Analysis}
\label{app:subsec:trajectory_filtering_theory}
Agentic RL training, which incorporates external feedback from tool interactions, is highly susceptible to training instability and model collapse \citep{xue2026simpletir,shang_rstar2-agent_2025}. Tool-augmented rollouts may fail for reasons unrelated to modeling quality, including malformed tool calls, runtime crashes, and solver timeouts. Treating such failures as ordinary negative rewards introduces substantial variance and biases the policy update toward avoiding tool use altogether. To ensure stable training and achieve better model performance, prior work has adopted trajectory filtering strategies to exclude obviously anomalous negative samples \citep{xue2026simpletir,shang_rstar2-agent_2025}. To verify whether such filtering is effective in the optimization modeling scenario, we investigate a dynamic sampling strategy with trajectory filtering, inspired by DAPO-style filtering ideas \citep{yu_dapo_2025} but adapted to optimization modeling.

For each modeling instance $x_i$, we repeatedly sample trajectories $\tau_{i,j}$ from the current policy until a minimum number of valid trajectories is obtained.
Formally, let $\tau_{i,j}$ denote the $j$-th sampled trajectory for instance $x_i$, and let $q(\tau_{i,j}) \in \{0,1\}$ be a binary quality indicator that tests whether the trajectory is usable for learning.
We define the set of retained trajectories as
\begin{align}
\mathcal{V}_i
&=
\{\, j \le K_i \;:\; q(\tau_{i,j}) = 1 \,\}, \nonumber\\
K_i
&=
\min\!\left\{
k \;:\;
\sum_{j \le k} q(\tau_{i,j}) \ge K_{\min}
\right\}
\wedge K_{\max},
\end{align}
where $K_{\min}$ specifies the minimum number of valid trajectories required for instance $x_i$, and $K_{\max}$ caps the total number of rollouts to control computational cost.

The quality gate $q(\tau)$ filters out trajectories that are structurally invalid or uninformative, including malformed tool-call schemas, incomplete final outputs, or execution failures that occur before any meaningful solver interaction. By conditioning learning on $\mathcal{V}_i$ rather than on all sampled trajectories, this mechanism reduces gradient noise while allocating more sampling budget to difficult instances with low validity rates.

\begin{algorithm}[h]
\caption{\textsc{PEARL}-Filter: NL-to-Opt via Agentic RL with adaptive sampling}
\label{alg:pearl_filter}
\DontPrintSemicolon
\small

\KwIn{Dataset $\mathcal{D}$ of optimization modeling instances;
reference policy $\pi_{\mathrm{ref}}$;
$K_{\min}, K_{\max}$;
clipping thresholds $\epsilon_{+}, \epsilon_{-}$}

Initialize policy $\pi_\theta \leftarrow \pi_{\mathrm{ref}}$\;

\While{not converged}{
  \ForEach{$x_i \sim \mathcal{D}$}{
    $\mathcal{V}_i \leftarrow \emptyset$, $j \leftarrow 0$\;
    {\hfill{\footnotesize\color{gray}//Valid solver-reaching trajectories}}\;

    \While{$|\mathcal{V}_i| < K_{\min}$ \textbf{and} $j < K_{\max}$}{
      Sample trajectory $\tau_{i,j} \sim \pi_\theta(\cdot \mid x_i)$\;
      {\hfill{\footnotesize\color{gray}//Multi-turn NL modeling with execution and solver calls}}\;

      \If{$q(\tau_{i,j}) = 1$}{
        $\mathcal{V}_i \leftarrow \mathcal{V}_i \cup \{j\}$\;
        {\hfill{\footnotesize\color{gray}//Executable and reaches solver/validation}}\;
      }
      $j \leftarrow j + 1$\;
    }

    \ForEach{$j \in \mathcal{V}_i$}{
      Compute return $R(\tau_{i,j})$\;
      {\hfill{\footnotesize{{//Trajectory-level reward}}}}\;
    }

    Compute group-normalized advantages $A_{i,j}$\;
    {\hfill{\footnotesize\color{gray}//Instance-wise credit assignment}}\;
  }

  Update $\pi_\theta$ via decoupled-clipped policy gradient\;
  {\hfill{\footnotesize\color{gray}//Agentic RL over multi-turn optimization modeling}}\;
}
\end{algorithm}

\subsubsection{Experiments}
\label{app:subsec:results_trajectory_filtering}

As described above, we adopt an adaptive sampling strategy with trajectory filtering to mitigate training instability in agentic RL. Here, we empirically evaluate whether such filtering is necessary in the optimization modeling domain. We use Qwen3-4B-Instruct-2507 as the base model and train on a combination of curated and open-source samples. Due to the large context requirement with \texttt{max\_response\_length} set to 32k tokens, {we use} 8 {high-performance} GPUs with context parallelization set to 4.

\begin{figure}[h]
    \centering
    \begin{subfigure}[b]{0.48\textwidth}
        \centering
        \includegraphics[width=\textwidth]{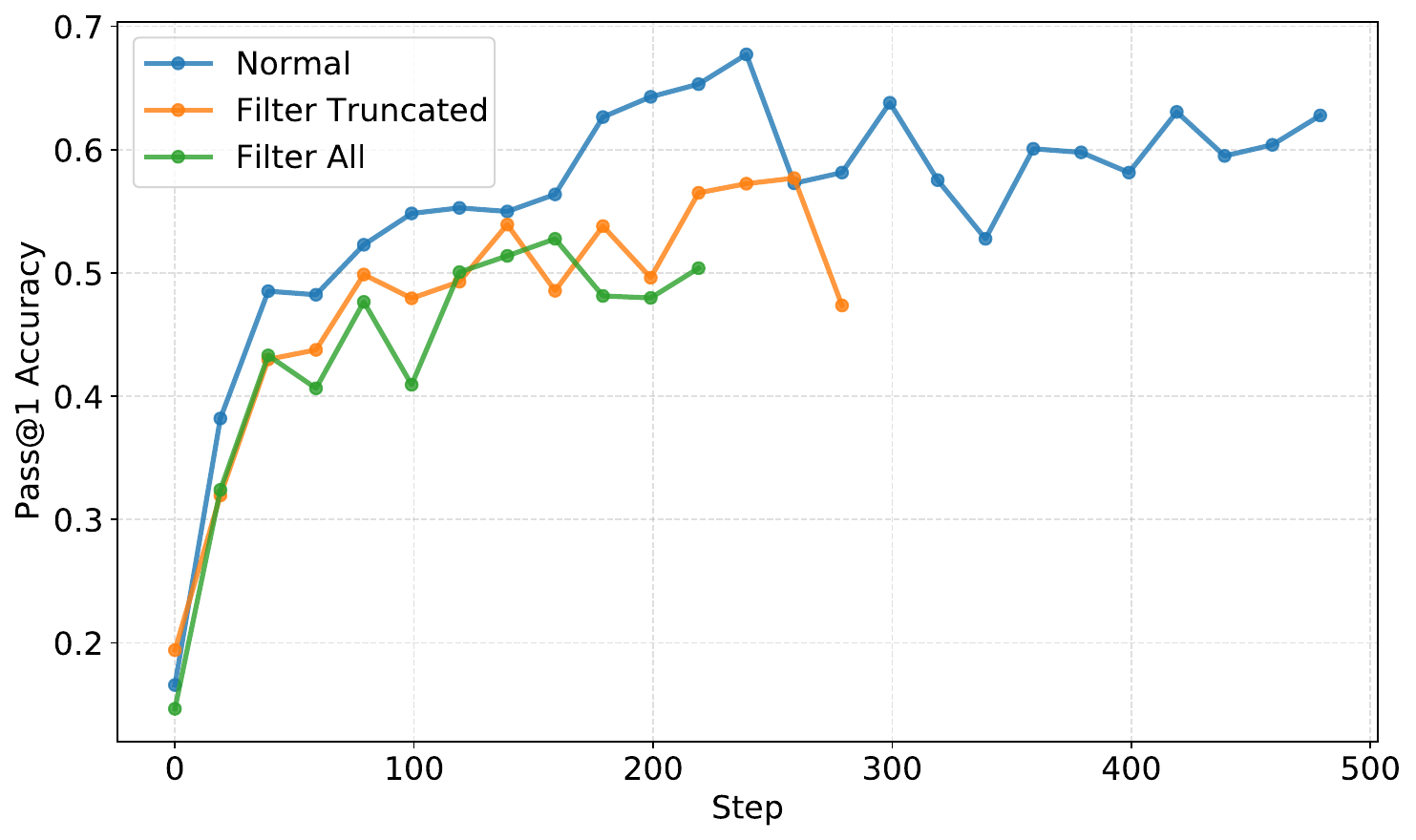}
        \caption{Pass@1 Accuracy (Avg. of MAMO-C \& IndustryOR)}
        \label{fig:filter_eval_avg}
    \end{subfigure}
    \hfill
    \begin{subfigure}[b]{0.48\textwidth}
        \centering
        \includegraphics[width=\textwidth]{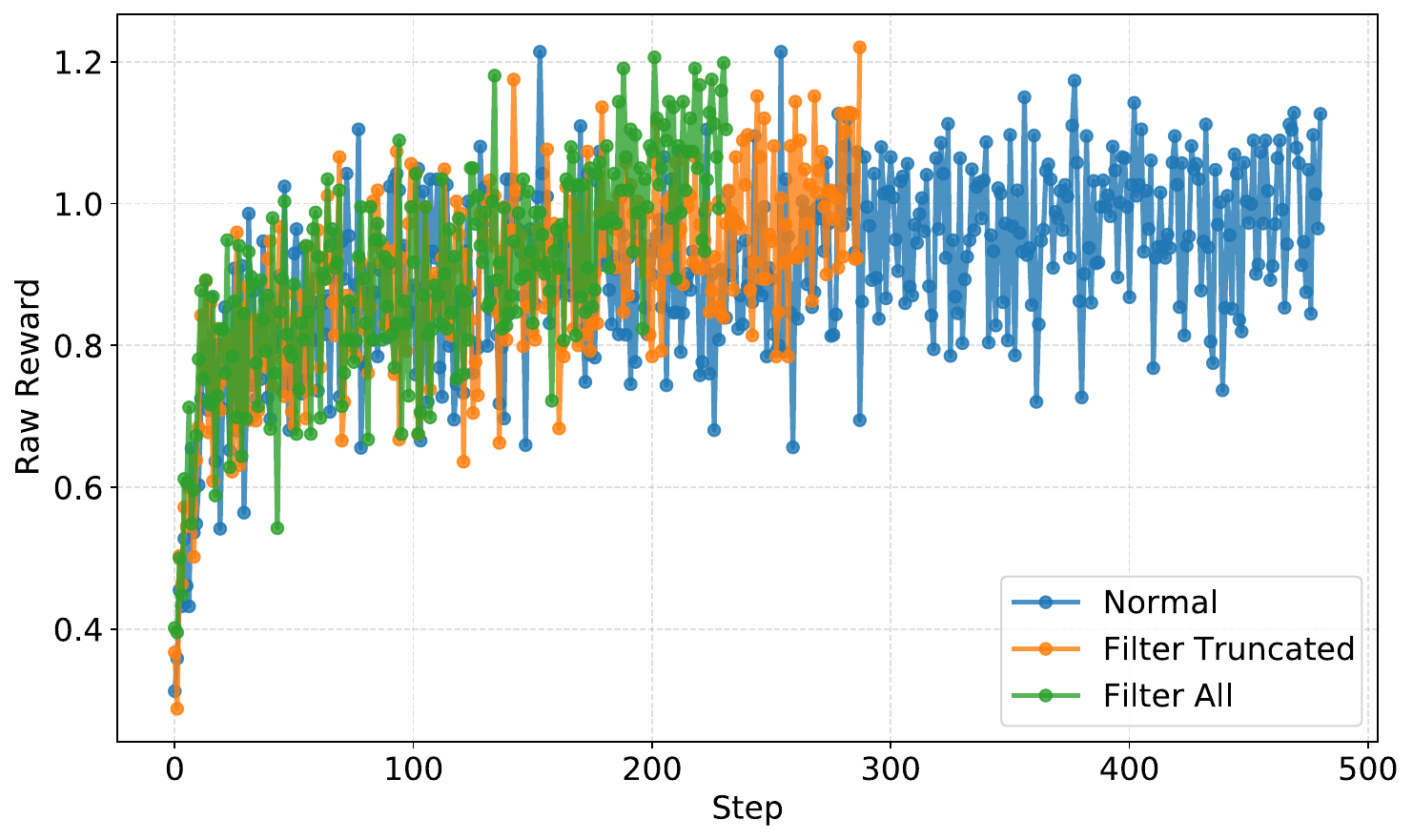}
        \caption{Raw Reward}
        \label{fig:filter_reward}
    \end{subfigure}

    \vspace{0.3cm}

    \begin{subfigure}[b]{0.48\textwidth}
        \centering
        \includegraphics[width=\textwidth]{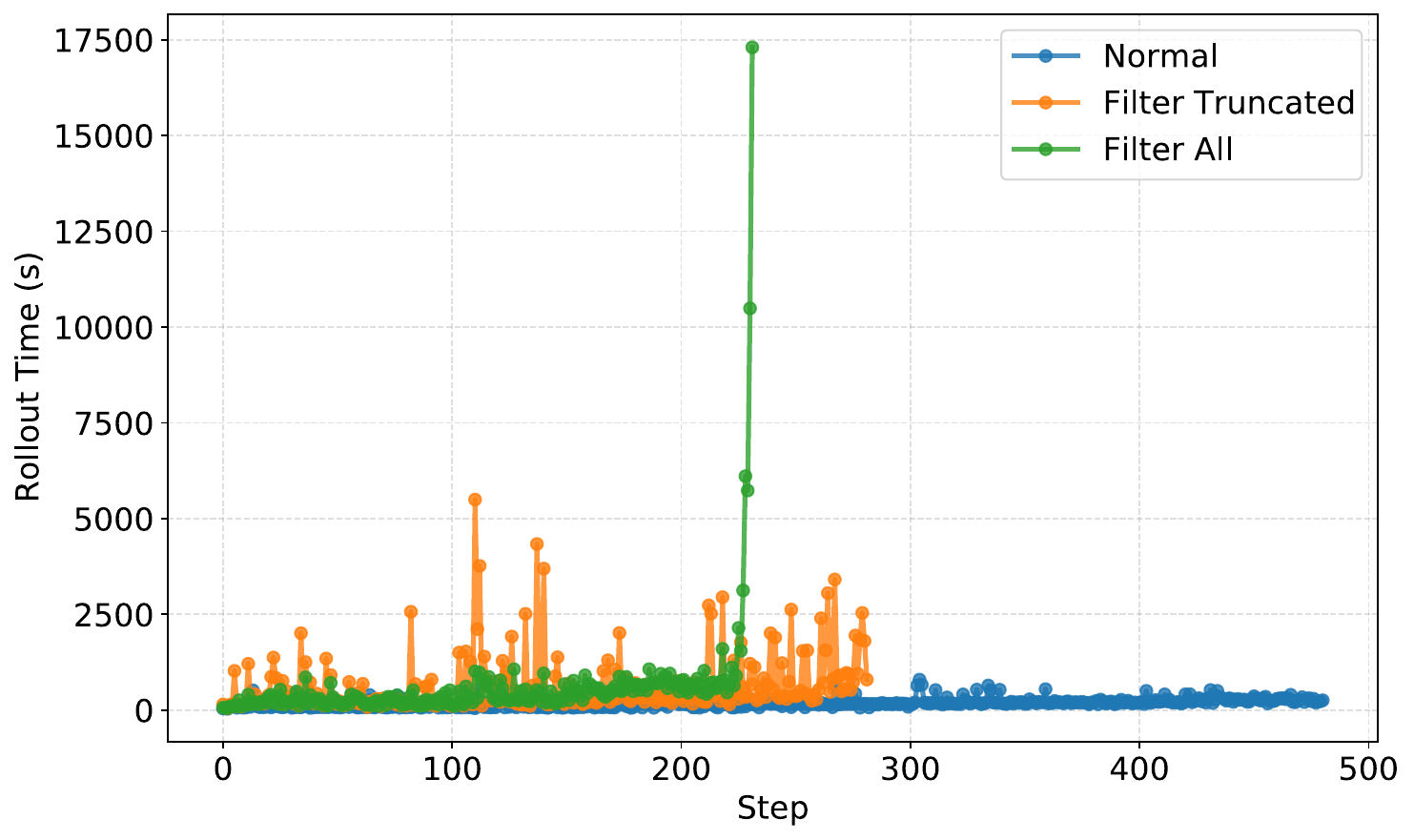}
        \caption{Rollout Time}
        \label{fig:filter_rollout_time}
    \end{subfigure}
    \hfill
    \begin{subfigure}[b]{0.48\textwidth}
        \centering
        \includegraphics[width=\textwidth]{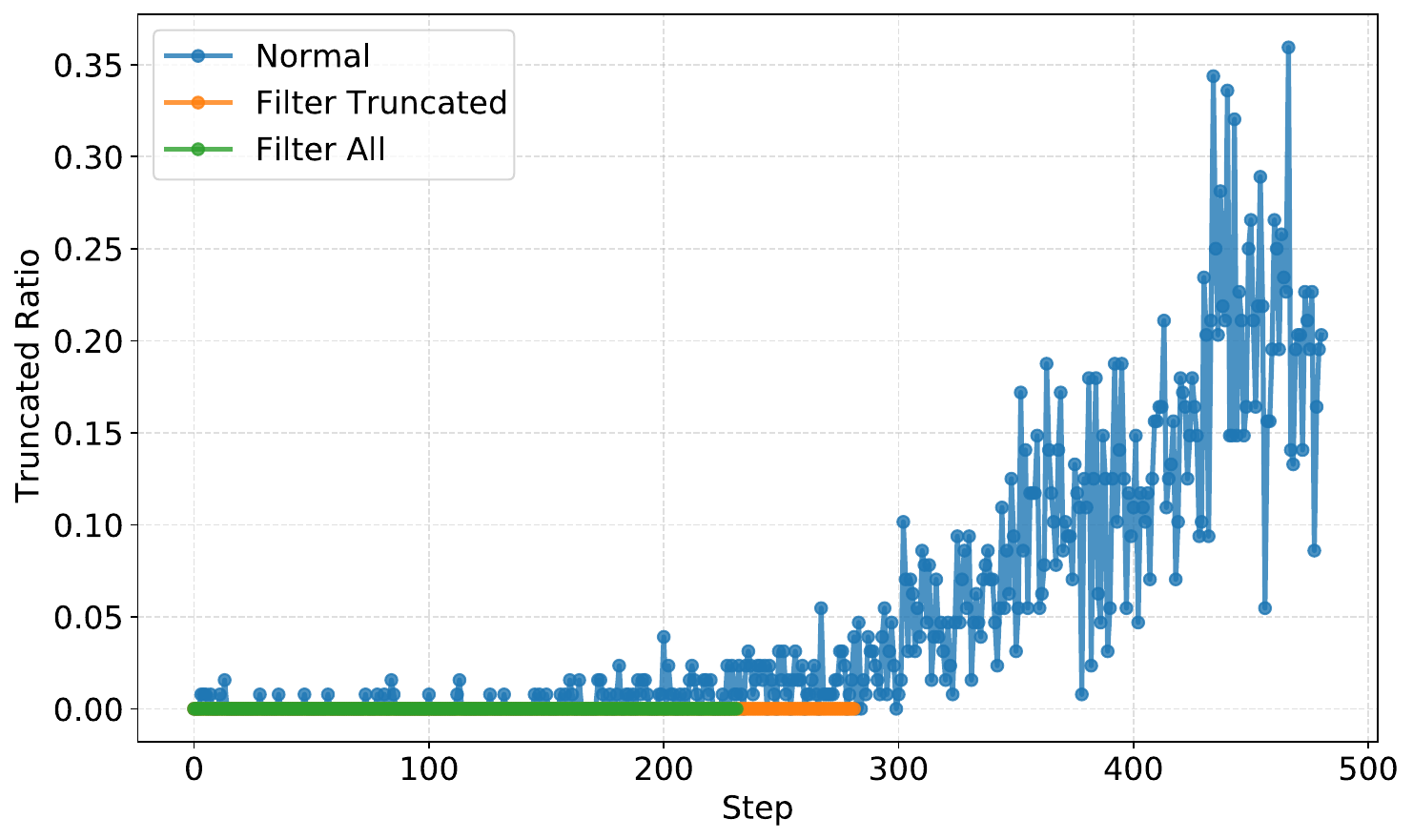}
        \caption{Truncated Ratio}
        \label{fig:filter_truncated_ratio}
    \end{subfigure}

    \caption{Impact of trajectory filtering strategies on PEARL training. We compare three settings: Normal (no filtering), Filter Truncated (filtering only truncated negative samples), and Filter All (filtering both truncated and format-violating negative samples). (a) Average Pass@1 accuracy on MAMO-C and IndustryOR benchmarks. (b) Training set raw reward. (c) Rollout time per step. (d) Truncated ratio during training.}
    \label{fig:trajectory_filtering}
\end{figure}

We explore three trajectory filtering strategies: (1) \textit{Normal}: no filtering of negative samples; (2) \textit{Filter Truncated}: filtering only truncated negative samples; and (3) \textit{Filter All}: filtering both truncated and format-violating negative samples. As shown in Figure~\ref{fig:trajectory_filtering}, under the Normal setting, the model's responses grow increasingly longer during training (reflected in extended rollout time in Figure~\ref{fig:filter_rollout_time}), leading to a progressively higher truncation ratio during the rollout process (Figure~\ref{fig:filter_truncated_ratio}). This results in more truncated negative samples being included in the training data.

Interestingly, despite the increasing presence of truncated negative samples, the model does not exhibit training collapse. In fact, as shown in Figure~\ref{fig:filter_eval_avg}, the \textit{Normal} setting achieves slightly better performance on the benchmarks (average of MAMO-C and IndustryOR) compared to the filtering strategies. {{The limited} number of tool calls (2-{-}4 per problem) {may help explain the stability observed in this setting. Training rewards also stabilize} ({Figure}~{\ref{fig:filter_reward}}). These {results} suggest that aggressive filtering {is unnecessary under the evaluated conditions}.}

\begin{figure}[h]
    \centering
    \includegraphics[width=0.70\linewidth]{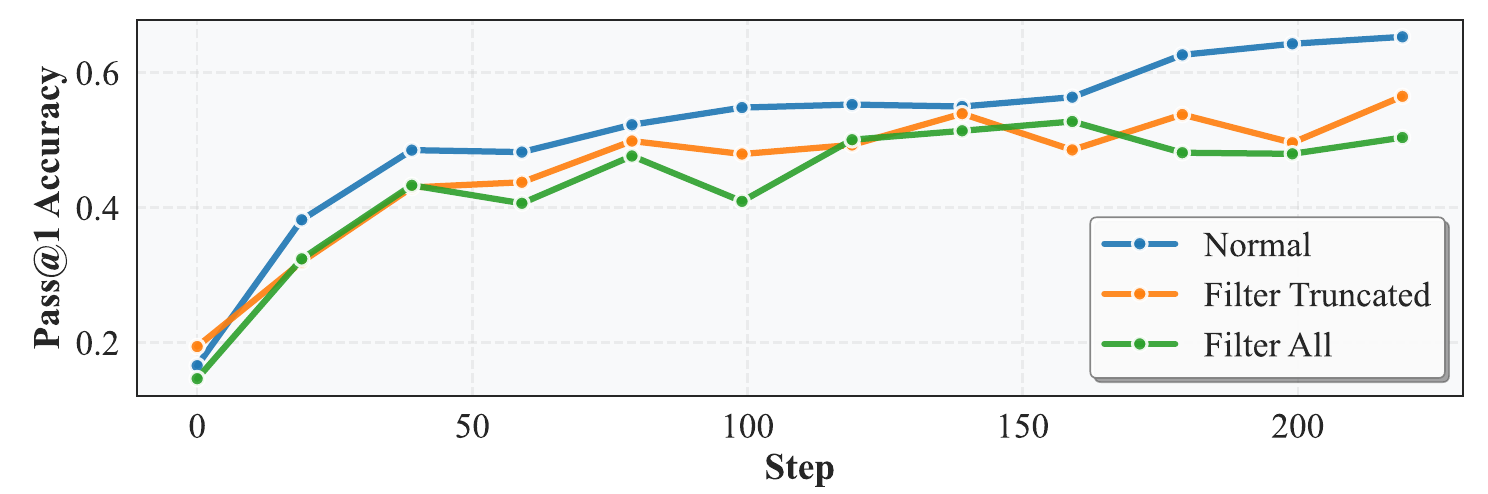}
    \caption{Impact of trajectory filtering strategies on PEARL training. We compare three settings---Normal, Filter Truncated, and Filter All---and report average Pass@1 accuracy on MAMO-C and IndustryOR throughout training.}
    \label{fig:trajectory_filtering_main}
\end{figure}

\subsection{Training Dynamics}
\label{app:subsec:training_dynamics}

\begin{figure}[h]
    \centering
    \begin{subfigure}[b]{0.48\textwidth}
        \centering
        \includegraphics[width=\textwidth]{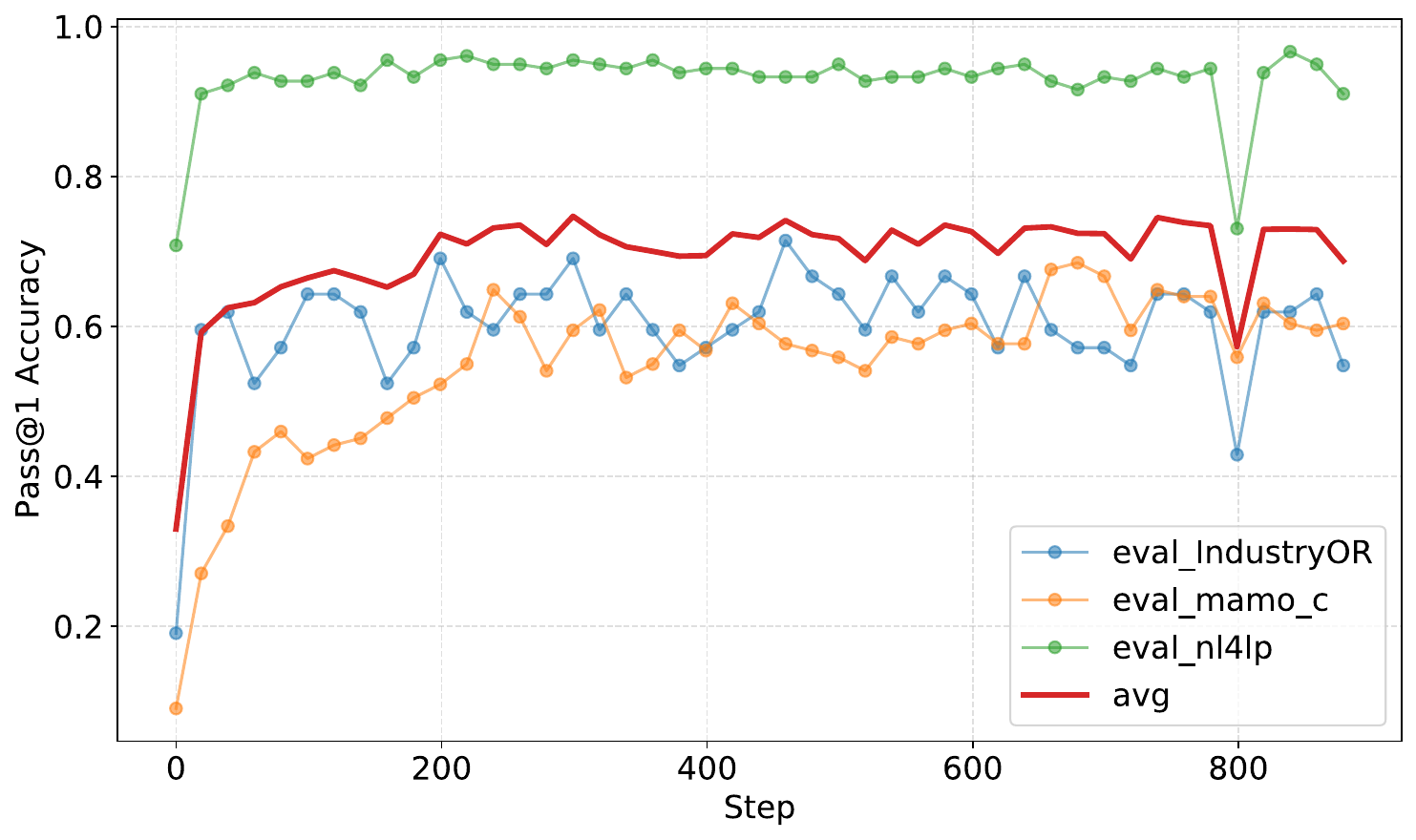}
        \caption{Pass@1 Accuracy on Three Benchmarks}
        \label{fig:training_accuracy}
    \end{subfigure}
    \hfill
    \begin{subfigure}[b]{0.48\textwidth}
        \centering
        \includegraphics[width=\textwidth]{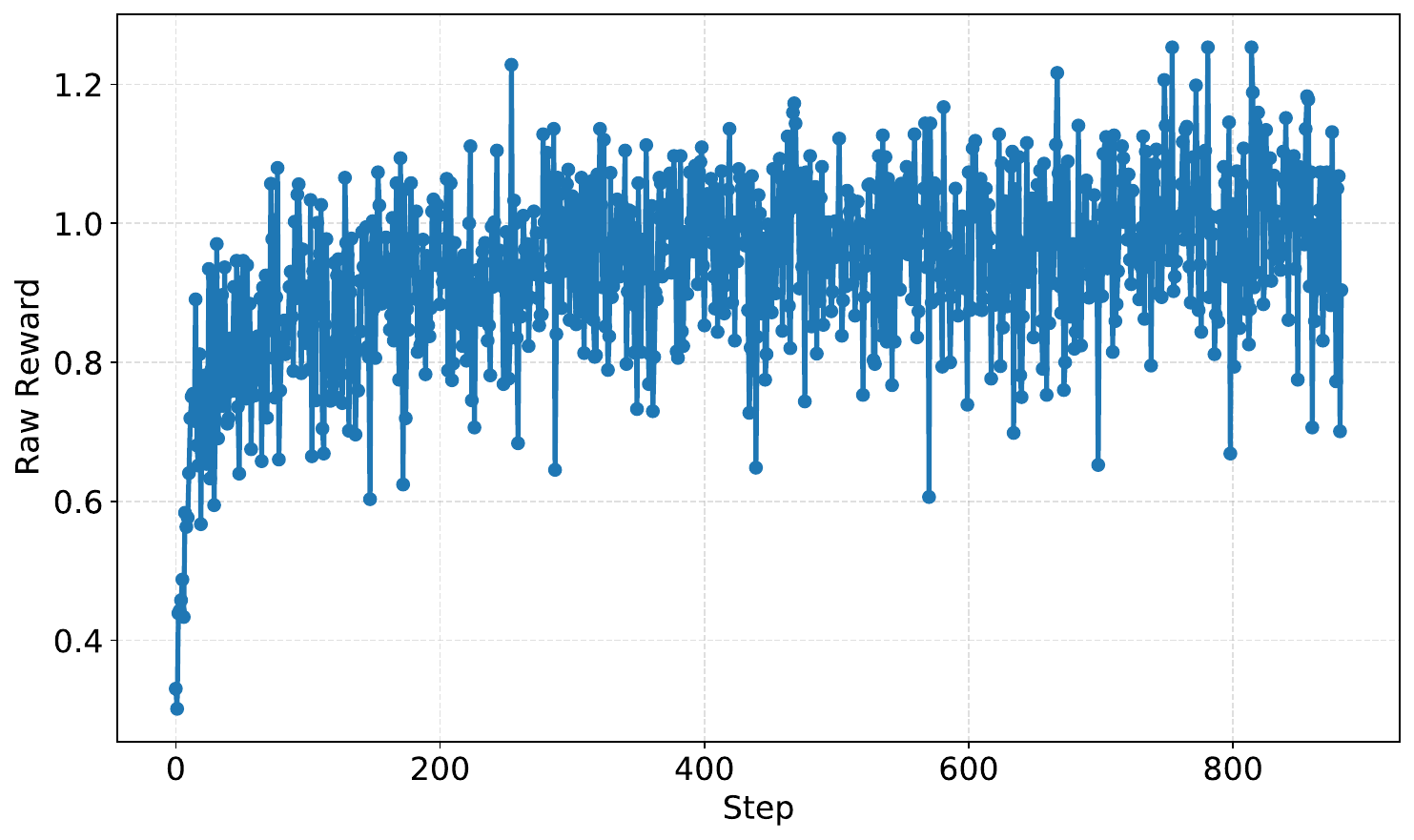}
        \caption{Raw Reward}
        \label{fig:training_reward}
    \end{subfigure}

    \vspace{0.3cm}

    \begin{subfigure}[b]{0.48\textwidth}
        \centering
        \includegraphics[width=\textwidth]{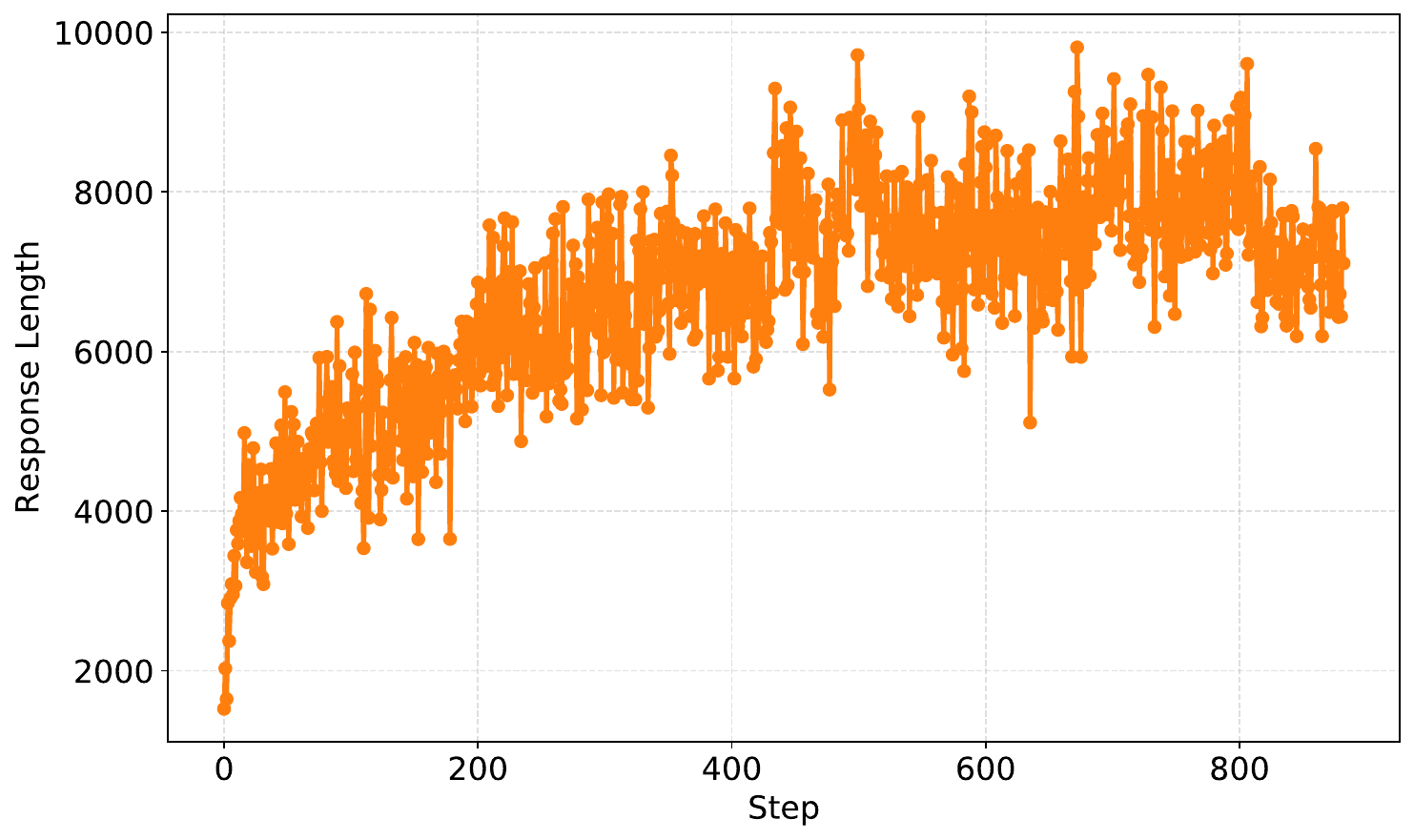}
        \caption{Response Length}
        \label{fig:training_response_length}
    \end{subfigure}
    \hfill
    \begin{subfigure}[b]{0.48\textwidth}
        \centering
        \includegraphics[width=\textwidth]{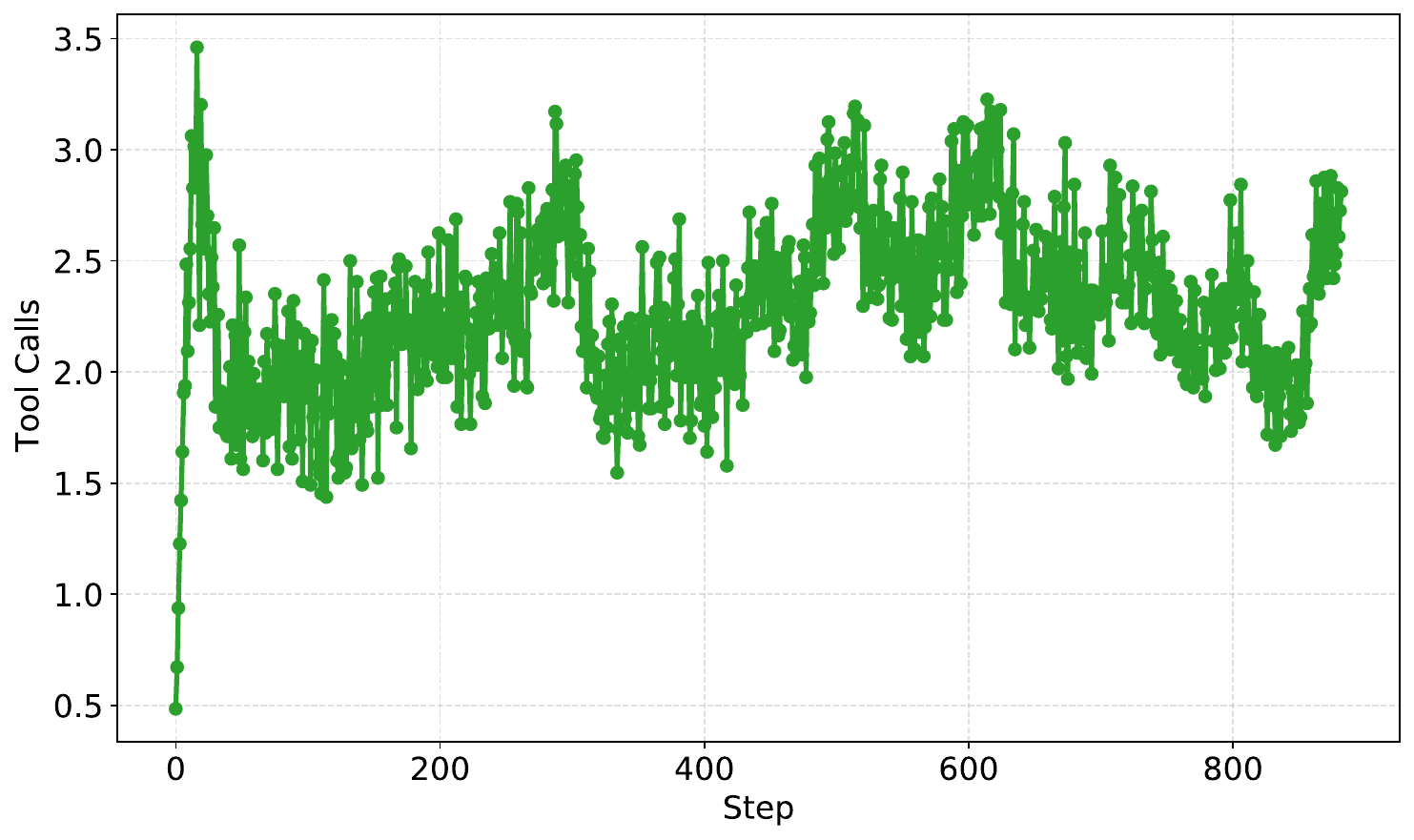}
        \caption{Tool Calls}
        \label{fig:training_tool_calls}
    \end{subfigure}

    \caption{Training dynamics of PEARL using Qwen3-4B-Instruct-2507 as the base model. The model is trained on open-source and curated data. (a) Pass@1 accuracy across three evaluation benchmarks (IndustryOR, MAMO-C, and NL4LP), showing consistent improvement during training. (b) Raw reward from the training set, demonstrating increasing performance that saturates in later stages. (c) Response length increases throughout training under GRPO optimization. (d) Tool calls rapidly converge, indicating the model quickly learns to utilize tools effectively{.} {{Online accuracy uses absolute objective error} ${\le 10^{-3}}${, without format bonuses}.}}
    \label{fig:training_dynamics}
\end{figure}

We visualize the training dynamics of PEARL in Figure~\ref{fig:training_dynamics}. The model is trained using Qwen3-4B-Instruct-2507 as the base model, with open-source and curated data as the training corpus. Throughout the training process, we monitor three key benchmarks (IndustryOR, MAMO-C, and NL4LP) as evaluation metrics. As shown in Figure~\ref{fig:training_accuracy}, the Pass@1 accuracy on all three benchmarks consistently increases during training, demonstrating the model's improved problem-solving capabilities.

The training set reward (Figure~\ref{fig:training_reward}) also exhibits an upward trend, though it begins to saturate in the later stages of training. {This {plateau} indicates {limited further improvement} in the {measured} training {reward}.} Notably, the model rapidly learns to utilize tool calls (Figure~\ref{fig:training_tool_calls}), with this metric converging early in the training process. This suggests that tool usage is one of the first capabilities the model acquires. Additionally, under GRPO training, we observe that the response length (Figure~\ref{fig:training_response_length}) continues to grow throughout the training process, reflecting the model's tendency to generate more detailed and comprehensive solutions.

\subsection{Tool-Use Depth and Termination}
\label{app:subsec:tool_use_depth}

\begin{table*}[h]
    \centering
    \caption{Scaffold-control comparison under the same PEARL inference process. Accuracy is pass@1 (\%); tool calls are averaged per instance.}
    \label{tab:scaffold_control_tool_calls}
    \begin{tabular}{lrrrr}
    \toprule
    \multirow{2}{*}{\textbf{Benchmark}}
    & \multicolumn{2}{c}{\textbf{Acc.}}
    & \multicolumn{2}{c}{\textbf{Tool Calls}} \\
    \cmidrule(lr){2-3}\cmidrule(lr){4-5}
    & \textbf{Base} & \textbf{PEARL} & \textbf{Base} & \textbf{PEARL} \\
    \midrule
    NL4Opt & 68.22 & 85.28 & 0.47 & 1.31 \\
    MAMO-Easy & 41.28 & 91.65 & 0.85 & 1.42 \\
    MAMO-Complex & 12.61 & 76.12 & 0.15 & 1.55 \\
    NLP4LP & 81.29 & 93.25 & 0.19 & 1.29 \\
    OptMATH & 5.42 & 16.87 & 0.23 & 4.43 \\
    IndustryOR & 19.05 & 67.86 & 0.69 & 1.62 \\
    ComplexOR & 11.11 & 41.66 & 0.28 & 2.28 \\
    Resocratic & 51.61 & 84.99 & 0.35 & 1.83 \\
    \bottomrule
    \end{tabular}%
\end{table*}

Average interaction depth alone is not sufficient evidence that PEARL benefits from a learned multi-turn policy: a fixed wrapper could also create additional opportunities for tool interaction. To isolate this effect, we evaluate the base Qwen3-4B-Instruct-2507 model under the same PEARL inference process, keeping the tool interface, execution-feedback loop, and repair budget fixed. For PEARL, we report the final checkpoint values from Table~\ref{tab:main_results_final} so that the accuracy column is aligned with the main result table. Table~\ref{tab:scaffold_control_tool_calls} reports both pass@1 accuracy and the average number of tool calls per instance.

The base model occasionally uses the available tools, but it does so much more sparsely and less effectively than trained PEARL. PEARL achieves higher accuracy on every benchmark, with especially large gains on MAMO-C, MAMO-E, IndustryOR, and Resocratic. These results suggest that the improvement is not due to the hand-written multi-turn scaffold alone. Rather, agentic training teaches a control policy for deciding when to execute code, how to interpret solver feedback, and whether to continue revising.

\begin{figure*}[h]
    \centering
    \includegraphics[width=0.70\textwidth]{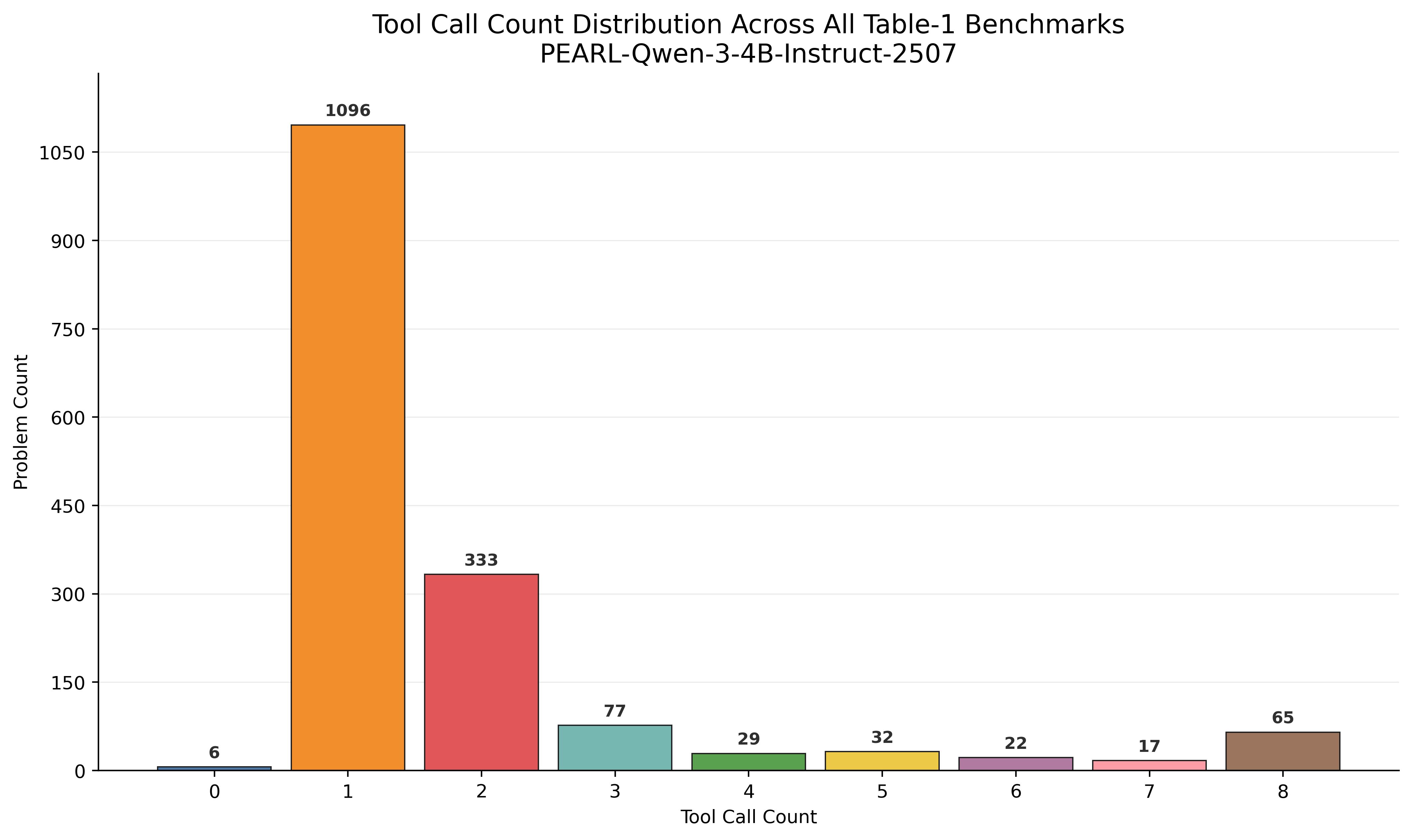}
    \caption{Distribution of tool-call counts for PEARL across the Table~\ref{tab:main_results_final} benchmarks. {{Counts are concentrated at} one or two calls{. The eight-turn budget includes final-answer generation}, {so the eight-call bin must not be interpreted as eight calls followed by an additional final-answer turn}.}}
    \label{fig:tool_call_count_distribution}
\end{figure*}

{The eight-turn inference budget {is} an upper bound {on generation turns, not} a target {number of tool calls}. {A valid} final \texttt{<answer>} action {permits early termination and consumes one turn}. {The inference implementation can also extract fallback code when the budget is exhausted, without generating an additional answer.} Figure~\ref{fig:tool_call_count_distribution} {shows a concentration} at one or two calls {and} a tail {extending to eight. Counts alone do not distinguish successful final-answer termination from budget exhaustion or fallback extraction}.} The benchmark-level averages in Table~\ref{tab:scaffold_control_tool_calls} similarly remain small for most datasets, rising most noticeably on OptMATH.

{{Figure}~{\ref{fig:rq2_training_choices}(a) and Figure}~{\ref{fig:training_tool_calls} show tool use emerging early and stabilizing around 2--4 calls, while the training reward in Figure}~{\ref{fig:training_reward} levels off later in training. Together} with {the concentration of evaluation trajectories at one or two tool calls in Figure}~{\ref{fig:tool_call_count_distribution}, these observations support the emergence of early} termination {behavior within the allowed} budget.}

\newpage

\section{PEARL Rollout Example}
\label{app:case_study}

\begin{tcolorbox}[
  breakable,
  title={PEARL Agent Response (Sanitized Excerpt)},
  enhanced,
  sharp corners,
  boxrule=0.4mm,
  colback=blue!1,
  colframe=blue!60!black,
  colbacktitle=blue!8,
  coltitle=black,
  boxed title style={sharp corners, boxrule=0.4mm, colframe=blue!60!black, colback=blue!8},
  fonttitle=\bfseries
]

\textit{We are given a complex production system with multiple stages. Let's model this as a \textbf{mixed-integer linear program}.}

\vspace{0.2cm}
\noindent\textbf{Decision Variables:}
\begin{itemize}[leftmargin=*,itemsep=1pt]
\item $R$: raw material units; $e_{\text{sold}}, e_{\text{toT}}, e_{\text{toBS}}$: engine block allocation
\item $t_{\text{sold}}, t_{\text{toBS}}$: transmission allocation; $b_{\text{sold}}$: brake systems sold
\end{itemize}

\noindent\textbf{Objective:} Maximize profit: $25e_{\text{sold}} + 35t_{\text{sold}} + 45b_{\text{sold}} - 68R - 5e_{\text{toT}} - 10e_{\text{toBS}} - 12t_{\text{toBS}}$

\noindent\textbf{Key Constraints:} Engine/transmission/brake system material balance, contract requirements ($b_{\text{sold}} \geq 4000$), market demand caps, machine time limit.

\vspace{0.2cm}
\noindent\textit{Now implementing in Pyomo...}

\vspace{0.1cm}
\noindent\colorbox{green!15}{\texttt{\textbf{<code>}}} \texttt{from pyomo.environ import *; model = ConcreteModel(); ...}

\noindent\colorbox{red!15}{\texttt{\textbf{<interpreter>}}} \textcolor{red!70!black}{\texttt{Error: Constraint 'eb\_balance[1]' does not have proper value. Found function object instead of expression.}}

\vspace{0.1cm}
\noindent\textcolor{blue!70!black}{\textbf{$\triangleright$ Reflection:}} \textit{Ah, I see the issue! I'm passing the function object instead of calling it. Let me fix the constraint definition...}

\vspace{0.1cm}
\noindent\colorbox{green!15}{\texttt{\textbf{<code>}}} \texttt{model.eb\_balance.add(eb\_balance\_rule(model)) \# Fixed: call the rule}

\noindent\colorbox{green!15}{\texttt{\textbf{<interpreter>}}} \texttt{Optimal Solution: R=3772.0, e\_sold=20000.0, e\_toT=2399.0, ...}  
\texttt{Maximum Profit: 851534.0}

\vspace{0.1cm}
\noindent\textcolor{blue!70!black}{\textbf{$\triangleright$ Reflection:}} \textit{We're getting non-integer values like 3772.0. However, these are actually integers displayed with .0 suffix. Let's verify the solution satisfies integrality and all constraints...}

\vspace{0.1cm}
\noindent\textit{Checking: $R=3772$ (integer $\checkmark$), $e_{\text{toT}}=2399$ (integer $\checkmark$), all constraints satisfied $\checkmark$}

\vspace{0.2cm}
\noindent\colorbox{orange!15}{\texttt{\textbf{<answer>}}}

\begin{tcolorbox}[
  colback=orange!3,
  colframe=orange!60!black,
  boxrule=0.3mm,
  sharp corners,
  left=2mm,
  right=2mm,
  top=1mm,
  bottom=1mm
]
\textbf{Modeling:} 7 decision variables (all $\mathbb{Z}^+$), 1 objective (maximize profit), 6 constraint groups (material balance, contract, demand, machine time)

\vspace{0.1cm}
\textbf{Implementation:} Complete Pyomo code with constraint rules, GLPK solver
\begin{lstlisting}[basicstyle=\ttfamily\scriptsize,breaklines=true,showstringspaces=false]
from pyomo.environ import *
model = ConcreteModel()
# Variables
model.R = Var(domain=NonNegativeIntegers)
model.e_sold = Var(domain=NonNegativeIntegers)
...
# Constraints
model.eb_balance = ConstraintList()
model.eb_balance.add(eb_balance_rule(model))
...
# Solve
solver = SolverFactory('glpk')
results = solver.solve(model)
print(f"Maximum Profit: {value(model.objective)}")
\end{lstlisting}
\end{tcolorbox}

\end{tcolorbox}

\begin{tcolorbox}[
  breakable,
  title={Problem Description},
  enhanced,
  sharp corners,
  boxrule=0.4mm,
  colback=blue!1,
  colframe=blue!60!black,
  colbacktitle=blue!8,
  coltitle=black,
  boxed title style={sharp corners, boxrule=0.4mm, colframe=blue!60!black, colback=blue!8},
  fonttitle=\bfseries
]
\begin{Verbatim}[breaklines,fontsize=\footnotesize]
AutoParts Ltd. manufactures three types of automotive components: Engine Block, Transmission, and Brake System. The company sources raw materials at a cost of $60 per unit. Each unit of raw material can be processed into 7 units of Engine Block and 3 units of Transmission. Processing one unit of raw material requires $8 in energy costs and 7 hours of machine time.\n\nEngine Block can be used in three ways:\n- Sold directly for $25 per unit.\n- Processed into 1 unit of Transmission, requiring 4 hours of machine time and $5 in additional costs.\n- Processed into 1 unit of Brake System, requiring 6 hours of machine time and $10 in additional costs.\n\nTransmission can be used in two ways:\n- Sold directly for $35 per unit.\n- Processed into 1 unit of Brake System, requiring 5 hours of machine time and $12 in additional costs.\n\nBrake System is sold for $45 per unit. The company has a maximum machine time capacity of 60,000 hours per year. Additionally, there is a contractual obligation to produce at least 4,000 units of Brake System annually. The maximum market demand for each component is as follows:\n- Engine Block: 20,000 units\n- Transmission: 15,000 units\n- Brake System: 12,000 units\n\nDetermine how AutoParts Ltd. can maximize its annual profit while meeting all constraints.
\end{Verbatim}

\end{tcolorbox}

\section{Prompt Templates}
\label{app:prompt_templates}

\newtcolorbox{prompttemplate}[1]{
  title={#1},
  breakable,
  boxrule=0.5mm,
  colback=white,
  colframe=black,
  fonttitle=\bfseries,
    listing options={
    upquote=true, 
    basicstyle=\ttfamily, 
    columns=fullflexible 
  }
}

\subsection{Training Prompt}
\label{app:training_prompt}

\begin{prompttemplate}{RL Training Prompt (System, Pyomo)}
\begin{Verbatim}[breaklines,breakanywhere]
You are an optimization modeling agent.

Goal:
- Convert the user problem statement into a correct optimization model and a runnable Python implementation.
- Use Python tool calls to compute, sanity-check, and verify your formulation when needed.

- The final code must be verified runnable, and you must validate the optimal value/solution against the problem statement: check all constraints are satisfied and no requirement is violated.

Modeling notes:
- Check variable meaning: if x is a quantity (integer/continuous), do NOT use binary-only logic like x_i + x_j <= 1.
- For logical constraints about whether a product is produced (x_i > 0), introduce indicator y_i in {0,1}.
- Link quantity and indicator with tight bounds: x_i <= M_i * y_i; add x_i >= y_i only if selecting implies x_i >= 1.
- Encode implications with indicators: (A => B) becomes y_A <= y_B; (A => not B) becomes y_A + y_B <= 1.
- Choose M_i as small as possible using known bounds (e.g., from resource constraints) to keep the MIP stable.
- Carefully distinguish variable types. Common nouns -> types: amount/production/flow/distance/time/cost = nonnegative continuous; counts (trips/machines/workers/vehicles/orders/batches) = nonnegative integer; yes/no decisions (select/open/visit/build) = binary; assignment/matching/edge-used (i->j) = binary; sequence/rank = integer (with subtour-elimination / uniqueness constraints).
- Keep units consistent across constraints (e.g., hours vs. units, kg vs. g); unit mistakes often cause 10x/1000x objective gaps.
- Implementation sanity: check solver termination is optimal before reading values; access indexed vars as model.x[i] (not getattr).

Implementation requirement:
- Use Pyomo to implement and solve the optimization model. Your code should follow this skeleton:
from pyomo.environ import *

model = ConcreteModel()

# Solve
You may use SolverFactory('glpk') for linear/MIP problems.
You may use SolverFactory('ipopt') or SolverFactory('gurobi') for nonlinear programming (NLP).
solver = SolverFactory('glpk')
results = solver.solve(model)

Tool calling:
- When you need to run Python, output exactly:
<code>
...python code...
</code>
- After a tool call, you will receive the execution result inside:
<interpreter>
...tool output...
</interpreter>
- You may call the tool multiple times.

Final answer:
- When you are done, output exactly one final block:
<answer>
# Modeling: 
define decision variables, objective, constraints (clear and complete).
# Implementation: 
include runnable code wrapped as:
'''python
...python code...
'''
</answer>
\end{Verbatim}
\end{prompttemplate}

\subsection{Direct Prompts}
\label{app:direct_prompt}

\begin{prompttemplate}{Direct Testing Prompt (Pyomo)}
\begin{Verbatim}[breaklines,breakanywhere]
You are a professional mathematical optimization expert. Please use Python and Pyomo to solve the following optimization problem.

# Question:
{description}

# Note:
- Use Pyomo to implement and solve the optimization model. Your code should follow this skeleton:
```python
import pyomo.environ as pyo

model = pyo.ConcreteModel()

solver = pyo.SolverFactory("glpk")
results = solver.solve(model)
```
- Make sure the model variable is named `model`.
- Carefully determine whether the variable is an integer or a continuous variable.
\end{Verbatim}
\end{prompttemplate}

\section{{Training Data and Benchmark Documentation}}
\label{app:data_collection}

{\paragraph{{Training data.}}
{The corpus contains approximately 13.8k underlying problems: about 10k cleaned public instances from OR-Instruct/ORLM}~{\citep{huang2025orlm}, OptMATH}~{\citep{lu2025optmath}, and OptiBench-style resources}~{\citep{yang_optibench_2024}, plus about 3.8k expert-curated instances. Optimization experts collected the curated problems from publicly available operations-research articles and websites. DeepSeek-V3.2 assisted with rewriting and diversifying descriptions, followed by expert formulation checks, manual curation, and solver verification. All retained training instances have valid reference objective values. These counts describe the problem pool; multi-turn trajectories are generated online during RL.}}

\begin{table}[htbp]

\centering

\caption{{Training problem pools and documented preparation steps. Counts refer to underlying problems, not online RL trajectories.}}
\label{tab:training_data_pools}
\small
\setlength{\tabcolsep}{4pt}
\renewcommand{\arraystretch}{1.1}
\begin{tabularx}{\linewidth}{@{}>{\raggedright\arraybackslash\leavevmode}p{0.22\linewidth}>{\centering\arraybackslash\leavevmode}p{0.15\linewidth}>{\raggedright\arraybackslash\leavevmode}X@{}}
\toprule
\textbf{{Problem pool}} & \textbf{{Approx. count}} & \textbf{{Sources and preparation}}\\
\midrule
{Cleaned public data} & {10k} & {OR-Instruct/ORLM, OptMATH, and OptiBench-style resources; cleaning and reference-objective verification.}\\
\addlinespace[0.55em]
{Expert-curated data} & {3.8k} & {Public OR articles and websites; LLM-assisted description rewriting, expert formulation checks, filtering, and~solver~verification.}\\
\bottomrule
\end{tabularx}
\end{table}

{\paragraph{{Evaluation data.}}
{For NL4Opt, MAMO, NLP4LP, IndustryOR, ComplexOR, and Resocratic, we use cleaned public benchmark resources associated with the LLM4OR survey}~{\citep{ijcai2025p1192}. OptMATH-Bench is the official 166-instance ICML-2025 test release}~{\citep{lu2025optmath}. Its release file is} \texttt{{benchmark/OptMATH\_Bench.json}} {in \url{https://github.com/optsuite/OptMATH}. Table}~{\ref{tab:benchmark_counts} lists the eight evaluation sets and their JSONL copies in the PEARL repository's} \texttt{{benchmark/}} {directory. Macro accuracy averages their scores; micro accuracy weights them by the instance counts.}}

\begin{table}[htbp]

\centering

\caption{{Benchmark instance counts used in the evaluation.}}
\label{tab:benchmark_counts}
\small
\begin{tabular}{lrl}
\toprule
{Benchmark} & {Instances} & {Released PEARL file}\\
\midrule
{NL4Opt} & {214} & \texttt{{nl4opt.jsonl}}\\
{MAMO EasyLP} & {545} & \texttt{{mamo\_easy\_lp\_clean.jsonl}}\\
{MAMO ComplexLP} & {111} & \texttt{{mamo\_complex.jsonl}}\\
{NLP4LP} & {178} & \texttt{{nl4lp.jsonl}}\\
{OptMATH-Bench} & {166} & \texttt{{OptMATH\_Bench.jsonl}}\\
{IndustryOR} & {42} & \texttt{{IndustryOR.jsonl}}\\
{ComplexOR} & {18} & \texttt{{complexor.jsonl}}\\
{Resocratic} & {403} & \texttt{{ReSocratic.jsonl}}\\
\midrule
{Total} & {1,677} & \\
\bottomrule
\end{tabular}
\end{table}

{\paragraph{{Data ablation.}}
{Table}~{\ref{tab:data_ablation} compares public-only, curated-only, and combined-data training with Qwen3-4B. Under the same PEARL training and evaluation protocol, adding the curated data improves macro- and micro-averaged Pass@1 by 6.28 and 1.43 percentage points, respectively, relative to public-only training.}}

\section{{Additional Experiments and Protocols}}
\label{app:comparison_protocols}

{This appendix provides additional comparisons and analyses. Appendix}~{\ref{app:sirl_retraining} compares PEARL with SIRL using the same backbone and training data, and also evaluates the released SIRL checkpoint with Gurobi. Appendix}~{\ref{app:data_token_usage} reports training-data ablations and generated-token usage. Appendix}~{\ref{app:workflow_scaffold_transfer} tests Qwen3-4B in the LLMOPT inference workflow without LLMOPT-specific fine-tuning and discusses comparisons with other workflow-based systems.}

{\subsection{{SIRL retraining on PEARL's training data}}
\label{app:sirl_retraining}
{To test whether additional training data alone explain PEARL's gains, we retrain SIRL on PEARL's 13.8k training instances using SIRL's released training pipeline. Both methods start from Qwen2.5-7B-Instruct and generate Pyomo code. For SIRL, we convert the data into the required problem--answer format without adding instances. We evaluate both models on the same eight benchmarks, with the same evaluation prompt, objective-value check, and tolerance.}}

\begin{table}[!htbp]

\centering

\caption{{Qwen2.5-7B comparisons (Pass@1, \%). Official SIRL uses its native Gurobi API and original training data; retrained SIRL and PEARL use Pyomo and the same 13.8k problems. Evaluation shares benchmark instances, validator, tolerance, and a common prompt template with interface-specific instructions. Results are point estimates, reported separately from Table}~{\ref{tab:main_results_final}.}}
\label{tab:sirl_controls}
\small
\begin{tabular}{lrrrr}
\toprule
{Benchmark} & ${n}$ & {PEARL} & {Official SIRL} & {Retrained SIRL}\\
 & & {Pyomo} & {Gurobi} & {Pyomo}\\
\midrule
{NL4Opt} & {214} & {90.19} & {87.85} & {84.51}\\
{MAMO-E} & {545} & {94.13} & {96.15} & {94.50}\\
{MAMO-C} & {111} & {87.39} & {79.28} & {44.14}\\
{NLP4LP} & {178} & {91.01} & {94.38} & {89.33}\\
{OptMATH-Bench} & {166} & {3.61} & {27.71} & {1.20}\\
{IndustryOR} & {42} & {57.14} & {35.71} & {42.86}\\
{ComplexOR} & {18} & {44.44} & {27.78} & {27.78}\\
{Resocratic} & {403} & {74.94} & {85.86} & {73.45}\\
\midrule
{Macro} & {--} & {67.86} & {66.84} & {57.22}\\
{Micro} & {1,677} & {77.82} & {82.29} & {73.03}\\
\bottomrule
\end{tabular}
\end{table}

{{Table}~{\ref{tab:sirl_controls} shows that PEARL reaches 67.86\% macro-averaged and 77.82\% micro-averaged Pass@1, compared with 57.22\% and 73.03\% for retrained SIRL. These are gains of 10.64 and 4.79 percentage points, respectively, with PEARL performing better on seven of eight benchmarks. Since both models use the same backbone and training data, the result supports a benefit from PEARL's training approach beyond access to additional data.}}

{{The two methods use the same training instances but learn differently: SIRL assigns terminal rewards to complete responses, whereas PEARL trains on multi-turn sequences of tool calls, feedback, and revisions. The comparison matches the backbone, training data, and modeling interface, but does not match training compute or hyperparameter-tuning effort.}}

{{Table}~{\ref{tab:sirl_controls} also includes the released SIRL checkpoint evaluated with Gurobi. We retain our evaluation-prompt template rather than SIRL's original prompt. The released checkpoint achieves 66.84\% macro-averaged and 82.29\% micro-averaged Pass@1. It was trained on a different dataset, which was not fully available to us, so it is a separate comparison rather than the data-matched baseline above. Table}~{\ref{tab:main_results_final} reports its results with Pyomo instead of Gurobi.}}

{{Retrained SIRL performs worse than the released checkpoint, but the two use different training data and modeling interfaces. Training budget, checkpoint selection, and hyperparameter tuning may also contribute to the difference. Further controlled experiments are needed to separate these effects.}}

{\subsection{{Data ablation and generated-token usage}}
\label{app:data_token_usage}
\noindent\textbf{{Training-data ablation.}} {We train PEARL with Qwen3-4B on three datasets: the cleaned public 10k instances, the curated 3.8k instances, and their combination. The data sources are described in Appendix}~{\ref{app:data_collection}. We keep the training and evaluation settings fixed. Table}~{\ref{tab:data_ablation} shows that adding the curated data to the public set raises macro-averaged Pass@1 from 64.12\% to 70.40\% and micro-averaged Pass@1 from 78.95\% to 80.38\%. The combined set gives the best results, showing that the curated instances are useful additions to the public data.}}

{\noindent\textbf{{Generated-token usage.}} {In a separate experiment, we evaluate PEARL, DeepSeek-V3.2, and DeepSeek-V3.2-Think on the same instances. All methods use the same solver, code parser, and objective-value check, but follow their respective inference procedures. PEARL averages 9,868.75 generated tokens per instance, compared with 1,465.51 for DeepSeek-V3.2 and 2,323.56 for DeepSeek-V3.2-Think (Table}~{\ref{tab:token_cost}). For PEARL, the count includes all model responses across the interaction, not just the final answer.}}

{{These counts measure how much text the models generate, not how much they cost to run. PEARL's 4B backbone supports local deployment and potentially lower per-token serving costs}~{\citep{abdin2024phi3,belcak2025small}, which can help offset the additional generation required for multi-turn modeling. Total cost also depends on hardware, batching, and the time spent constructing and solving the generated optimization models}~{\citep{li2026costly}. We do not measure end-to-end latency or monetary cost in this experiment. The additional experiments in this appendix are reported as point estimates, without standard deviations.}}

\begin{table}[htbp]

\centering

\caption{{Effect of training data on PEARL-Qwen3-4B. Training and evaluation settings are unchanged across rows. Scores are macro- and micro-averaged Pass@1 (\%). The combined-data row matches the direct-RL results in Table}~{\ref{tab:sft_ablation}.}}
\label{tab:data_ablation}
\small
\begin{tabular}{lrr}
\toprule
{Training problem pool} & {Macro} & {Micro}\\
\midrule
{Cleaned public 10k} & {64.12} & {78.95}\\
{Curated 3.8k} & {67.61} & {79.23}\\
{Combined 13.8k} & {70.40} & {80.38}\\
\bottomrule
\end{tabular}
\end{table}

\begin{table}[htbp]

\centering

\caption{{Mean generated tokens, execution success, and Pass@1 on the same evaluation instances. All methods use the same solver, code parser, and objective-value check, but retain their own inference procedures. Token counts cover all model responses for each instance, including intermediate turns, and are averaged across instances.}}
\label{tab:token_cost}
\small
\begin{tabular}{lrrr}
\toprule
{Method} & {Mean tokens} & {Execution success (\%)} & {Pass@1 (\%)}\\
\midrule
{PEARL-Qwen3-4B} & {9,868.75} & {90.88} & {79.43}\\
{DeepSeek-V3.2} & {1,465.51} & {91.29} & {80.68}\\
{DeepSeek-V3.2-Think} & {2,323.56} & {91.06} & {79.90}\\
\bottomrule
\end{tabular}
\end{table}

\Needspace{8\baselineskip}
{\subsection{{LLMOPT and other workflow-based methods}}
\label{app:workflow_scaffold_transfer}
{We run Qwen3-4B within LLMOPT's inference workflow}~{\citep{jiang2025llmopt}, without LLMOPT-specific fine-tuning. The model reaches 5.30\% macro-averaged and 6.14\% micro-averaged Pass@1 (Table}~{\ref{tab:llmopt_scaffold_transfer}). Using the workflow alone therefore does not give strong results with this backbone. These scores are not an evaluation of the released LLMOPT-Qwen2.5-14B model, which has undergone task-specific fine-tuning.}}

\begin{table}[htbp]

\centering

\caption{{Pass@1 (\%) for Qwen3-4B using LLMOPT's inference workflow without LLMOPT-specific fine-tuning. These scores do not represent the released LLMOPT-Qwen2.5-14B model.}}
\label{tab:llmopt_scaffold_transfer}
\small
\begin{tabular}{lr}
\toprule
{Benchmark} & {Pass@1 (\%)}\\
\midrule
{NL4Opt} & {4.21}\\
{MAMO-E} & {8.62}\\
{MAMO-C} & {7.21}\\
{NLP4LP} & {5.62}\\
{OptMATH-Bench} & {0.00}\\
{IndustryOR} & {4.76}\\
{ComplexOR} & {5.56}\\
{Resocratic} & {6.45}\\
\midrule
{Macro} & {5.30}\\
{Micro} & {6.14}\\
\bottomrule
\end{tabular}
\end{table}

{{Other released workflows require inputs beyond a natural-language description. OptiMUS}~{\citep{pmlr-v235-ahmaditeshnizi24a} expects preprocessed target and data files, together with a Gurobi executor for linear and mixed-integer linear programs. Chain-of-Experts}~{\citep{xiao_chain--experts_2024} requires a function signature and input/output examples for each instance. Applying these releases to our benchmarks would require additional code to prepare those inputs; the results would depend in part on how this conversion is implemented.}}

{{OptiMind}~{\citep{chen_optimind_2025} also uses multi-turn correction, but with a different backbone and training dataset. Its published scores therefore do not provide a comparison in which those factors are held fixed.}}

{{To test the benefit of PEARL training without changing the backbone or tool access, Appendix}~{\ref{app:subsec:tool_use_depth} (Table}~{\ref{tab:scaffold_control_tool_calls}) compares PEARL with the base Qwen3-4B model using the same tools and turn budget. We do not report equivalent tool-enabled comparisons for the larger DeepSeek and Qwen models, or a comparison with a general-purpose coding agent. Our results therefore do not establish that optimization-specific training is required for every agent workflow.}}

\section{Limitations and Broader Impact}
\label{app:limitations_impact}

{Our evaluation provides evidence for tool-integrated agentic optimization modeling, but it has several limitations. First, PEARL operates under a bounded turn budget and fixed observation limits. This makes training and evaluation tractable, but it may under-estimate problems that require long debugging chains, richer data inspection, or interaction with multiple domain-specific tools. Multi-turn inference also incurs additional serving cost relative to one-shot generation, so future systems should learn when interaction is worth the cost and when a direct solution is sufficient.}

{Second, the benchmark suite covers linear, integer, mixed-integer, nonlinear, and second-order cone programs, but it still under-represents stochastic, robust, online, and highly domain-specific enterprise optimization problems. Real deployments may involve longer documents, richer data interfaces, domain-specific modeling languages, interactive human oversight, or larger industrial instances{, motivating broader evaluation on industrial-scale benchmarks}~{\citep{li2026industrial}}.} %The training data also combines cleaned open-source instances with newly constructed and manually curated examples, so residual collection bias and imperfect labels may remain.

{Third, solver execution and objective-value checks are strong but partial validation signals. They make evaluation auditable and practically meaningful, but they {may fail to detect} ambiguous specifications or formulations that satisfy the tested instance while {violating} broader requirements{. In particular, an under- or over-constrained model may attain the reference optimum and receive the objective-agreement reward despite being semantically incorrect. We do not quantify this false-positive rate. Constraint-satisfaction problems without an optimization objective are outside the current evaluation and would require a verifier for the predicted assignment}. This limitation is common in optimization-modeling evaluation: objective agreement, executability, formulation matching, and solver-checked outputs are practical proxies for correctness, but none can fully score all mathematically equivalent formulations or latent structural defects. Future work should therefore develop richer validators, uncertainty-aware evaluation, and training procedures for broader classes of real-world optimization workflows{.}}

{The potential positive impact of this line of research is substantial. Reliable interactive optimization modeling could lower the expertise barrier that currently limits the use of mathematical optimization in domains such as logistics, manufacturing, planning, energy systems, and resource allocation. By emphasizing solver execution and validation feedback, PEARL may also reduce modeling errors and encourage a more transparent style of LLM-based decision-support generation centered on verifiable artifacts rather than surface-level text.}

{At the same time, this capability carries meaningful risks. Optimization models are often used in high-stakes settings, and flawed formulations may yield infeasible, unsafe, unfair, or economically harmful decisions while still appearing superficially plausible. A system that generates convincing but incorrect solver code could create a false sense of reliability, especially for non-expert users. PEARL should therefore be understood as a sandboxed modeling and verification system, not as evidence that current agents are ready for autonomous deployment in decision-critical environments. All outputs require human interpretation and validation before real-world use. We hope this work helps promote stronger solver-grounded verification, better structural interpretability, clearer human oversight, and safer standards for optimization-oriented AI systems.}

\end{document}